\documentclass{article}

\usepackage{fullpage}
\usepackage{multirow}
\usepackage{amsmath, amsfonts, amssymb}
\usepackage{bm}
\usepackage[dvips]{graphicx}
\usepackage{enumerate}
\usepackage{microtype}
\usepackage{hyperref}
\usepackage{hyperref}
\usepackage{url}
\usepackage{fullpage}
\usepackage{multirow}
\usepackage{amsmath, amsfonts, amssymb}
\usepackage{bm}
\usepackage[dvips]{graphicx}
\usepackage{enumerate}
\usepackage{microtype}
\usepackage{hyperref}
\usepackage{verbatim}

\newcommand{\Prob}{\mathcal{P}}

\newcommand{\tr}{\text{tr}}

\newcommand{\x}{\mathbf{x}}
\newcommand{\y}{\mathbf{y}}
\newcommand{\n}{\mathbf{n}}

\newcommand{\bx}{\bold{x}}
\newcommand{\by}{\bold{y}}

\newtheorem{theorem}{Theorem}[section]

\newtheorem{proposition}[theorem]{Proposition}

\usepackage{graphicx} 
\usepackage{subfigure} 
\usepackage{epsfig,wrapfig}
\usepackage{xcolor}
\usepackage[numbers,sort&compress,comma]{natbib}
\usepackage{bibspacing}

\usepackage{algorithm}
\usepackage[noend]{algorithmic}

\usepackage{hyperref}

\usepackage{todonotes}

\newenvironment{itemize**}%
  {\vspace{-2pt}\begin{itemize}%
    \setlength{\topsep}{2pt}%
    \setlength{\partopsep}{2pt}%
    \setlength{\parsep}{2pt}%
    \setlength{\parskip}{2pt}%
    \setlength{\itemsep}{2pt}}%
  {\end{itemize}\vspace{-6pt}}

\usepackage[accepted]{icml2014} 
\icmltitlerunning{Learning the Parameters of Determinantal Point Process Kernels}

\begin{document} 

\twocolumn[
\icmltitle{Learning the Parameters of Determinantal Point Process Kernels}

\icmlauthor{Raja Hafiz Affandi}{rajara@wharton.upenn.edu}
\icmladdress{University of Pennsylvania}
\icmlauthor{Emily B. Fox}{ebfox@stat.washington.edu}
\icmladdress{University of Washington}
\icmlauthor{Ryan P. Adams}{rpa@seas.harvard.edu}
\icmladdress{Harvard University}
\icmlauthor{Ben Taskar}{taskar@cs.washington.edu}
\icmladdress{University of Washington}
\vskip 0.2in
]

\begin{abstract} 
Determinantal point processes (DPPs) are well-suited for modeling repulsion and have proven useful in many applications where diversity is desired. While DPPs have many appealing properties, such as efficient sampling, learning the parameters of a DPP is still considered a difficult problem due to the non-convex nature of the likelihood function. In this paper, we propose using Bayesian methods to learn the DPP kernel parameters. These methods are applicable in large-scale and continuous DPP settings even when the exact form of the eigendecomposition is unknown. We demonstrate the utility of our DPP learning methods in studying the progression of diabetic neuropathy based on spatial distribution of nerve fibers, and in studying human perception of diversity in images. 
\end{abstract}
 
\section{Introduction}
\label{sec:intro}
%
 A determinantal point process (DPP) provides a distribution over configurations of points. The defining characteristic of the DPP is that it is a repulsive point process, which makes it useful for modeling diversity. Recently, DPPs have played an increasingly important role in machine learning and statistics with applications both in the discrete setting---where they are used as a diverse subset selection method \cite{kulesza2010structured,kulesza2011k,gillenwater2012discovering,affandi2012markov,snoek2013determinantal,Affandi:AISTATS2013}--- and in the continuous setting for generating point configurations that tend to be spread out\cite{affandi2013approximate,zou2012priors}.

Formally, given a space ${\Omega\subseteq\mathbb{R}^d}$, a specific point configuration ${A\subseteq\Omega}$, and a positive semi-definite kernel function~${L:\Omega\times\Omega\rightarrow \mathbb{R}}$, the probability density under a DPP with kernel $L$ is given by
\begin{equation}
\label{eq:DPP}
\Prob_L(A) \propto \det(L_A)~,
\end{equation}
where $L_A$ is the ${|A|\times|A|}$ matrix with entries $L(\bx,\by)$ for each~${\bx,\by\in A}$.  This defines a repulsive point process since point configurations that are more spread out according to the metric defined by the kernel $L$ have higher densities. To see this, recall that the subdeterminant in Eq.~\eqref{eq:DPP} is proportional to the square of the volume spanned by the kernel vectors associated with the points in $A$.

Building on work of \citet{kulesza2010structured}, it is intuitive to decompose the kernel $L$ as
\begin{equation}
L(\bx,\by)=q(\bx)k(\bx,\by)q(\by)~,  
\end{equation}
where $q(\bx)$ can be interpreted as the quality function at point $\bx$ and $k(\bx,\by)$ as the similarity kernel between points $\bx$ and $\by$. The ability to bias the quality in certain locations while still maintaining diversity via the similarity kernel offers great modeling flexibility.

One of the remarkable aspects of DPPs is that they offer efficient algorithms for inference, including computing the marginal and conditional probabilities \cite{kulesza2012determinantal}, sampling \cite{hough2006determinantal,kulesza2010structured,Affandi:AISTATS2013,affandi2013approximate}, and restricting to fixed-sized point configurations ($k$-DPPs)\cite{kulesza2011k}. However, an important component of DPP modeling, learning the DPP kernel parameters, is still considered a difficult, open problem.  Even in the discrete $\Omega$ setting, DPP kernel learning has been conjectured to be NP-hard \cite{kulesza2012determinantal}.  Intuitively, the issue arises from the fact that in seeking to maximize the log-likelihood of Eq.~\eqref{eq:DPP}, the numerator yields a concave log-determinant term whereas the normalizer contributes a convex term, leading to a non-convex objective.  This non-convexity holds even under various simplifying assumptions on the form of~$L$.

Attempts to partially learn the kernel have been studied by, for example, learning the parametric form of the quality function $q(\bx)$ for fixed similarity $k(\bx,\by)$ \cite{kulesza2011learning}, or learning a weighting on a fixed set of kernel experts \cite{kulesza2011k}. So far, the only attempt to learn the parameters of the similarity kernel $k(\bx,\by)$ has used Nelder-Mead optimization \cite{lavancier2012statistical}, which lacks theoretical guarantees about convergence to a stationary point.

In this paper, we consider parametric forms for the quality function $q(\bx)$ and similarity kernel $k(\bx,\by)$ and propose Bayesian methods to learn the DPP kernel parameters $\Theta$. In addition to capturing posterior uncertainty rather than a single point estimate, these methods can be easily modified to efficiently learn large-scale and continuous DPPs where the eigenstructures are either unknown or are inefficient to compute. In contrast, gradient ascent algorithms for maximum likelihood estimation (MLE) require kernels $L$ that are differentiable with respect to $\Theta$ in the discrete $\Omega$ case.  In the continuous $\Omega$ case, the eigenvalues must additionally have a known, differentiable functional form, which only occurs in limited scenarios.  

In Sec.~\ref{sec:background}, we review DPPs and their fixed-sized counterpart ($k$-DPPs). We then explore likelihood maximization algorithms for learning DPP and $k$-DPP kernels. After examining the shortcomings of the MLE approach, we propose a set of techniques for Bayesian posterior inference of the kernel parameters in Sec.~\ref{sec:method}, and explore modifications to accommodate learning large-scale and continuous DPPs. In Sec.~\ref{sec:moments}, we derive a set of DPP moments assuming a known kernel eigenstructure and explore using these moments as a model-checking technique. In low-dimensional settings, we can use a method of moments approach to learn the kernel parameters via numerical techniques. Finally, we test our methods on both simulated and real-world data.  Specifically, in Sec.~\ref{sec:applications} we use DPP learning to study the progression of diabetic neuropathy based on spatial distribution of nerve fibers and also to study human perception of diversity of images.

\section{Background}
\label{sec:background}
\subsection{Discrete DPPs/$k$-DPPs}
For a discrete base set ${\Omega=\{\bx_1,\bx_2,\ldots,\bx_N\}}$, a DPP defined by an ${N\times N}$ positive semi-definite kernel matrix~$L$ is a probability measure on the $2^\Omega$ possible subsets $A$ of $\Omega$:
\begin{equation}
\Prob_L(A)=\frac{\det(L_A)}{\det(L+I)}~.
\end{equation}
Here, ${L_A\equiv[L_{ij}]_{\bx_i,\bx_j\in A}}$ is the submatrix of $L$ indexed by the elements in $A$ and $I$ is the ${N\times N}$ identity matrix \cite{borodin2005eynard}.


In many applications, we are instead interested in the probability distribution which gives positive mass only to subsets of a fixed size, $k$. In  these cases, we consider fixed-sized DPPs (or $k$-DPPs) with probability distribution on sets $A$ of cardinality $k$ given by
\begin{equation}
\Prob_L^k(A)=\frac{\det(L_A)}{e_k(\lambda_1,\ldots,\lambda_N)}~,
\end{equation}
where ${\lambda_1,\ldots,\lambda_N}$ are eigenvalues of~$L$ and
$e_k(\lambda_1,\ldots,\lambda_N)$ is the $k$th elementary symmetric polynomial \cite{kulesza2011k}. 
Note that~$e_k(\lambda_1,\ldots,\lambda_N)$ can be efficiently computed using recursion \cite{kulesza2012determinantal}.

\subsection{Continuous DPPs/$k$-DPPs}
Consider now the case where ${\Omega\subseteq\mathbb{R}^d}$ is a continuous space. DPPs extend to this case naturally, with~$L$ now a kernel operator instead of a matrix. Again appealing to Eq.~\eqref{eq:DPP}, the DPP probability density for point configurations~${A \subset \Omega}$ is given by
\begin{equation}
\label{eq:ContDPP}
\Prob_L(A)=\frac{\det(L_A)}{\prod_{n=1}^\infty(\lambda_n+1)}~,
\end{equation}
where $\lambda_1,\lambda_2,\ldots$ are eigenvalues of the operator $L$.

The $k$-DPP also extends to the continuous case with  
\begin{equation}
\Prob_L^k(A)=\frac{\det(L_A)}{e_k(\lambda_{1:\infty})}~,
\end{equation}
where~${\lambda_{1:\infty} = (\lambda_1,\lambda_2,\ldots)}$.

In contrast to the discrete case, the eigenvalues $\lambda_i$ for continuous DPP kernels are generally unknown; exceptions include a few kernels such as the exponentiated quadratic. However, \citet{affandi2013approximate} showed that a low-rank approximation to $L$ can be used to recover an approximation to a finite truncation of the eigenvalues representing an important part of the eigenspectrum. This enables us to approximate the normalizing constants of both DPPs and $k$-DPPs, and will play a crucial role in our proposed methods of Sec.~\ref{sec:largescale}.

\section{Learning Parametric DPPs}
\label{sec:method}
Assume that we are given a training set consisting of samples $A^1,A^2,\ldots,A^T$, and that we model these data using a DPP/$k$-DPP with parametric kernel
\begin{equation}
L(\bx,\by;\Theta)=q(\bx;\Theta)k(\bx,\by;\Theta)q(\by;\Theta)~,
\end{equation}
with parameters $\Theta$. We denote the associated kernel matrix for a set $A^t$ by $L_{A^t}(\Theta)$ and the full kernel matrix/operator by $L(\Theta)$.  Likewise, we denote the kernel eigenvalues by $\lambda_i(\Theta)$.  In this section, we explore various methods for DPP/$k$-DPP learning.
%
\subsection{Learning using Optimization Methods}
\label{sec:MLE}
To learn the parameters $\Theta$ of a discrete DPP model, we can maximize the log-likelihood
\begin{equation}
\mathcal{L}(\Theta)=\sum_{t=1}^T\log\det(L_{A^t}(\Theta))-T\log\det(L(\Theta)+I)~.
\end{equation}

\citet{lavancier2012statistical} suggests that the Nelder-Mead simplex algorithm \cite{nelder1965simplex} can be used to maximize $\mathcal{L}(\Theta)$. This method is based on evaluating the objective function at the vertices of a simplex, then iteratively shrinking the simplex towards an optimal point. While this method is convenient since it does not require explicit knowledge of derivates of $\mathcal{L}(\Theta)$, it is regarded as a heuristic search method and is known for its failure to necessarily converge to a stationary point \cite{mckinnon1998convergence}. 

Gradient ascent and stochastic gradient ascent provide more attractive approaches because of their theoretical guarantees, but require knowledge of the gradient of~$\mathcal{L}(\Theta)$.  In the discrete DPP setting, this gradient can be computed straightforwardly, and we provide examples for discrete Gaussian and polynomial kernels in the Supplementary Material. We note, however, that these methods are still susceptible to convergence to local optima due to the non-convex likelihood landscape. 

The log likelihood of the $k$-DPP kernel parameter is

\vspace{-25pt}
\begin{align}
\mathcal{L}(\Theta)=\sum_{t=1}^T\log\det(L_{A^t}(\Theta))-T\log\sum_{|B|=k}\det(L_B(\Theta))~,
\end{align}
\vspace{-22pt}

which presents an addition complication due to needing a sum over ${n\choose k}$ terms in the gradient.

For continuous DPPs/$k$-DPPs, gradient ascent can only be used in cases where the exact eigendecomposition of the kernel operator is known with a differentiable form for the eigenvalues (see Eq.~\eqref{eq:ContDPP}). This restricts the applicability of gradient-based likelihood maximization to a limited set of scenarios, such as a DPP 
with Gaussian quality function and similarity kernel. Furthermore, for kernel operators with infinite rank (such as the Gaussian), an explicit truncation has to be made, resulting in an approximate gradient of $\mathcal{L}(\Theta)$. Unfortunately, such approximate gradients are not unbiased estimates of the true gradient, so the theory associated with attractive stochastic gradient based approaches does not hold.
\subsection{Bayesian Learning for Discrete DPPs}
Instead of optimizing the likelihood to get an MLE, here we propose a Bayesian approach to that samples from the posterior distribution over kernel parameters:
\begin{equation}
\label{eq:PostDPP}
\Prob(\Theta|A^1,\ldots,A^T)\propto \Prob(\Theta)\prod_{t=1}^T\frac{\det(L_{A^t}(\Theta))}{\det(L(\Theta)+I)}
\end{equation}
for the DPP and, for the $k$-DPP,
\begin{equation}
\label{eq:PostkDPP}
\Prob(\Theta|A^1,\ldots,A^T)\propto \Prob(\Theta)\prod_{t=1}^T\frac{\det(L_{A^t}(\Theta))}{e_k(\lambda_1(\Theta),\ldots,\lambda_N(\Theta))}.
\end{equation}
Here, $\Prob(\Theta)$ is the prior on $\Theta$. Since neither Eq.~\eqref{eq:PostDPP} nor Eq.~\eqref{eq:PostkDPP} yield a closed form posterior, we resort to approximate techniques based on Markov chain Monte Carlo (MCMC). We highlight two techniques: random-walk Metropolis-Hastings (MH) and slice sampling, although other MCMC methods can be employed without loss of generality. 

\comment{\begin{figure}
  \centering
     \includegraphics[scale=0.25]{figs/LearnContDiag2DAllGamma}
      \includegraphics[scale=0.25]{figs/LearnContDiag2DAllSigma}
\todo[inline]{What do red and blue mean?}
  \caption{Learning a 2D discrete DPP with Gaussian quality and similarity and two different starting points (denoted by colors) using random walk MH (solid) and gradient ascent (dashed). Relative Frobenious norm error of the (\emph{left}) quality function parameter, $\frac{\|\Gamma-\Gamma_t\|_F}{\|\Gamma\|_F}$, and (\emph{right}) similarity kernel parameter, $\frac{\|\Sigma-\Sigma_t\|_F}{\|\Sigma\|_F}$.}
  \label{fig:MLE}
\end{figure}}
In random-walk MH, we use a proposal distribution~$f(\hat{\Theta}|\Theta_i)$ to generate a candidate value $\hat{\Theta}$ given the current parameters $\Theta_i$, which are then accepted or rejected with probability $\min\{r,1\}$ where
\begin{equation}
r=\left(\frac{\Prob(\hat{\Theta}|A^1,\ldots,A^T)}{\Prob(\Theta_i|A^1,\ldots,A^T)}\frac{f(\Theta_i|\hat{\Theta})}{f(\hat{\Theta}|\Theta_i)}\right)~.
\end{equation}
The proposal distribution $f(\hat{\Theta}|\Theta_i)$ is chosen to have mean $\Theta_i$. 
The hyperparameters of $f(\hat{\Theta}|\Theta_i)$ tune the width of the distribution, determining the average step size.  
See Alg.~1 
of the Supplementary Material.


While random-walk MH can provide a straightforward means of sampling from the posterior, its efficiency requires tuning the proposal distribution.  Choosing an aggressive proposal can result in a high rejection rate, while choosing a conservative proposal can result in inefficient exploration of the parameter space. 
To avoid the need to tune the proposal distribution, we can instead use slice sampling \cite{neal2003slice}, which performs a local search for an acceptable point while still satisfying detailed balance conditions.  We first describe this method in the univariate case, following the ``linear stepping-out'' approach described in \citet{neal2003slice}. Given the current parameter $\Theta_i$, we first sample $y\sim \textrm{Uniform}[0,\Prob(\Theta_i|A^1,\ldots,A^T)]$. This defines our \emph{slice} with all values of $\Theta$ with~$\Prob(\Theta|A^1,\ldots,A^T)$ greater than $y$ included in the slice.  We then define a random interval around $\Theta_i$ with width $w$ 
that is linearly expanded until neither endpoint is in the slice.
We propose $\hat{\Theta}$ uniformly in the interval. If $\hat{\Theta}$ is in the slice, it is accepted. Otherwise, $\hat{\Theta}$ becomes the new boundary of the interval,
 shrinking it so as to still include the current state of the Markov chain. This procedure is repeated until a proposed $\hat{\Theta}$ is accepted. See Alg. 2 of the Supplementary Material. 

There are many ways to extend this algorithm to a multidimensional setting. We consider the simplest extension proposed by \citet{neal2003slice} where we use hyperrectangles instead of intervals. A hyperrectangle region is constructed around $\Theta_i$ and the edge in each dimension is expanded or shrunk depending on whether its endpoints lie inside or outside the slice. One could alternatively consider coordinate-wise or random-direction approaches to multidimensional slice sampling.

\begin{figure}
  \centering
     \includegraphics[scale=0.25]{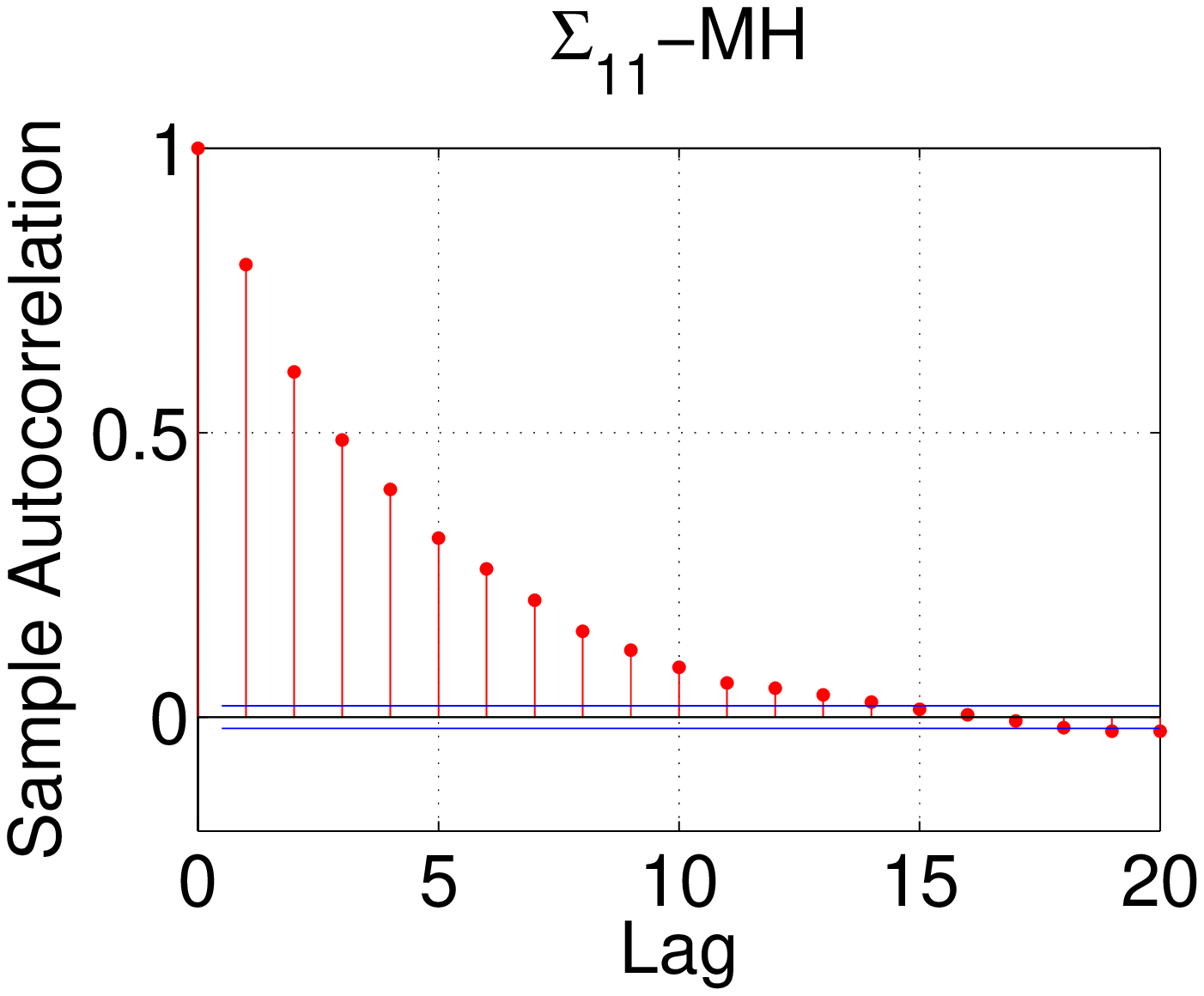}
      \includegraphics[scale=0.25]{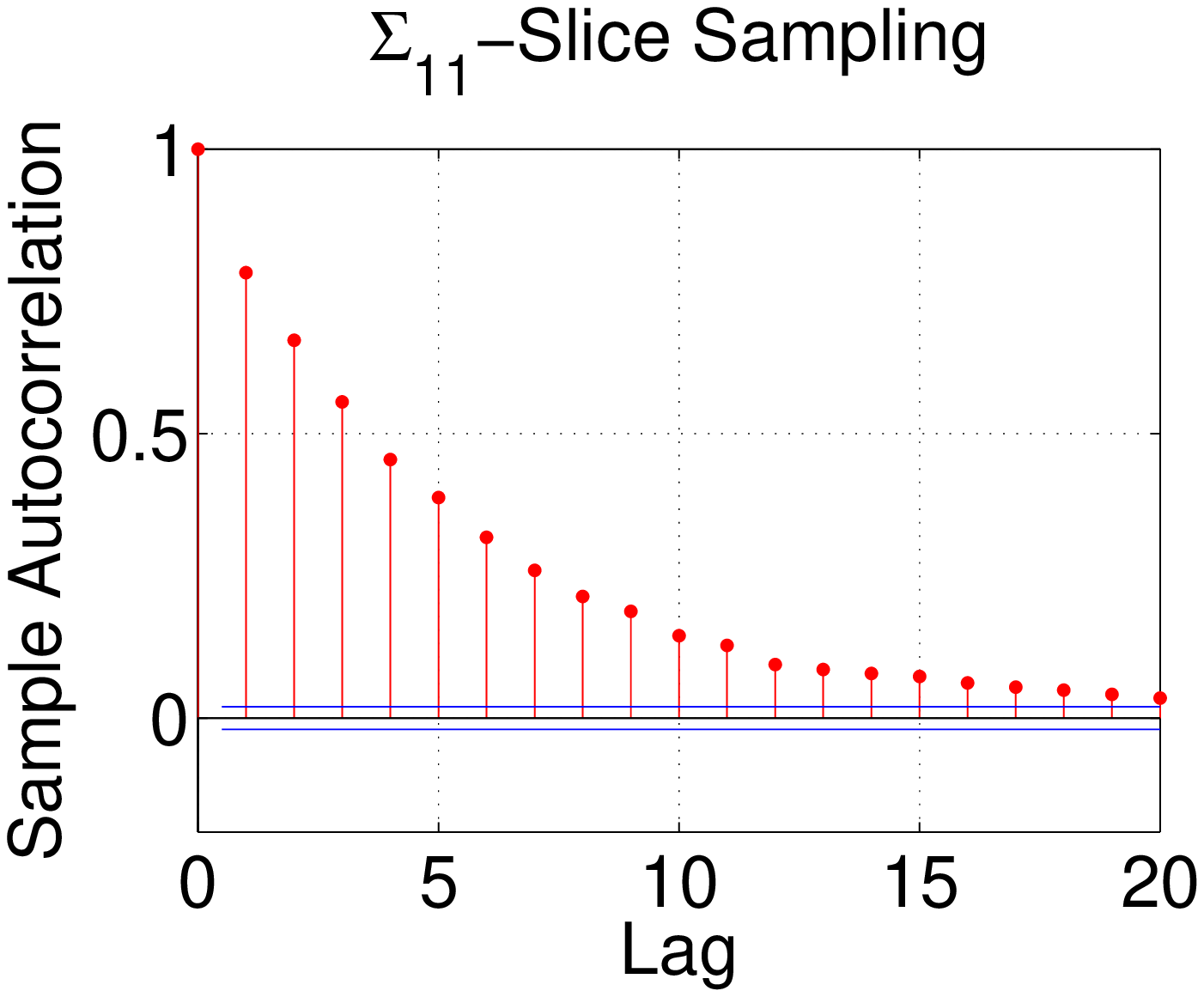}
\vspace{-10pt}
  \caption{Sample autocorrelation function for posterior samples of the slowest mixing parameter of the kernel in Eq.~\eqref{eq:qdiscrete} and Eq.~\eqref{eq:kdiscrete} sampled using MH and slice sampling.}
  \label{fig:conv}
\vspace{-10pt}
\end{figure}

As an illustrative example, we consider synthetic data 
generated from 
a two-dimensional discrete DPP using a kernel where 
\vspace{-5pt}
\begin{align}
\label{eq:qdiscrete}
q(\bx_i)&=\exp{\left\{-\frac{1}{2}\bx_i^\top\Gamma^{-1}\bx_i\right\}}\\
\label{eq:kdiscrete}
k(\bx_i,\bx_j)&=\exp{\left\{-\frac{1}{2}(\bx_i\!-\!\bx_j)^\top\Sigma^{-1}(\bx_i\!-\!\bx_j)\right\}}~,
\end{align}
where ${\Gamma=\mbox{diag}(0.5,0.5)}$ and ${\Sigma=\mbox{diag}(0.1,0.2)}$. 
We consider $\Omega$ to be a grid of 100 points evenly spaced in a $10\times 10$ unit square and simulate 100 samples from a DPP with kernel as above.  We then condition on these simulated data and perform posterior inference of the kernel parameters using MCMC. Fig.~\ref{fig:conv} shows the sample autocorrelation function of the slowest mixing parameter, $\Sigma_{11}$, 
learned using random-walk MH and slice sampling. Furthermore, we ran a Gelman-Rubin test~\cite{gelman1992inference} on 5 chains starting from overdispersed starting positions and found that the average partial scale reduction function across the four parameters to be 1.016 for MH and 1.023 for slice sampling, indicating fast mixing of the posterior samples.

%
\subsection{Bayesian Learning for Large-Scale Discrete and Continuous DPPs}\label{sec:largescale}
In the large-scale discrete or continuous settings, evaluating the normalizers~$\det(L(\Theta)+I)$ or~$\prod_{n=1}^\infty(\lambda_n(\Theta)+1)$, respectively, can be inefficient or infeasible. Even in cases where an explicit form of the truncated eigenvalues can be computed, this will only lead to approximate MLE solutions, as discussed in Sec.~\ref{sec:MLE}. 

On the surface, it seems that most MCMC algorithms will suffer from the same problem since they require knowledge of the likelihood as well. However, we argue that for most of these algorithms, an upper and lower bound of the 
posterior probability is sufficient as long as we can control the accuracy of these bounds. In particular, denote the upper and lower bounds by $\Prob^+(\Theta|A^1,\ldots,A^T)$ and $\Prob^-(\Theta|A^1,\ldots,A^T)$, respectively. In the random-walk MH algorithm 
we can then compute the upper and lower bounds on the acceptance ratio, 
\begin{align}
r^+&=\left(\frac{\Prob^+(\hat{\Theta}|A^1,\ldots,A^T)}{\Prob^-(\Theta_i|A^1,\ldots,A^T)}\frac{f(\Theta_i|\hat{\Theta})}{f(\hat{\Theta}|\Theta_i)}\right)\\
r^-&=\left(\frac{\Prob^-(\hat{\Theta}|A^1,\ldots,A^T)}{\Prob^+(\Theta_i|A^1,\ldots,A^T)}\frac{f(\Theta_i|\hat{\Theta})}{f(\hat{\Theta}|\Theta_i)}\right)~.
\end{align}
We can precompute the threshold~${u \sim \textrm{Uniform}[0,1]}$, so we can still sometimes accept or reject the proposal~$\hat{\Theta}$ even if these bounds have not completely converged.  All that is necessary is for~${u<\min\{1,r^-\}}$ (immediately reject) or~${u>\min\{1,r^+\}}$ (immediately accept).  In the case that~${u\in(r^{-},r^{+})}$, we can perform further computations to increase the accuracy of our bounds until a decision can be made.  As we only sample~$u$ once in the beginning, this iterative procedure yields a Markov chain with the exact target posterior as its stationary distribution; all we have done is ``short-circuit'' the computation once we have bounded the acceptance ratio~$r$ away from~$u$.  We show this procedure in Alg.~3 of the Supplementary Material.
\comment{\begin{algorithm}[tb]
 \caption{Random-Walk Metropolis-Hastings with Posterior Bounds}
  \label{alg:RWMH-EB}
\begin{algorithmic}
\STATE{Input:} Dimension: $D$, , Starting point: $\Theta_0$, Prior distribution: $\Prob(\Theta)$,  Proposal distribution $f(\hat{\Theta}|\Theta)$ with mean $\Theta$, samples: $\mathcal{A}=[A^{1},\ldots,A^{T}]$.
\STATE $\Theta=\Theta_0$
\FOR{$i=0:\tau$}
	\STATE $\hat{\Theta}\sim f(\hat{\Theta}|\Theta_i)$
	\STATE $r_{+}=\infty, r_{-}=-\infty$
\STATE $u\sim$ Uniform[0,1] 
\WHILE{$u\in[r_{-},r_{+}]$}               
\STATE $r^{+}=\left(\frac{\Prob^+(\hat{\Theta}|A^1,\ldots,A^T)}{\Prob^-(\Theta_i|A^1,\ldots,A^T)}\frac{f(\Theta_i|\hat{\Theta})}{f(\hat{\Theta}|\Theta_i)}\right)$
\STATE $r^{-}=\left(\frac{\Prob^-(\hat{\Theta}|A^1,\ldots,A^T)}{\Prob^+(\Theta_i|A^1,\ldots,A^T)}\frac{f(\Theta_i|\hat{\Theta})}{f(\hat{\Theta}|\Theta_i)}\right)$
\STATE Increase tightness on $\Prob^+$ and $\Prob^-$
\ENDWHILE
\IF{$u<\min\{1,r^-\}$}
	\STATE $\Theta_t=\hat{\Theta}$
\ENDIF
\ENDFOR
\STATE{Output: $\Theta_{0:\tau}$}
\end{algorithmic}
\end{algorithm}}

The same idea applies to slice sampling. In the first step of generating a slice, instead of sampling $y\sim\textrm{Uniform}[0,\Prob(\Theta_i|A^1,\ldots,A^T)]$, we use a rejection sampling scheme first propose a candidate slice as 
\begin{equation}
\hat{y}\sim\textrm{Uniform}[0,\Prob^{+}(\Theta_i|A^1,\ldots,A^T)]~.
\end{equation}
We then decide whether $\hat{y}<\Prob^{-}(\Theta_i|A^1,\ldots,A^T)$, in which case we know $\hat{y}<\Prob(\Theta_i|A^1,\ldots,A^T)$ and we accept $\hat{y}$ as the slice and set ${y = \hat{y}}$.  In the case where $\hat{y}\in(\Prob^{-}(\Theta_i|A^1,\ldots,A^T),\Prob^{+}(\Theta_i|A^1,\ldots,A^T))$, we keep increasing the tightness of our bounds until a decision can be made. If at any point $\hat{y}$ exceeds the newly computed $\Prob^{+}(\Theta_i|A^1,\ldots,A^T)$, we know that $\hat{y} >\Prob(\Theta_i|A^1,\ldots,A^T)$ so we reject the proposal.  In this case, we generate a new $\hat{y}$ and repeat. 

Upon accepting a slice $y$, the subsequent steps for proposing a parameter $\hat{\Theta}$ proceed in a similarly modified manner.  For the interval computation, the endpoints $\Theta_e$ are each examined to decide whether $y<\Prob^{-}(\Theta_e|A^1,\ldots,A^T)$ (endpoint is not in slice) or $y>\Prob^{+}(\Theta_e|A^1,\ldots,A^T)$ (endpoint is in slice).  The tightness of the posterior bounds is increased until a decision can be made and the interval can be adjusted, if need be. After convergence of the interval, $\hat{\Theta}$ is generated uniformly over the interval and is likewise tested for acceptance.
We illustrate this procedure in 
Fig.~1 of the Supplementary Material.

\begin{figure}
  \centering
     \includegraphics[scale=0.25]{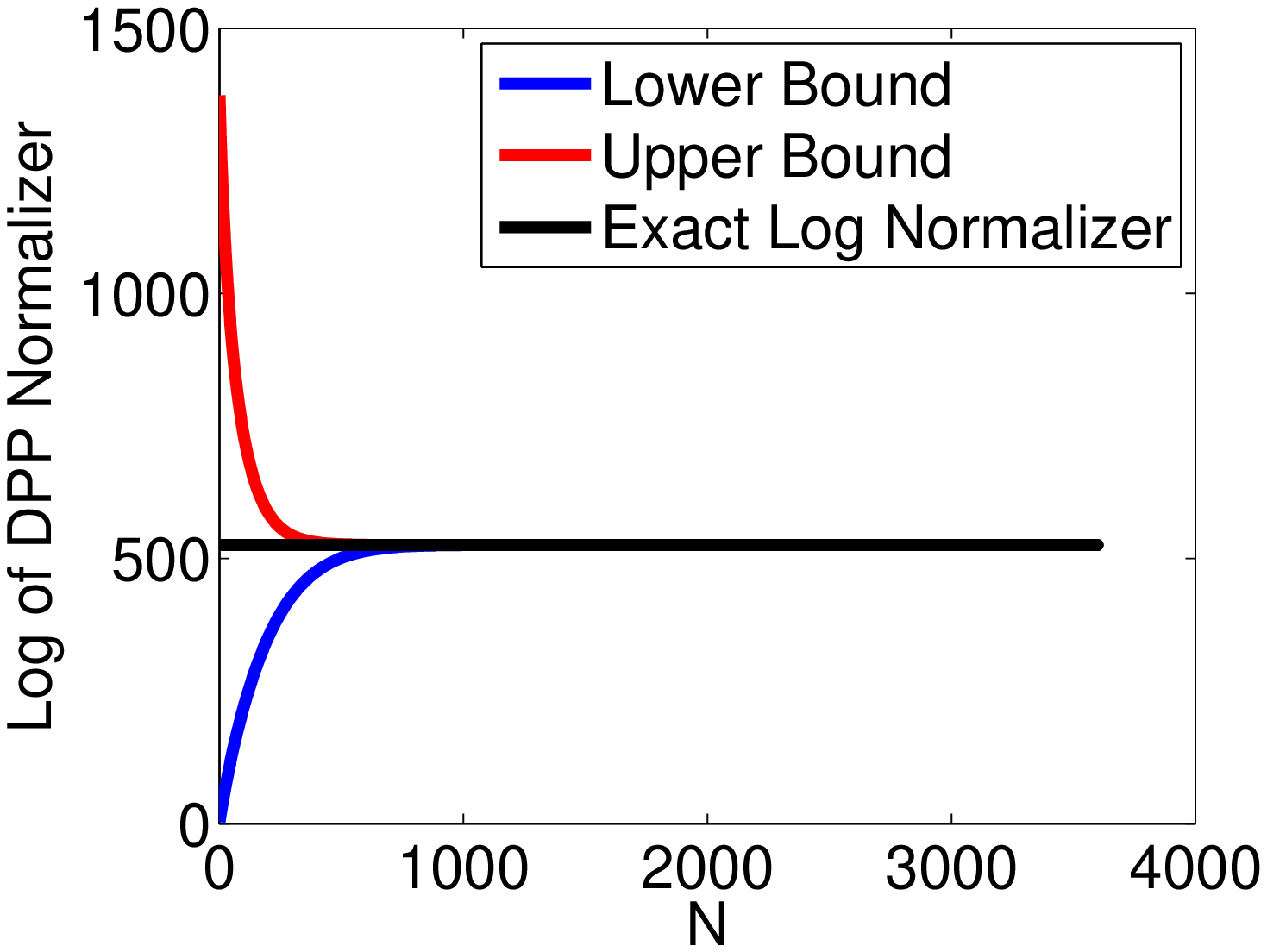}
      \includegraphics[scale=0.25]{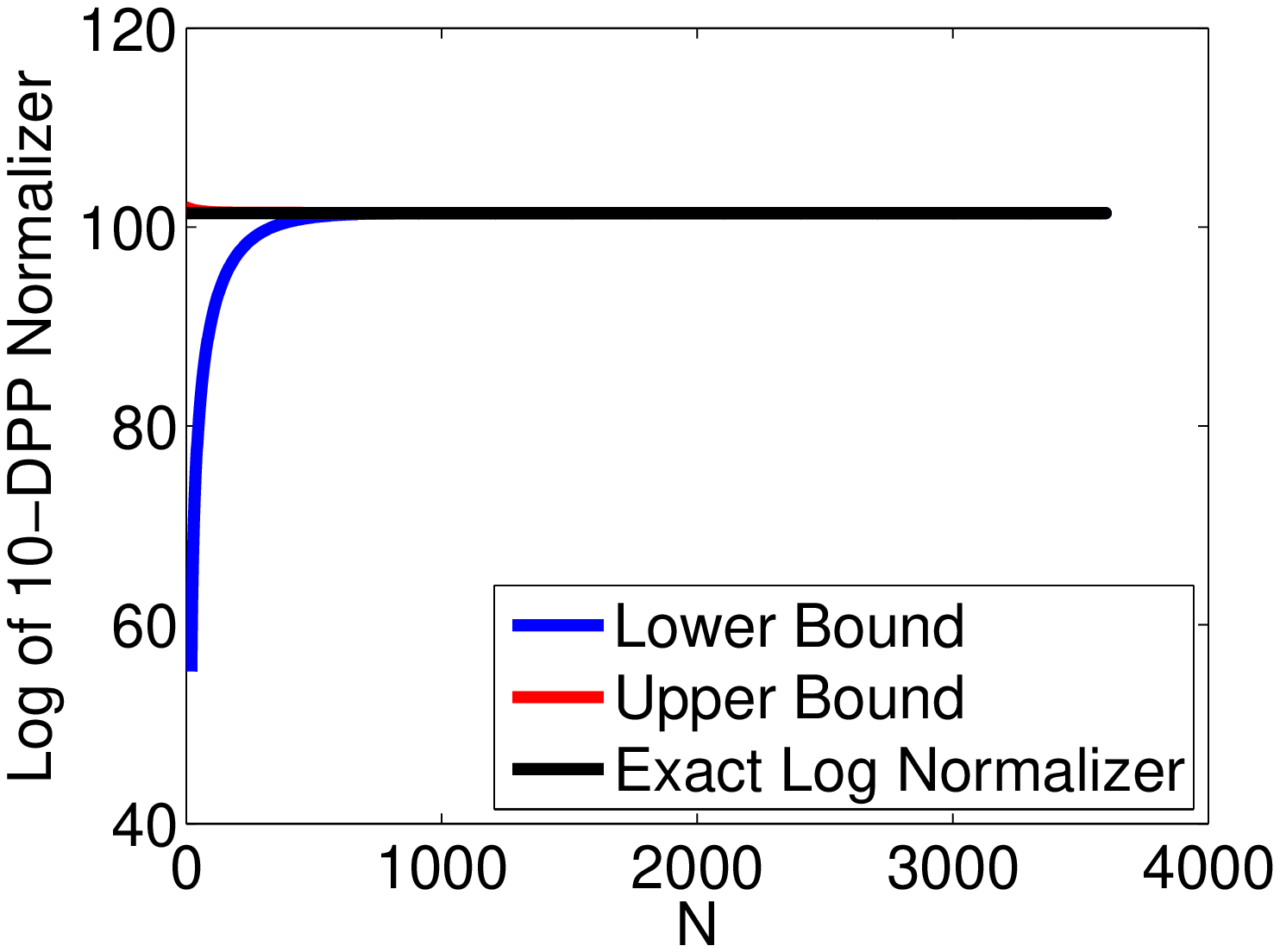}  
\vspace{-10pt}
  \caption{Normalizer bounds for a discrete DPP (\emph{left}) and a 10-DPP (\emph{right}) with Gaussian quality and similarity as in Eqs.~\eqref{eq:qdiscrete} and \eqref{eq:kdiscrete} and $\Omega$ a grid of 3600 points.}
  \label{fig:Bounds}
\vspace{-10pt}
\end{figure}

The lower and upper bounds of the posterior probability can in fact be incorporated in many MCMC-type algorithms. This provides a convenient and efficient way to garner posterior samples assuming that tightening the bounds can be done efficiently. In our case, the upper and lower bounds for the posterior probability depends on the truncation of the kernel eigenvalues and can be arbitrarily tightened by including more terms in the truncation. In the discrete DPP/$k$-DPP settings, the eigenvalues can be efficiently computed to a specified point using methods such as power law iterations. The corresponding bounds for a $3600 \times 3600$ Gaussian kernel example are shown in Fig.~\ref{fig:Bounds}.  In the continuous setting, explicit truncation can be done when the kernel has Gaussian quality and similarity, as we show in Sec.~\ref{sec:LearnGaussianCont}. For other continuous DPP kernels, low-rank approximations can be used \cite{affandi2013approximate} resulting in approximate posterior samples. In contrast, a gradient ascent algorithm for MLE is not even feasible: we do not know the form of the approximated eigenvalues, so we cannot take their derivative.

Explicit forms for the posterior probability bounds of $\Theta$ for DPPs and $k$-DPPs as a function of the eigenvalue truncations follow from Prop. \ref{prop:DPP} and \ref{prop:kDPP} combined with Eqs.~\eqref{eq:PostDPP} and \eqref{eq:PostkDPP}, respectively. Proofs are in the Supplementary Material.
\begin{proposition}
\label{prop:DPP}
Let $\lambda_{1:\infty}$ be the eigenvalues of kernel $L$. Then 
\begin{equation}
\prod_{n=1}^M(1+\lambda_n)\le \prod_{n=1}^\infty(1+\lambda_n)
\end{equation}
and
\begin{equation}
\prod_{n=1}^\infty(1+\lambda_n)\le \exp\bigg\{ \textrm{\normalfont tr}(L)-\sum_{n=1}^M\lambda_n\bigg\}\left[\prod_{n=1}^M(1+\lambda_n)\right]~.
\end{equation}
\end{proposition}
\begin{proposition}
\label{prop:kDPP}
Let $\lambda_{1:\infty}$ be the eigenvalues of kernel $L$. Then 
\begin{equation}
e_k(\lambda_{1:M})\le e_k(\lambda_{1:\infty})
\end{equation}
and
\begin{equation}
e_k(\lambda_{1:\infty})\le \sum_{j=0}^k \frac{(\textrm{\normalfont tr}(L)-\sum_{n=1}^M\lambda_n)^j}{j!}e_{k-j}(\lambda_{1:M})~.
\end{equation}
\end{proposition}
Finally note that the expression $\tr(L)$ in the bounds can be easily computed as either $\sum_{n=1}^N L_{ii}$ in the discrete case or $\int_{\Omega}L(\bx,\bx)d\bx$ in the continuous case. 
%

\section{Method of Moments}
\label{sec:moments}

Convergence and mixing of MCMC samplers can be challenging to assess.  Although generic techniques such as Gelman-Rubin diagnostics~\cite{gelman1992inference} are applicable, we additionally provide a set of tools more directly tailored to the DPP by deriving a set of theoretical moments.  When performing posterior inference of kernel parameters, we can check whether the moments of our data match the theoretical moments given by the posterior samples. This can be done in cases where the eigenstructure is fully known. 

In the discrete case, we first need to compute the marginal probabilities. \citet{borodin2009determinantal} shows that the marginal kernel, $K$, can be computed directly from $L$:
\begin{equation}
K=L(I+L)^{-1}~.
\end{equation} 
The $m$th moment can then be calculated via
\begin{equation}
\label{eq:momdiscrete}
\mathbb{E}[\bx^m]=\sum_{i=1}^N \bx_i^m K(\bx_i,\bx_i)~.
\end{equation} 
In the continuous case, given the eigendecomposition of the kernel operator, $L(\bx,\by)=\sum_{n=1}^\infty \lambda_n\phi_n(\bx)^*\phi_n(\by)$ (where $\phi_n(\bx)^*$ denotes the complex conjugate of the~$n$th eigenfunction), the $m$th moment is
\begin{equation}
\label{eq:momCont}
\mathbb{E}[\bx^m]=\int_{\Omega}\sum_{n=1}^\infty\frac{\lambda_n}{\lambda_n+1}\bx^m\phi_n(\bx)^2d\bx~.
\end{equation} 

Note that this generally cannot be evaluated in closed form since the eigendecompositions of most kernel operators are not known. However, in certain cases where the eigenfunctions are known analytically, the moments can be directly computed. For a kernel defined by Gaussian quality and similarity (see Sec.~\ref{sec:LearnGaussianCont}), the eigendecomposition can be performed using Hermite polynomials. 
In the Supplementary Material, we derive the $m$th moment in this setting. 



Unfortunately, the method of moments can be challenging to use for direct parameter learning since Eqs.~\eqref{eq:momdiscrete} and \eqref{eq:momCont} are not analytically available in most cases.  In low dimensions, these quantities can be estimated numerically, but it remains an open question as to how these moments should be estimated for large-scale problems.

\section{Experiments}
\label{sec:applications}
\subsection{Simulations}
\label{sec:LearnGaussianCont}
We provide an explicit example of Bayesian learning for a continuous DPP with the kernel defined by
\vspace{-5pt} 
\begin{align}
\label{eq:qGaussian}
q(\bx)&=\sqrt{\alpha}\prod_{d=1}^D\frac{1}{\sqrt{\pi\rho_d}}\exp\left\{-\frac{x_d^2}{2\rho_d}\right\}\\
\label{eq:kGaussian}
k(\bx,\by)&=\prod_{d=1}^D\exp\left\{-\frac{(x_d-y_d)^2}{2\sigma_d}\right\}, \bx,\by\in\mathbb{R}^D.
\end{align}
Here, $\Theta = \{\alpha,\rho_d,\sigma_d\}$ and the eigenvalues of the operator $L(\Theta)$ are given by \cite{fasshauer2012stable},
\begin{equation}
\lambda_\mathbf{m}(\Theta)=\alpha\prod_{d=1}^D\sqrt{\frac{1}{\frac{\beta_d^2+1}{2}+\frac{1}{2\gamma_d}}}\bigg(\frac{1}{\gamma_d(\beta_d^2+1)+1}\bigg)^{m_d-1}, \end{equation}
where $\gamma_d=\frac{\sigma_d}{\rho_d}$, $\beta_d=(1+\frac{2}{\gamma_d})^{\frac{1}{4}}$, and $\mathbf{m} = (m_1,\dots,m_D)$ is a multi-index.

Furthermore, the trace of $L(\Theta)$ can be easily computed as
\vspace{-10pt}
\begin{equation}
\tr(L(\Theta))=\int_{\mathbb{R}^d}\alpha\prod_{d=1}^D\frac{1}{\pi\rho_d}\exp\left\{-\frac{x_d^2}{2\rho_d}\right\}d\bx=\alpha~.
\end{equation}
\vspace{-15pt}
\comment{Thus, upper and lower bounds for the likelihood can be explicitly calculated and our proposed Bayesian learning algorithms are applicable.}

We test our Bayesian learning algorithms on simulated data generated from a 2-dimensional isotropic kernel (${\sigma_d=\sigma}$, ${\rho_d=\rho}$ for ${d=1,2}$) using Gibbs sampling \cite{affandi2013approximate}. We then learn the parameters under weakly informative inverse gamma priors on $\sigma$, $\rho$ and $\alpha$. Details are in the Supplementary Material. \comment{Since the expected number of points is given by ${\sum_{n=1}^\infty\frac{\lambda_n}{\lambda_n+1}}$ \cite{kulesza2010structured},}We tweak the $(\alpha,\rho,\sigma)$ used for simulation so that we have the following three scenarios: 
\vspace{-7pt}
\begin{itemize**}
\item[(i)] 10 DPP samples with average number of points=18 using $(\alpha,\rho,\sigma)=(1000,1,1)$
\item[(ii)] 1000 DPP samples with average number of points=18 using $(\alpha,\rho,\sigma)=(1000,1,1)$
\item[(iii)] 10 DPP samples with average number of points=77 using $(\alpha,\rho,\sigma)=(100,0.7,0.05)$.
\end{itemize**}
\vspace{-3pt}
\begin{figure*}
  \centering
     \includegraphics[scale=0.2]{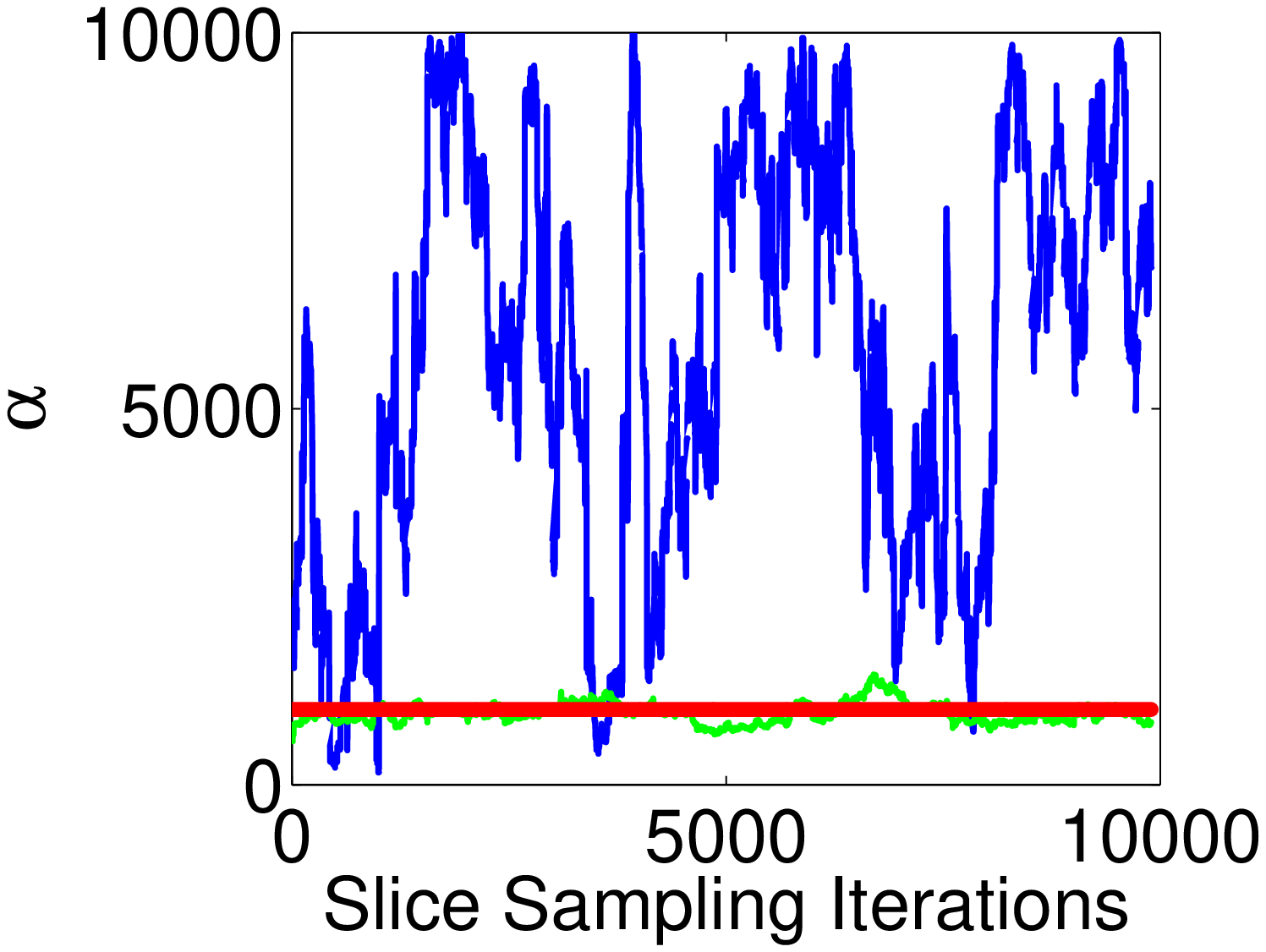}
      \includegraphics[scale=0.2]{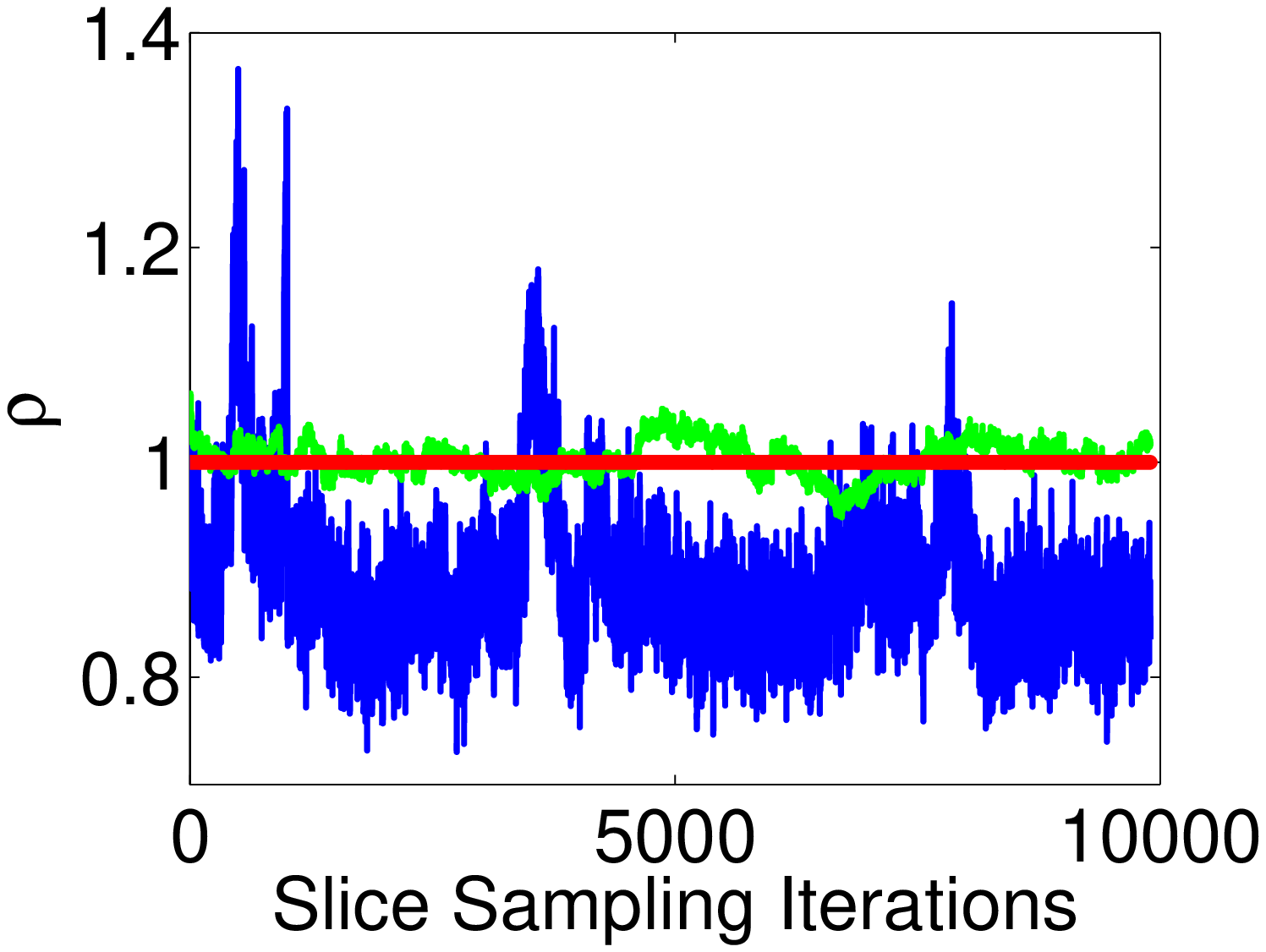}
\includegraphics[scale=0.2]{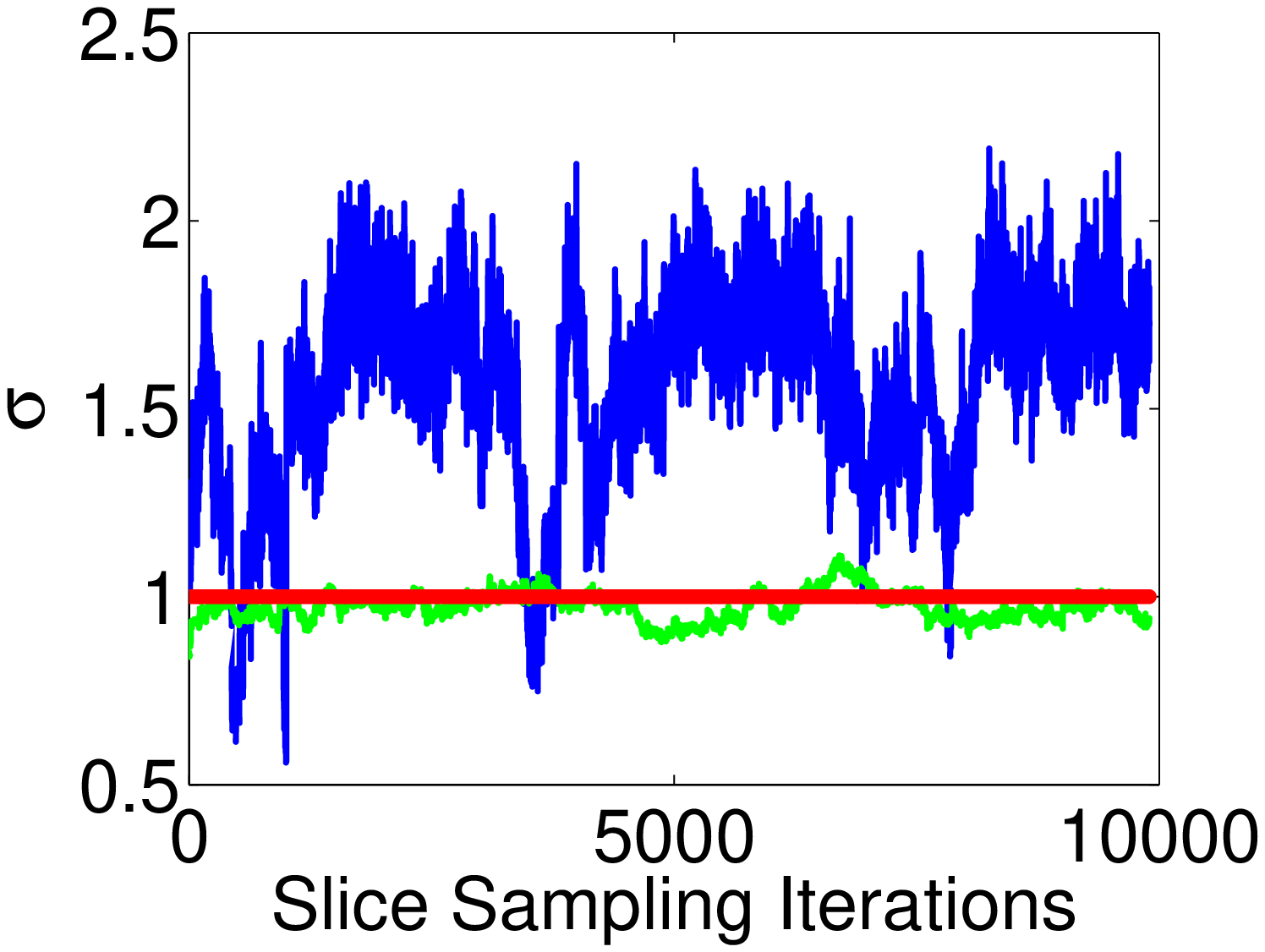}
\includegraphics[scale=0.2]{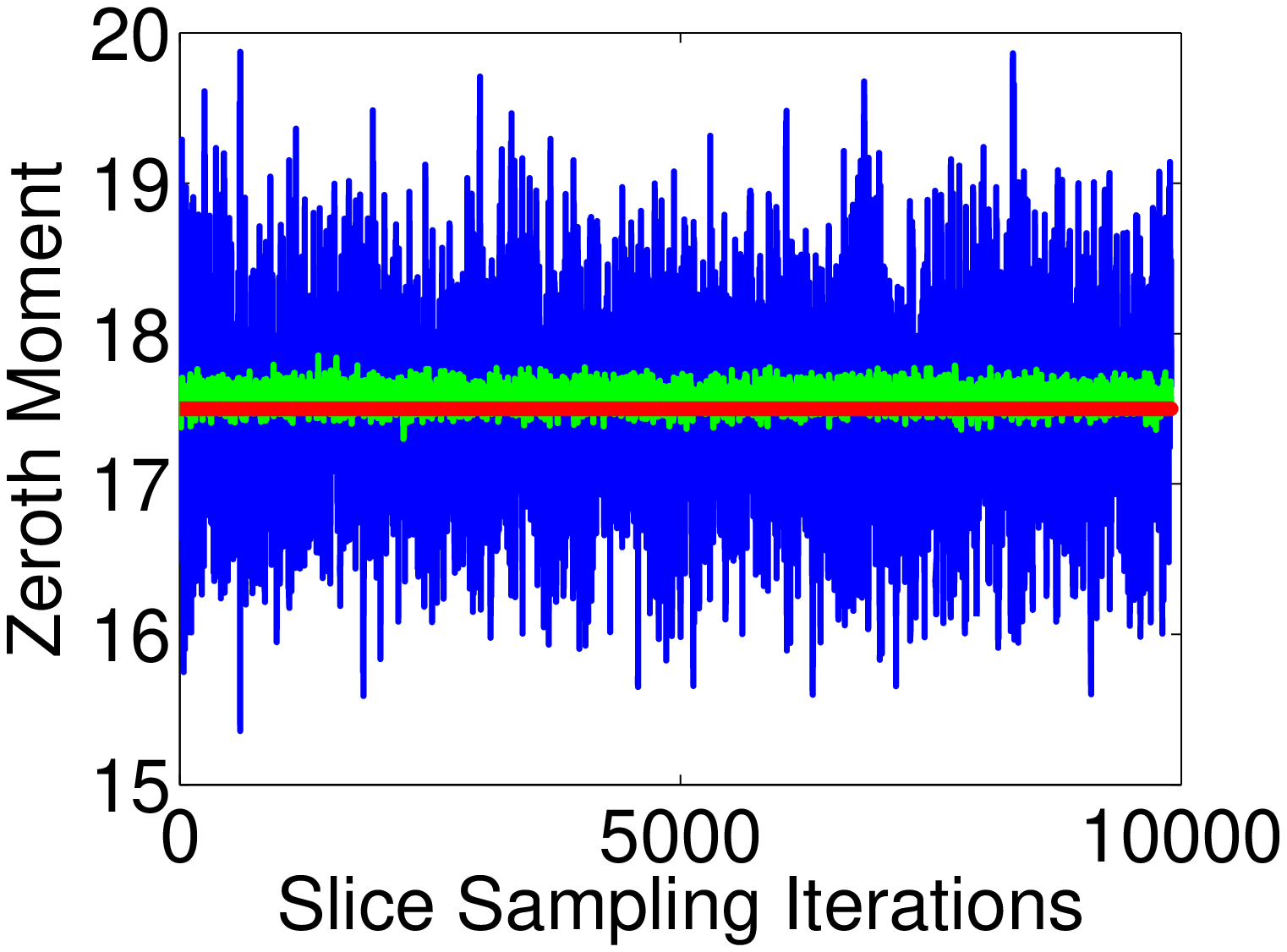}
\includegraphics[scale=0.2]{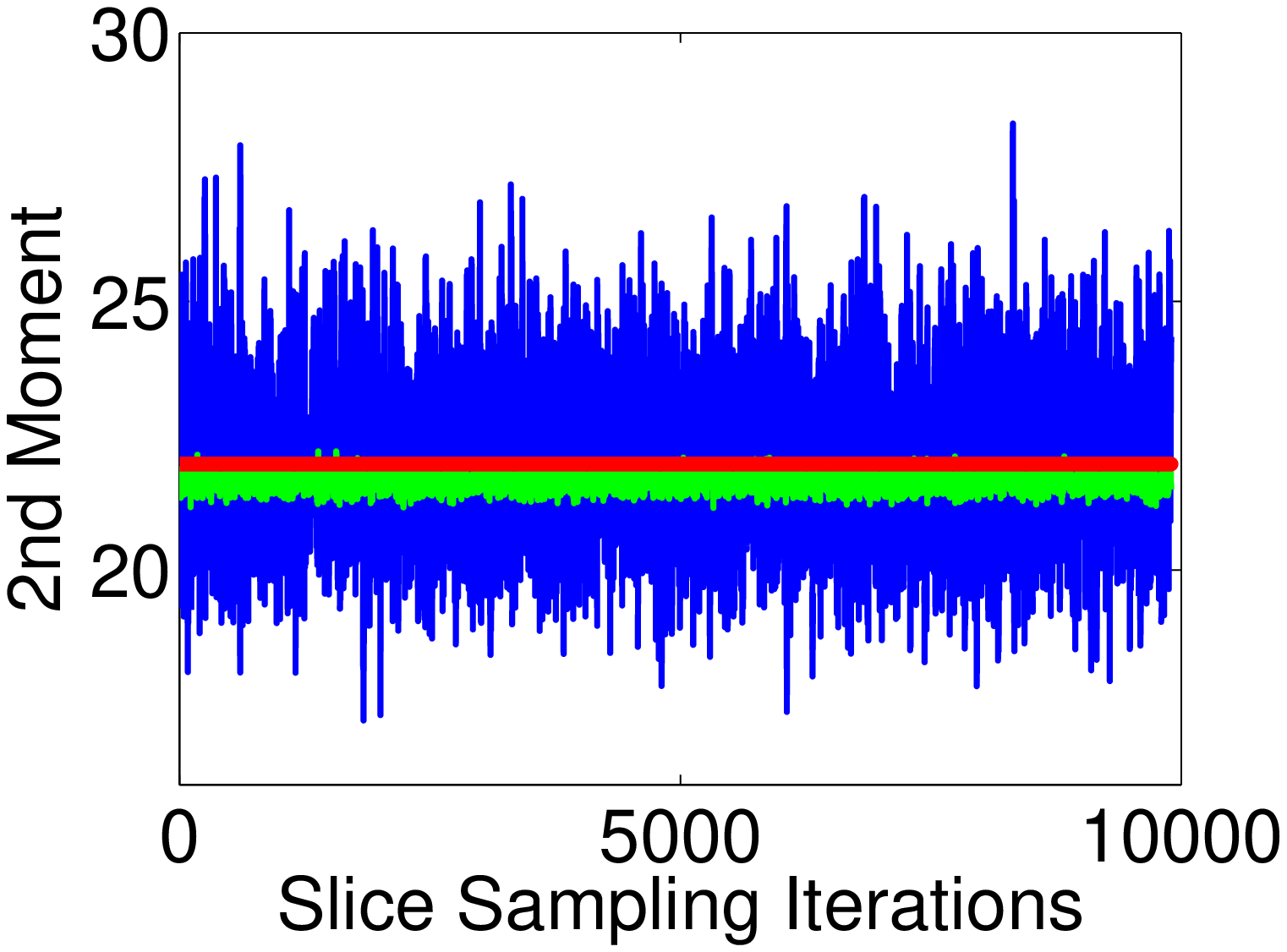}\\
     \includegraphics[scale=0.2]{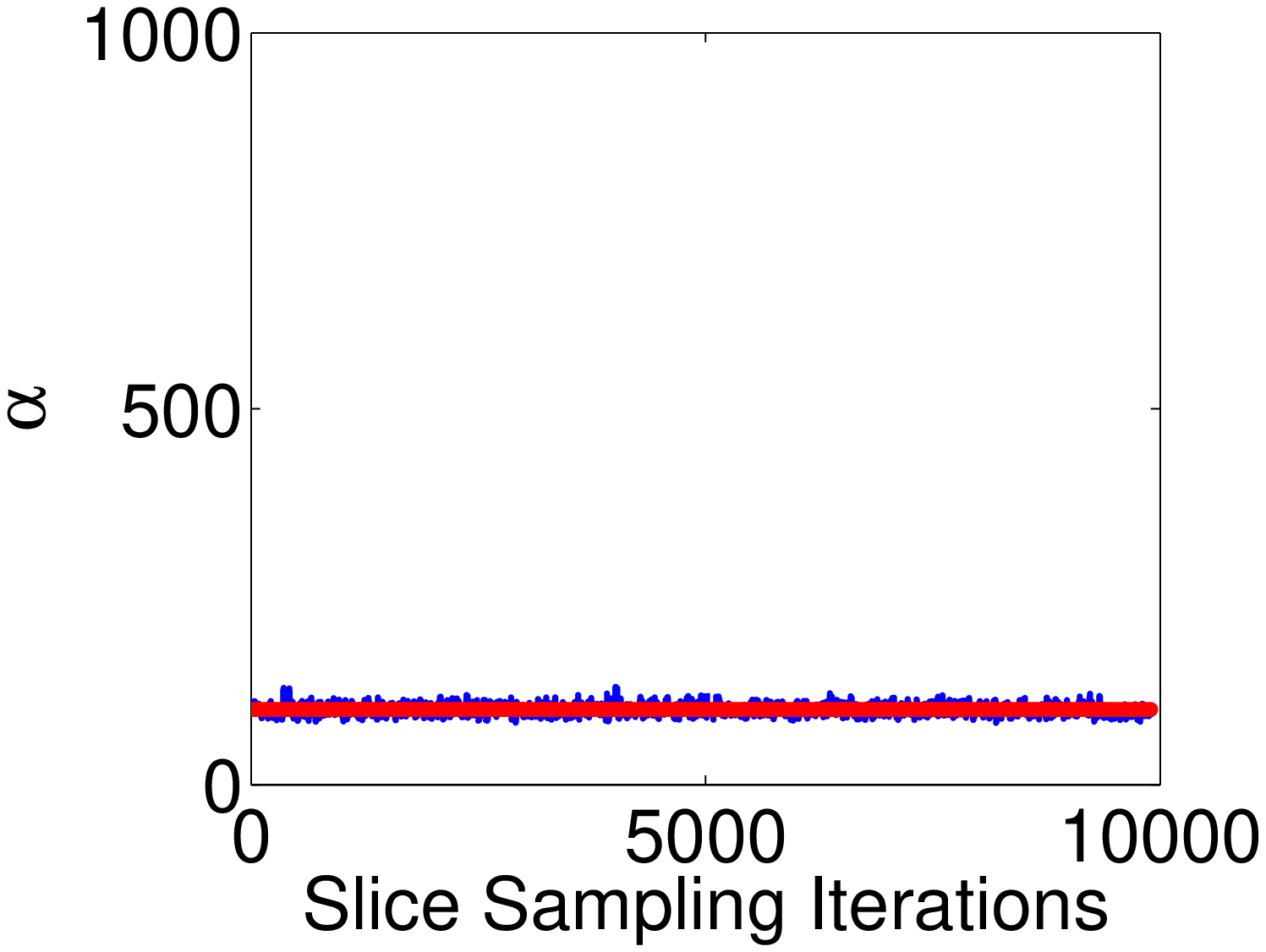}
      \includegraphics[scale=0.2]{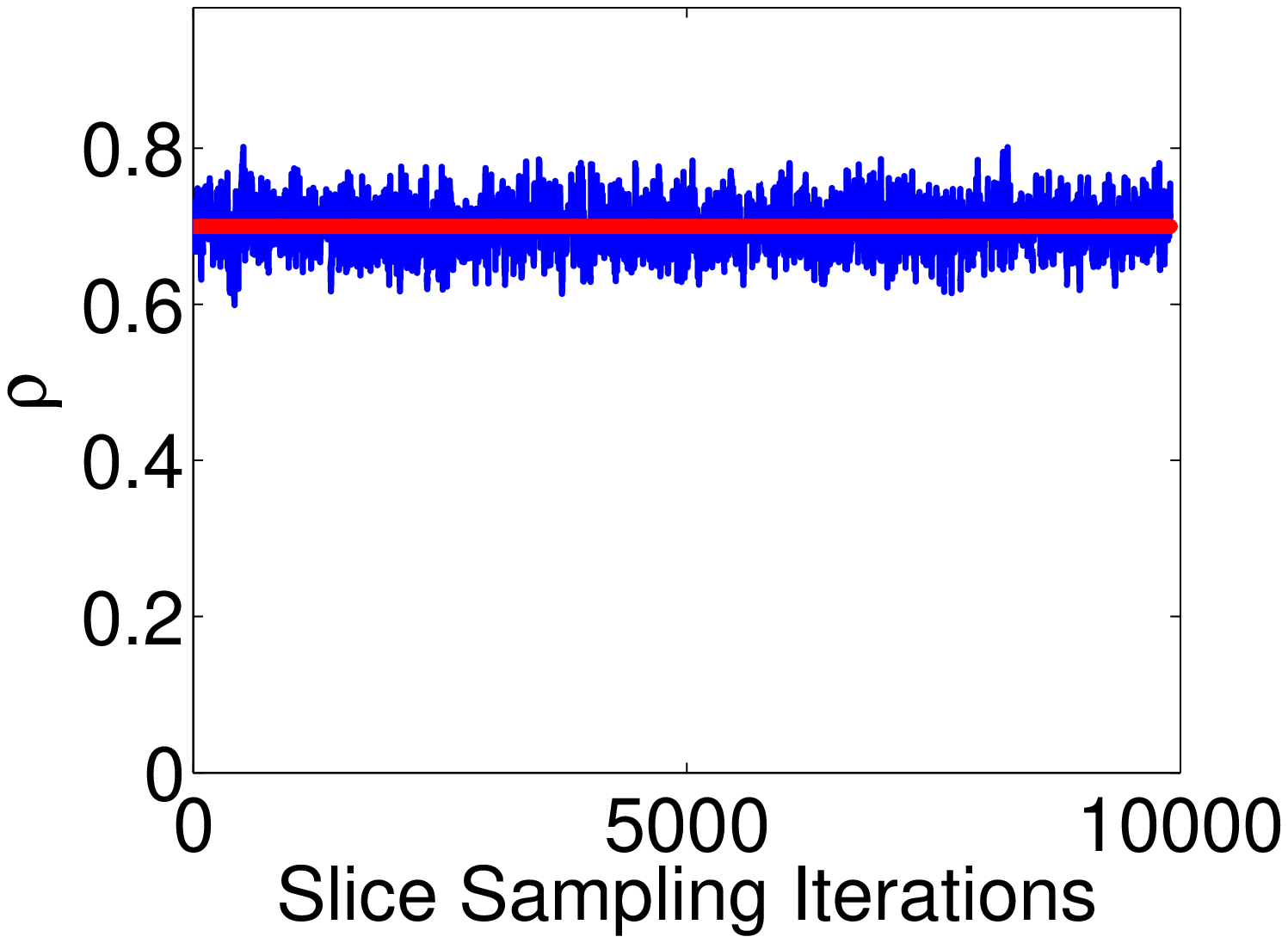}
\includegraphics[scale=0.2]{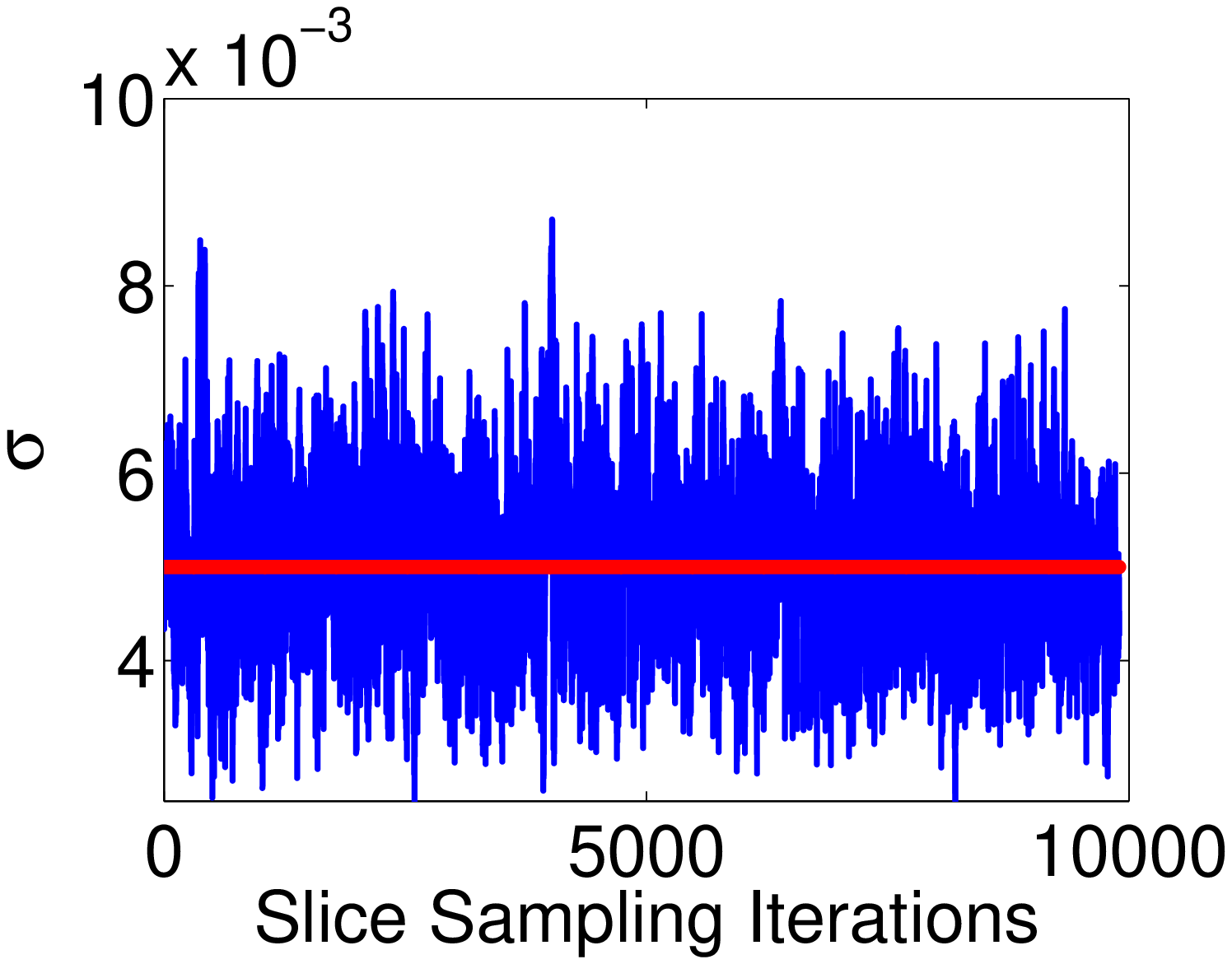}
\includegraphics[scale=0.2]{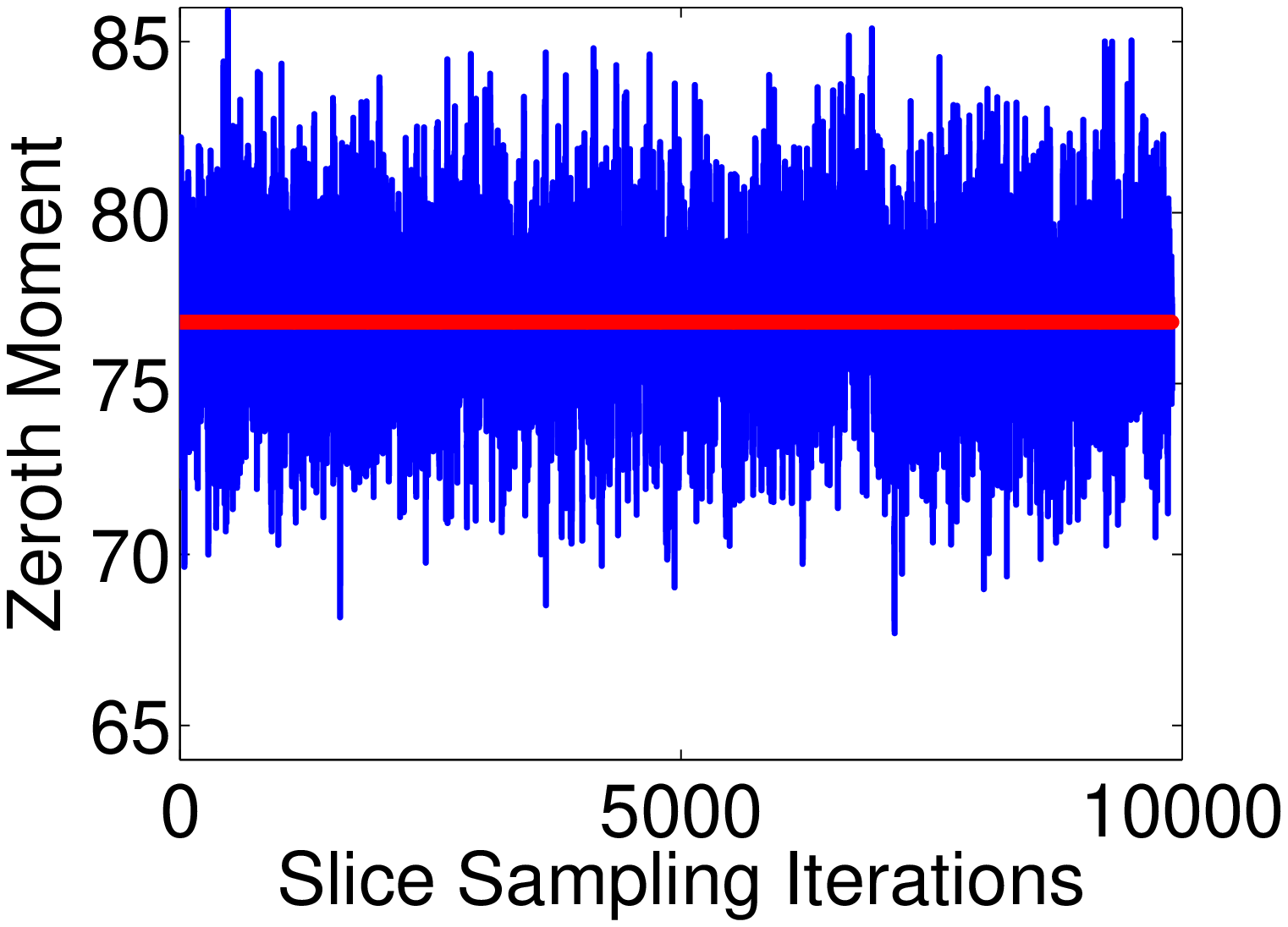}
\includegraphics[scale=0.2]{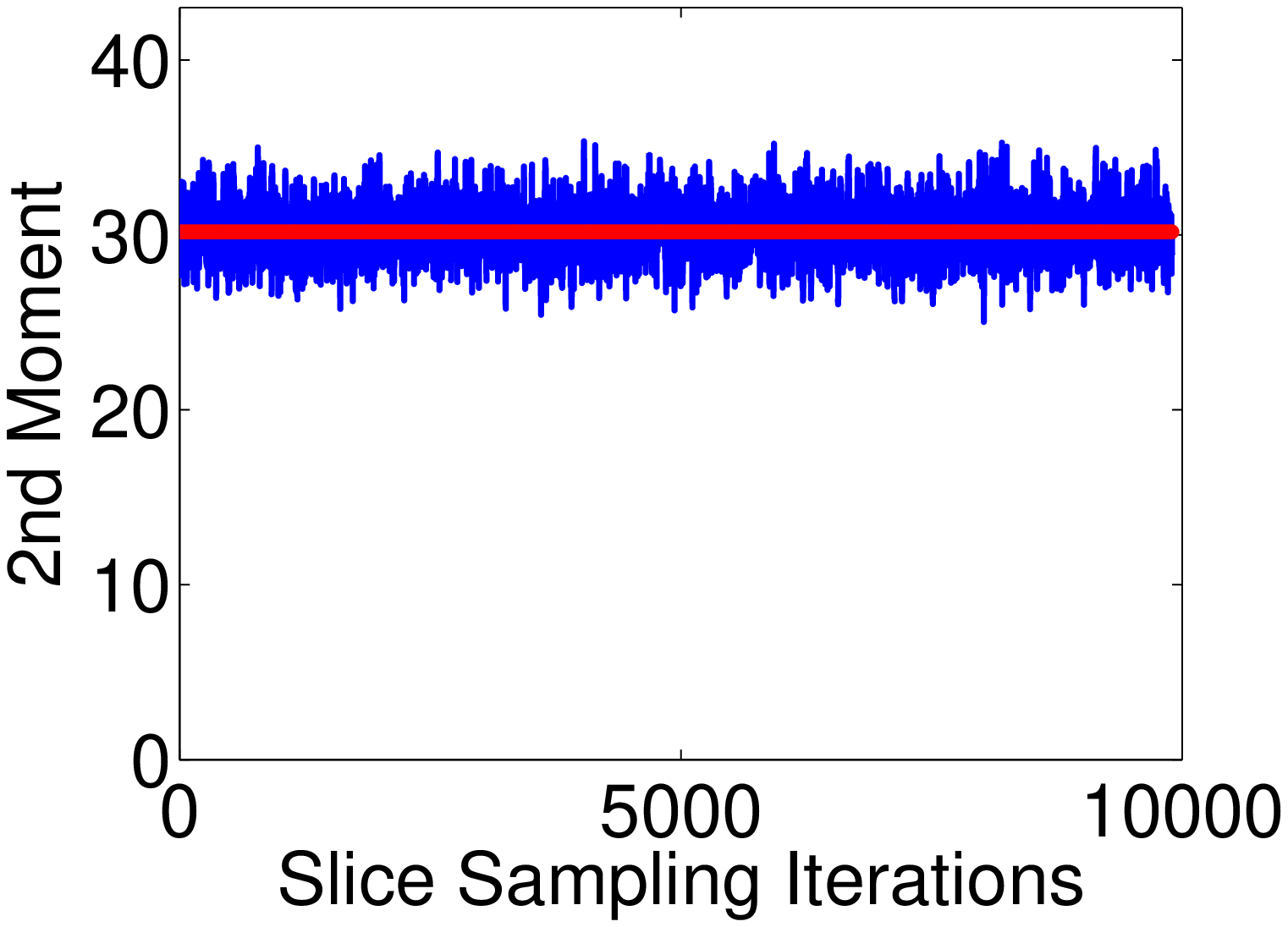}
\vspace{-7pt}
 \caption{Posterior samples and moment estimates under the parameters learned for a continuous DPP with Gaussian quality and similarity. The first 3 columns show posterior samples for (from right to left) $\alpha$, $\rho$ and $\sigma$. The last two columns show the zeroth and second moment estimates. The top row are samples from Scenario (i) (blue) and Scenario (ii) (green) while the second row are samples from Scenario (iii), plotted with the same relative scales to the other scenarios. Red lines indicate the true parameter values that generated the data and their associated theoretical moments.}
  \label{fig:Synth}
\vspace{-10pt}
\end{figure*}

Fig.~\ref{fig:Synth} shows trace plots of the posterior samples for all three scenarios. In the first scenario, the parameter estimates vary wildly whereas in the other two scenarios, the posterior estimates are more stable. In all the cases, the zeroth and second moment estimated from the posterior samples are in the neighborhood of the corresponding empirical moments.  
This leads us to believe that the posterior is broad in cases where 
we have both a small number of samples and few points in each sample.  The posterior becomes more peaked as the total number of points increases. Note that using a stationary similarity kernel allows us to garner information either from few sets with many points or many sets of few points.
\vspace{-10pt}
\paragraph{Dispersion Measure} In many applications, we are interested in quantifying the overdispersion 
of point process data.  In spatial statistics, one standard quantity used to measure dispersion is the Ripley $K$-function~\cite{ripley1977modelling}. Here, instead, we would like to use the learned parameters of the DPP to measure overdispersion as repulsion. An important characteristic of a measure of repulsion is that it should be invariant to scaling. In the Supplementary Material, we derive results that, as the data are scaled from $\bx$ to $\eta\bx$, the parameters scale from $(\alpha,\sigma_i,\rho_i)$ to $(\alpha,\eta\sigma_i,\eta\rho_i)$. This suggests that an appropriate scale-invariant measure of repulsion is~${\gamma_i=\sigma_i/\rho_i}$.

\subsection{Applications}
\subsubsection{Diabetic Neuropathy}
\begin{figure}
\centering
    \includegraphics[scale=0.16]{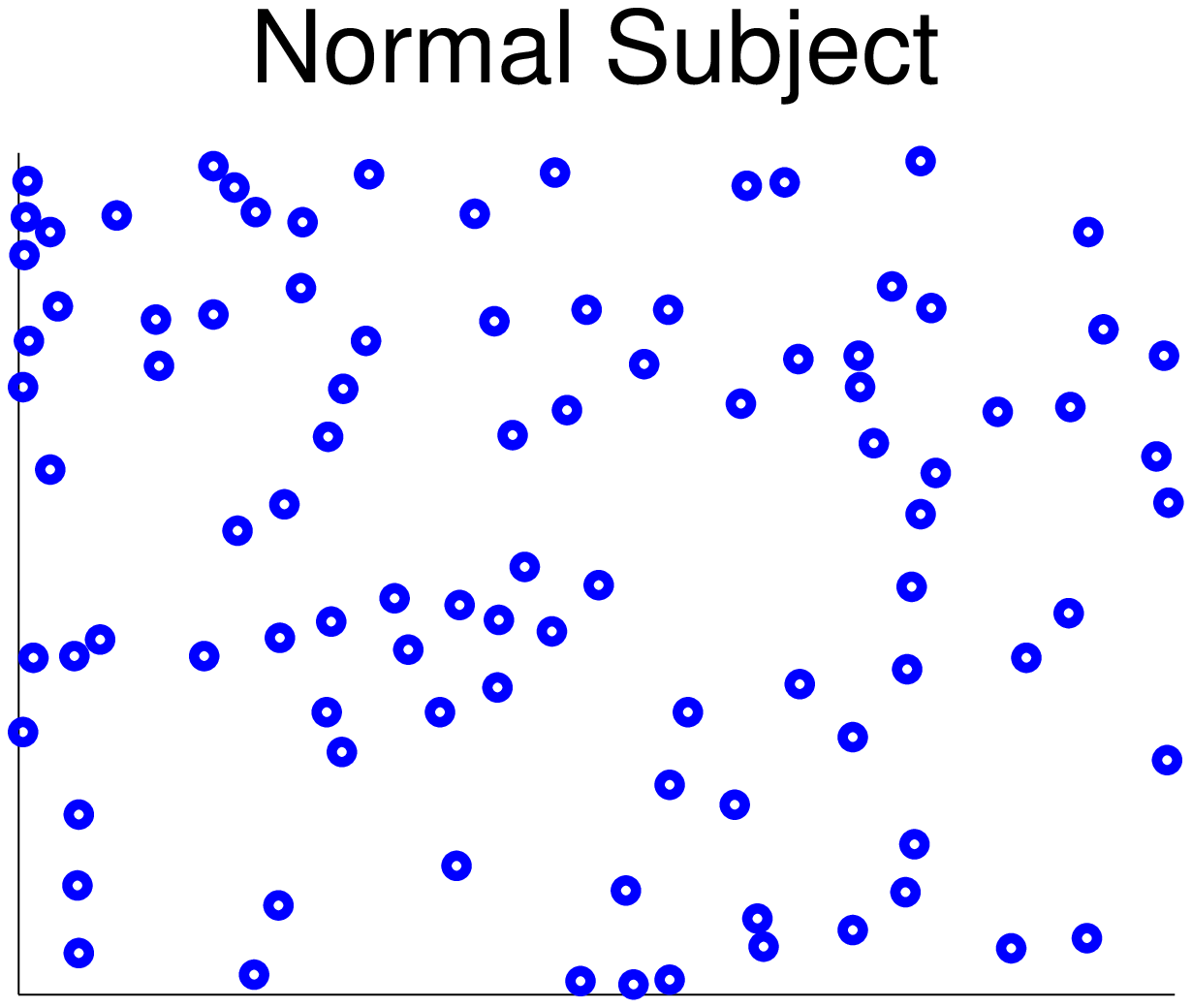} 
    \includegraphics[scale=0.16]{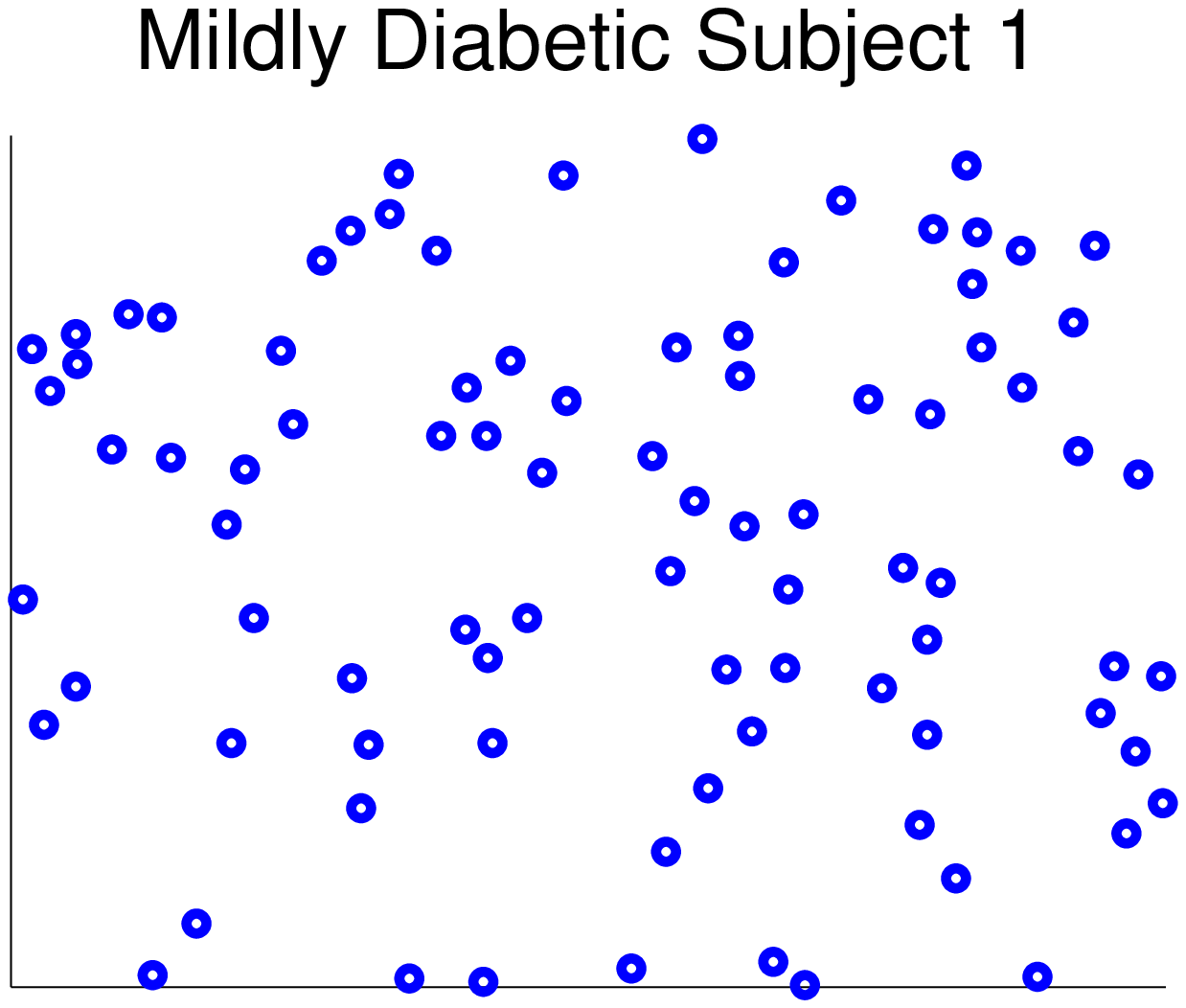} 
    \includegraphics[scale=0.16]{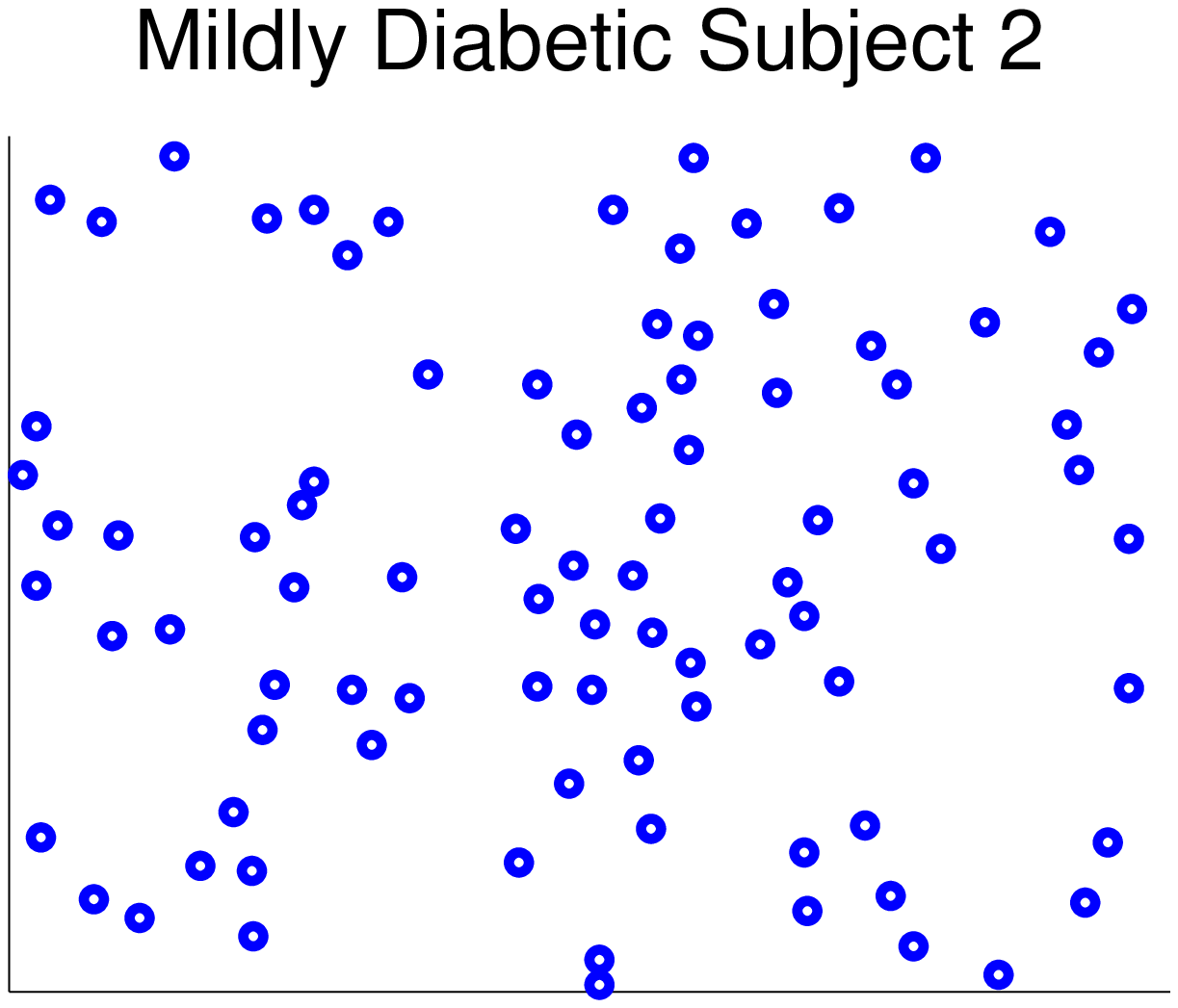}\\
    \includegraphics[scale=0.16]{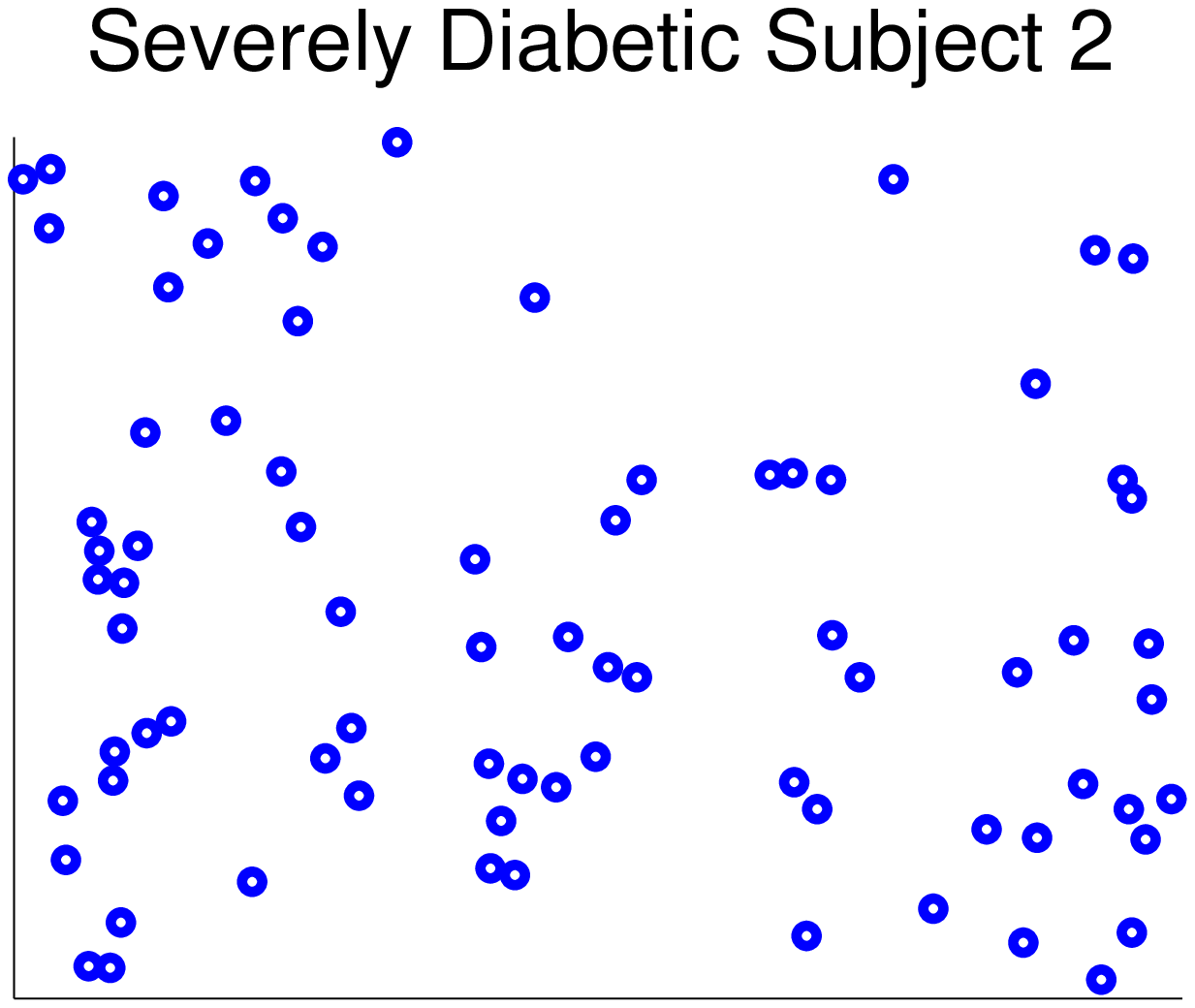} 
    \includegraphics[scale=0.16]{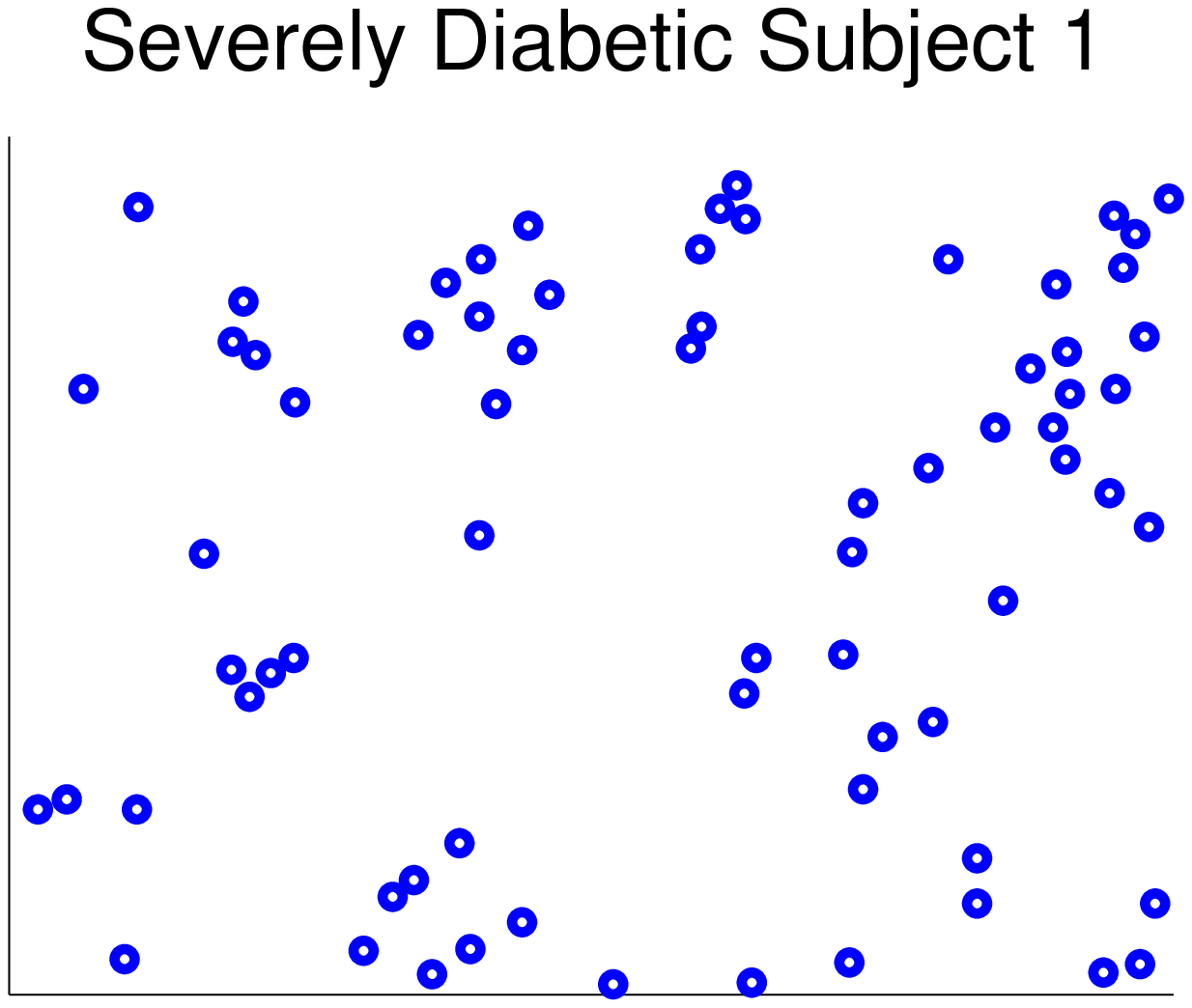} 
    \includegraphics[scale=0.16]{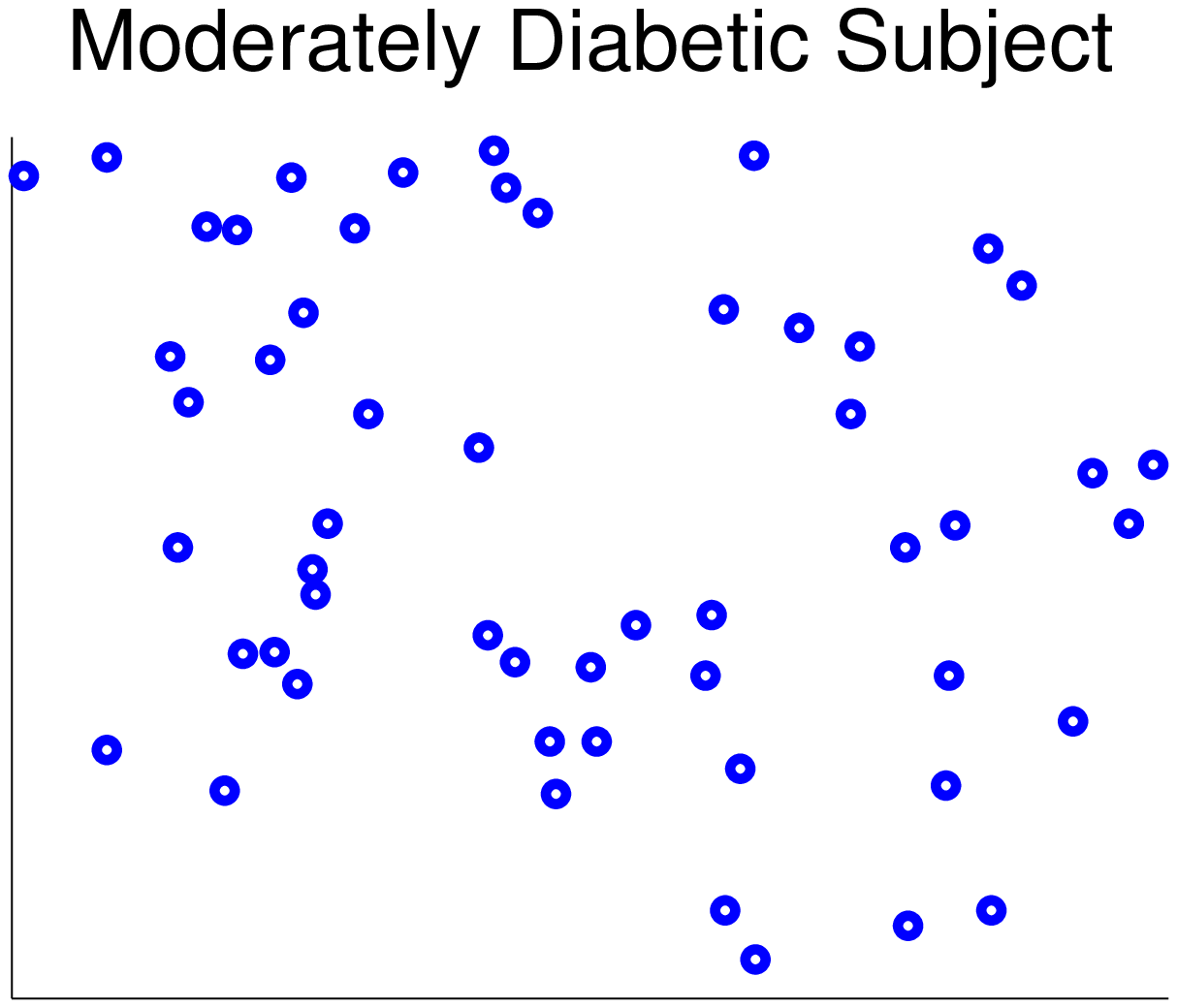}\\
\vspace{-15pt}
  \caption{Nerve fiber samples. Clockwise: (i) Normal subject, (ii) Mildly Diabetic Subject 1, (iii) Mildly Diabetic Subject 2,(iv) Moderately Diabetic subject, (v) Severely Diabetic Subject 1 and (vi) Severely Diabetic Subject 2.}
  \label{fig:nervefibers}
\vspace{-10pt}
\end{figure}

Recent breakthroughs in skin tissue imaging have spurred interest in studying the spatial patterns of nerve fibers in diabetic patients. It has been observed that these nerve fibers appear to become more clustered as diabetes progresses. \citet{waller2011second} previously analyzed this phenomena based on 6 thigh nerve fiber samples. These samples were collected from 5 diabetic patients at different stages of diabetic neuropathy and one healthy subject. On average, there are 79 points in each sample (see Fig.~\ref{fig:nervefibers}). \citet{waller2011second} analyzed the Ripley $K$-function 
and concluded that\comment{while the comparison between healthy and mildly diabetic tissue samples was not statistically significant, the difference between the healthy and moderately diabetic samples is somewhat significant (p-value=0.03) and} the difference between the healthy and severely diabetic samples is highly significant. 

We instead study the differences between these samples by learning the kernel parameters of a DPP and quantifying the level of repulsion of the point process.  Due to the small sample size, we consider a 2-class study of Normal/Mildly Diabetic versus Moderately/Severely Diabetic. We perform two analyses. In the first, we directly quantify the level of repulsion based on our scale-invariant statistic, $\gamma = \sigma/\rho$ (see Sec.~\ref{sec:LearnGaussianCont}).  In the second, we perform a leave-one-out classification by training the parameters on the two classes with one sample left out. We then evaluate the likelihood of the held-out sample under the two learned classes. We repeat this for all six samples. 

We model our data using a 2-dimensional continuous DPP with Gaussian quality and similarity as in Eqs.~\eqref{eq:qGaussian} and \eqref{eq:kGaussian}. Since there is no observed preferred direction in the data, we use an isotropic kernel (${\sigma_d=\sigma}$ and ${\rho_d=\rho}$ for ${d=1,2}$). We place weakly informative inverse gamma priors on $(\alpha,\rho,\sigma)$, as specified in the Supplementary Material, and learn the parameters using slice sampling with eigenvalue bounds as outlined in Sec.~\ref{sec:largescale}.  The results shown in Fig.~\ref{fig:nervefiberclass} indicate that our $\gamma$ measure clearly separates the two classes, concurring with the results of \citet{waller2011second}.  Furthermore, we are able to correctly classify all six samples. While the results are preliminary, being based on only six observations, they show promise for this task. 
\comment{\begin{figure}
\vspace{-5pt}
\centering
    \includegraphics[scale=0.5]{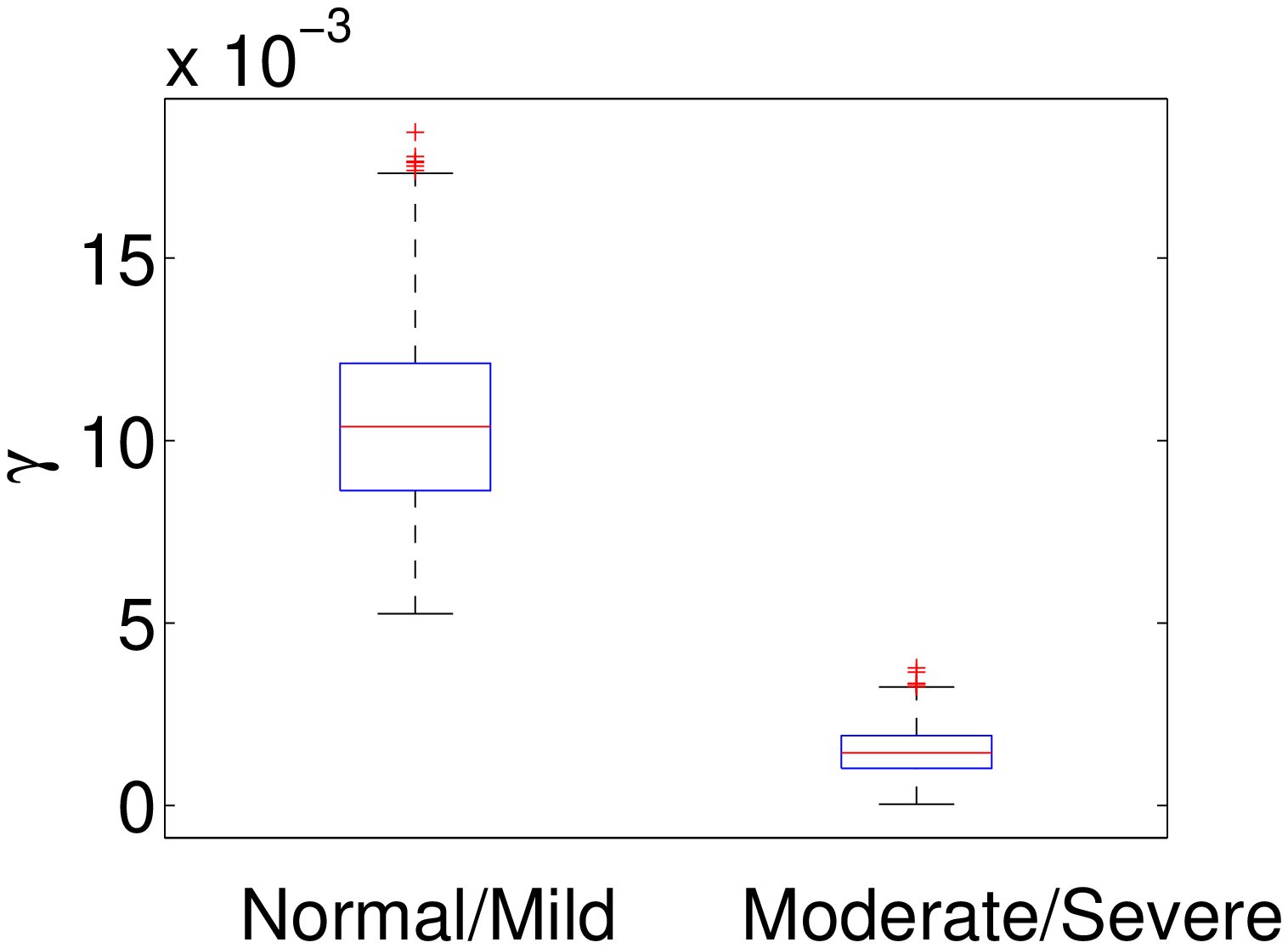}
\vspace{-20pt}
\caption{Learning the scale-invariant repulsive parameter, $\gamma$ for each group of diabetic neuropathy.}
\todo[inline]{Merge with figure on classification using minipage}
\vspace{-30pt}
\end{figure}} 
\begin{figure*}
\begin{minipage}{0.25\linewidth}
    \includegraphics[scale=0.25]{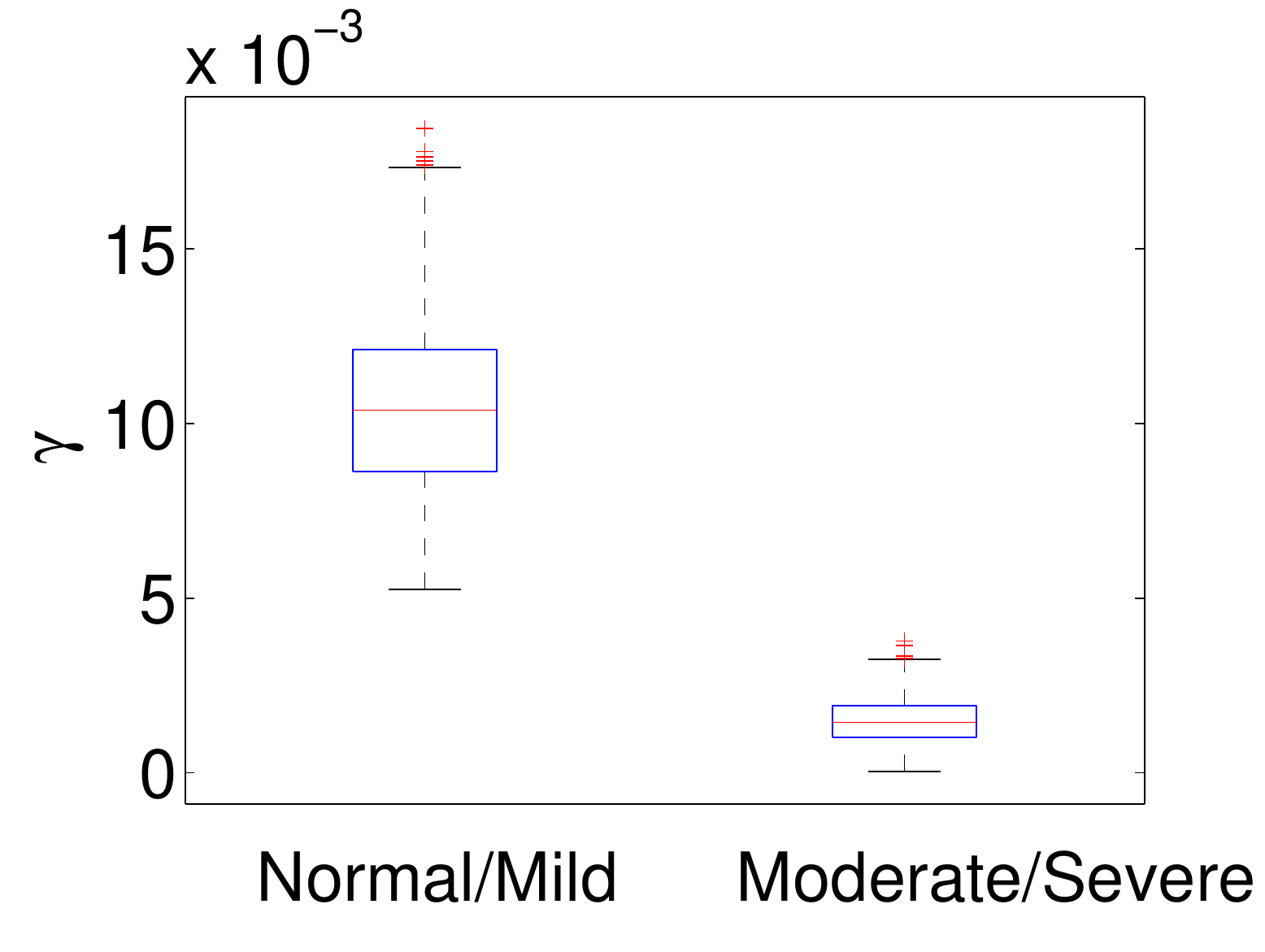}
\vspace{-10pt}
\end{minipage}
\begin{minipage}{0.75\linewidth}
    \includegraphics[scale=0.25]{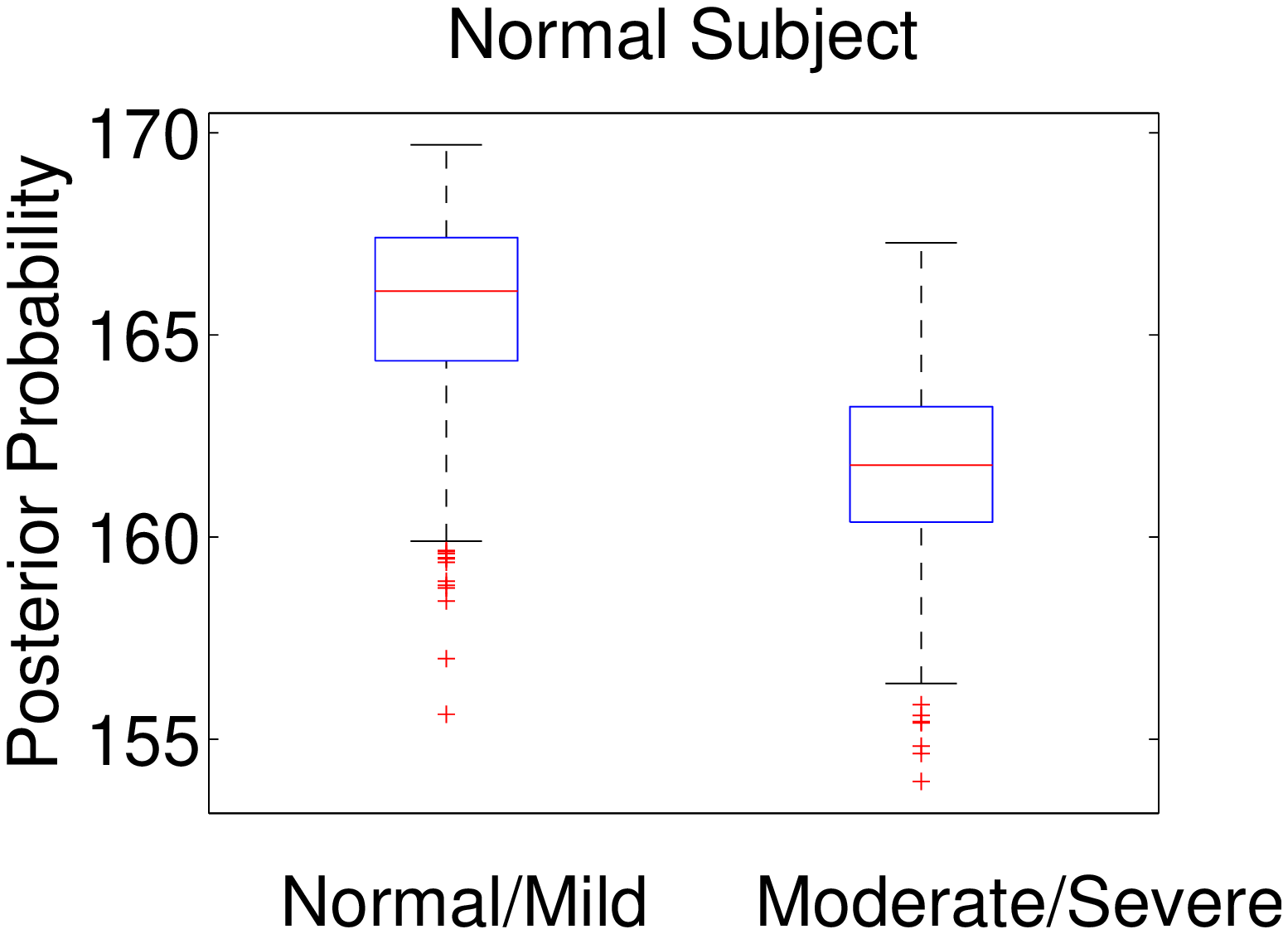} 
    \includegraphics[scale=0.25]{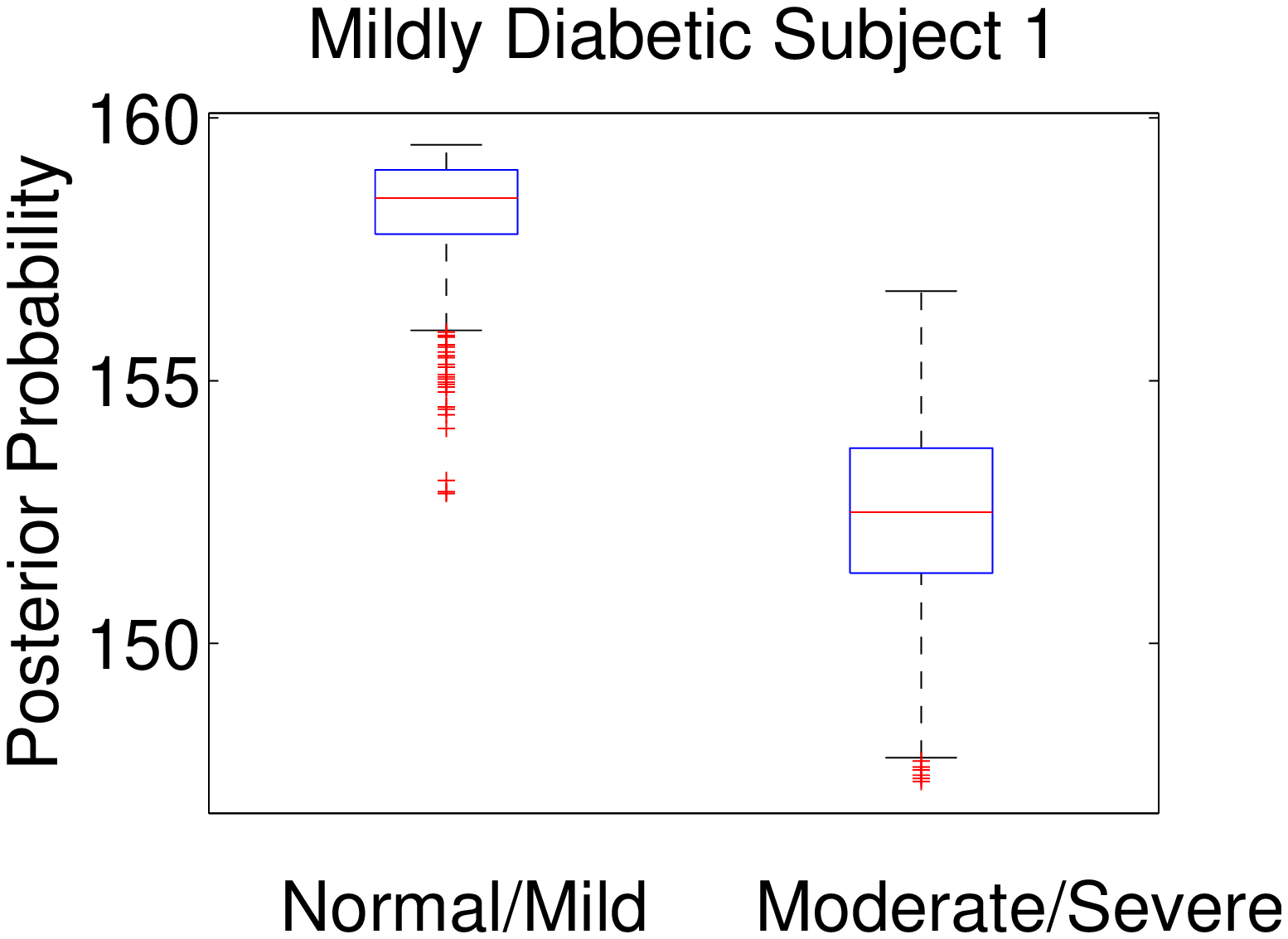} 
    \includegraphics[scale=0.25]{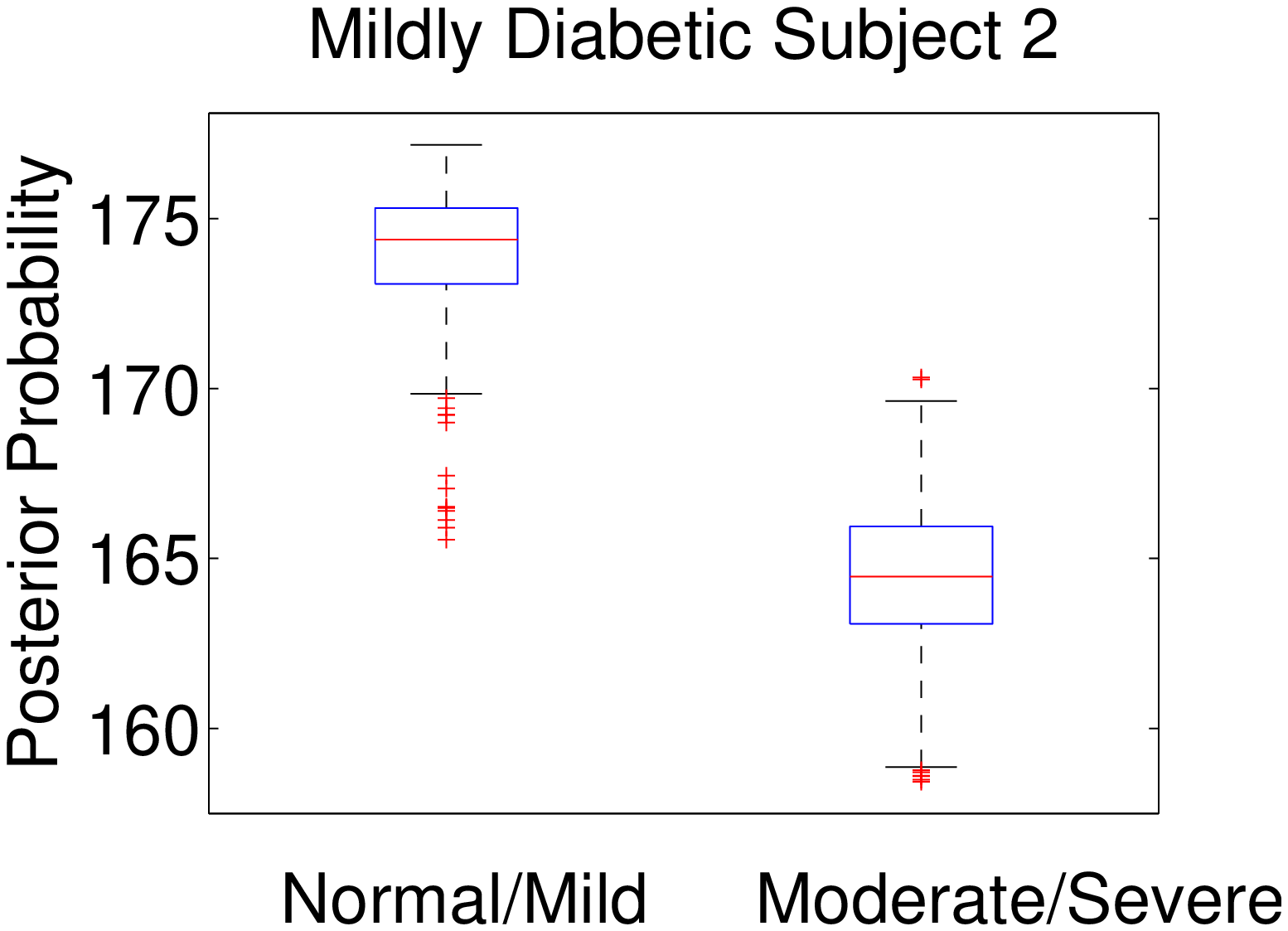}\\
    \includegraphics[scale=0.25]{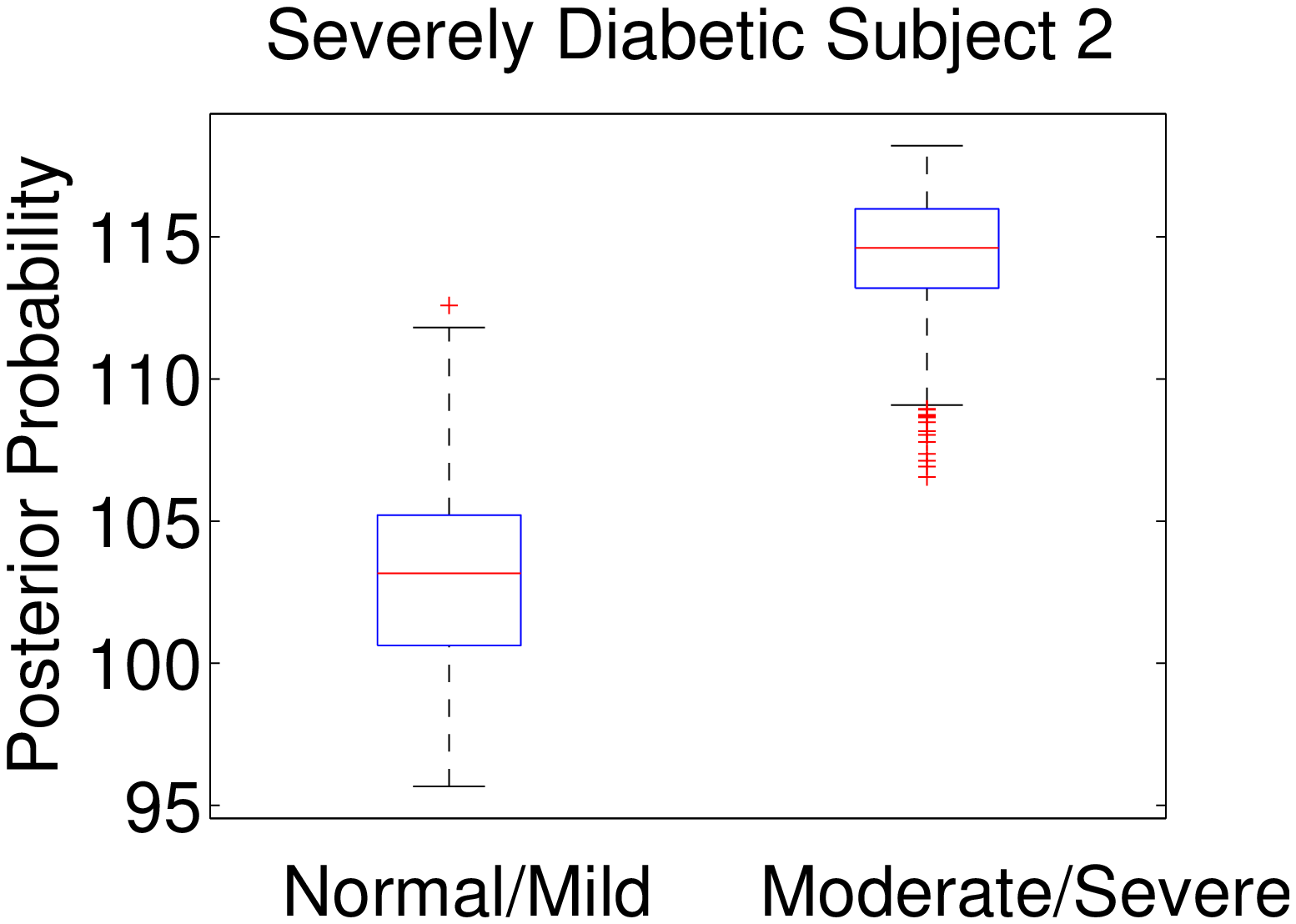} 
    \includegraphics[scale=0.25]{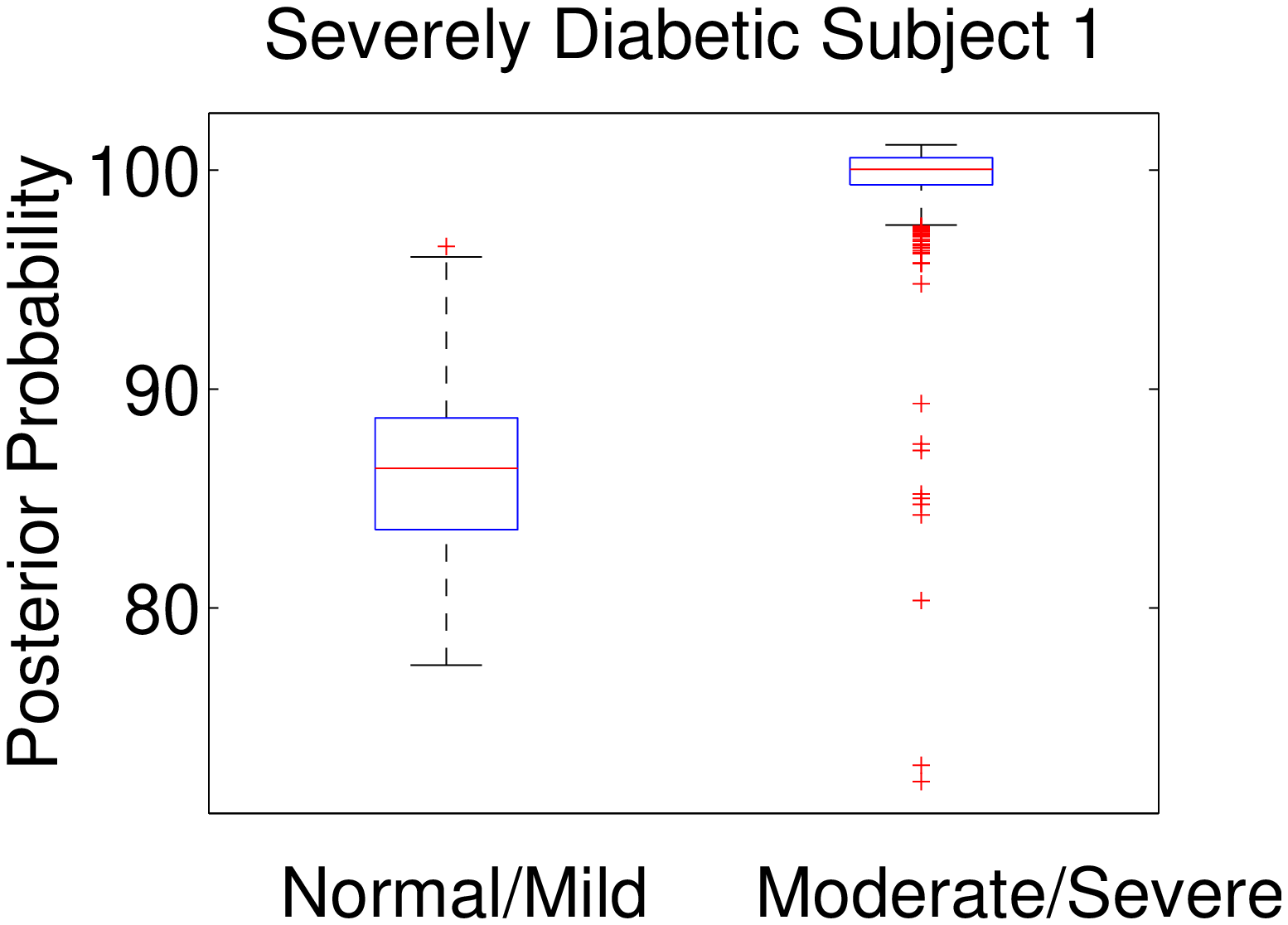} 
    \includegraphics[scale=0.25]{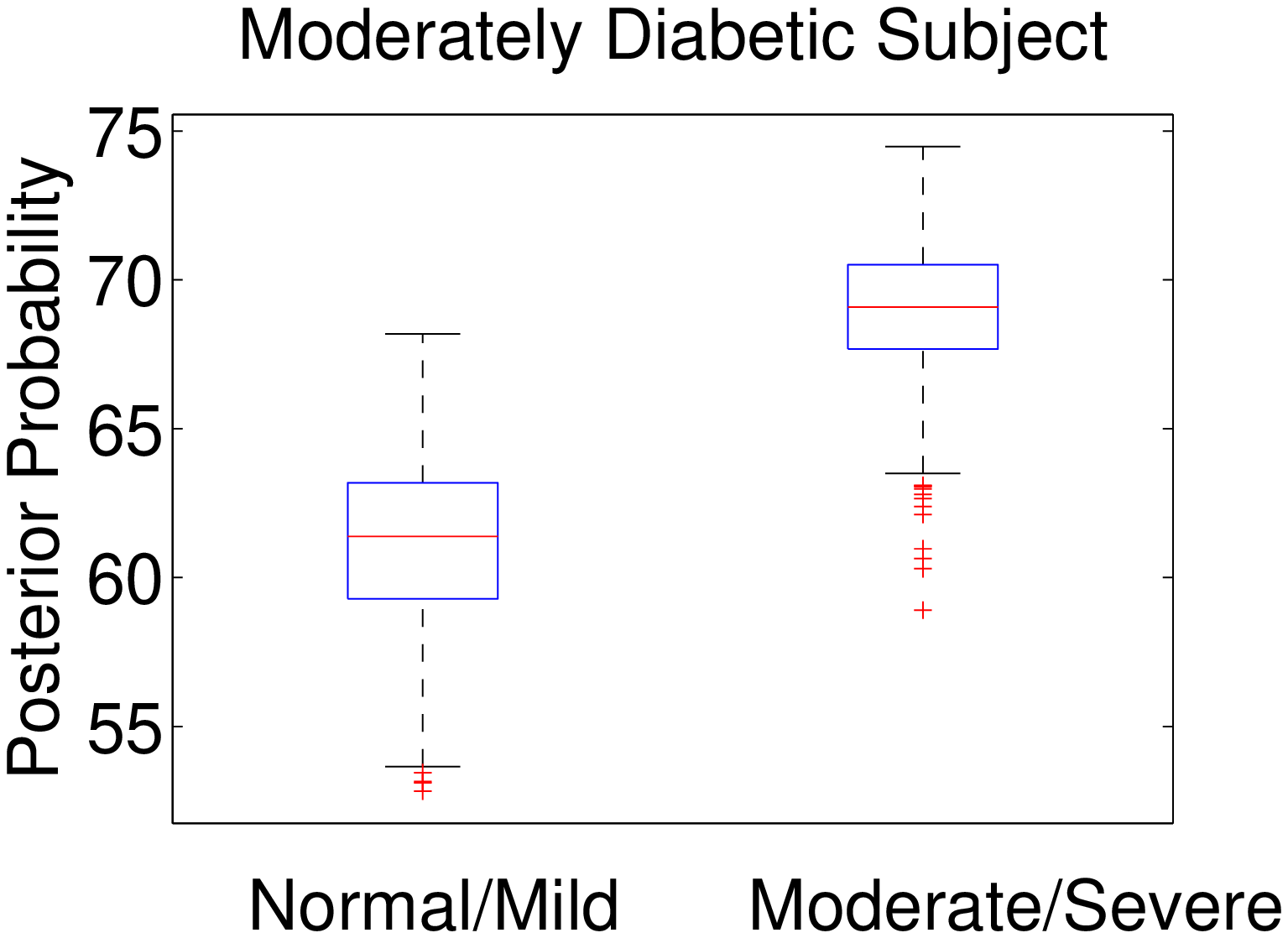}\\
\end{minipage}
\vspace{-25pt}
\caption{Leftmost plot shows the repulsion measure, $\gamma$,  and the rest show the leave-one out log-likelihood of each sample under the two learned DPP classes: Normal/Mildly Diabetic (left box) and Moderately/Severely Diabetic (right box). The ordering of log-likelihood plots aligns with that of Fig.~\ref{fig:nervefibers}.}
  \label{fig:nervefiberclass}
\vspace{-10pt}
\end{figure*}

\vspace{-5pt}
\subsubsection{Diversity in Images}
We also examine DPP learning for quantifying how visual features relate to human perception of diversity in different image categories.  This is useful in applications such as image search, where it is desirable to present users with a set of images that are not only relevant to the query, but diverse as well.

Building on work by \citet{kulesza2011k}, three image categories---$\textsf{cars}$, $\textsf{dogs}$ and $\textsf{cities}$---were studied. Within each category, 8-12 subcategories (such as \emph{Ford} for $\textsf{cars}$, \emph{London} for $\textsf{cities}$ and \emph{poodle} for $\textsf{dogs}$) were queried from Google Image Search and the top 64 results were retrieved. For a subcategory $\textsf{subcat}$, these images form our base set $\Omega_{\sf{subcat}}$. To assess human perception of diversity, human annotated sets of size six were generated from these base sets.  However, it is challenging to ask a human to coherently select six diverse images from a set of 64 total.  Instead, \citet{kulesza2011k} generated a \emph{partial result set} of five images from a 5-DPP on each $\Omega_{\sf{subcat}}$ with a kernel based on the SIFT256 features (see Supplementary Material).  Human annotators (via Amazon Mechanical Turk) were then presented with two images selected at random from the remaining subcategory images and asked to add the image they felt was least similar to the partial result set. These experiments resulted in about 500 samples spread evenly spread evenly across the different subcategories.   

We aim to study how the human annotated sets differ from the top six Google results, $\textsf{Top-6}$. As in \citet{kulesza2011k}, we extracted three types of features from the images---$\textsf{color}$ features, $\textsf{SIFT}$ descriptors \cite{vedaldi2010vlfeat,lowe1999object} and $\textsf{GIST}$ descriptors \cite{oliva2006building} described in the Supplementary Material. We denote these features for image~$i$ as $f_i^{\sf{color}}$, $f_i^{\sf{SIFT}}$, and $f_i^{\sf{GIST}}$, respectively.  For each subcategory, we model our data as a discrete 6-DPP on $\Omega_{\sf{subcat}}$ with kernel
\begin{equation}
L^{\sf{subcat}}_{i,j}=\exp\left\{-\sum_{\sf{feat}}\frac{\|f_i^{\sf{feat}}-f_j^{\sf{feat}}\|_2^2}{\sigma^{\sf{cat}}_{\sf{feat}}}\right\}
\end{equation}
for ${\textsf{feat}\in\{\textsf{color},\textsf{SIFT},\textsf{GIST}\}}$ and $i,j$ indexing the 64 images in $\Omega_{\sf{subcat}}$. Here, we assume that each category has the same parameters across subcategories, namely, ${\sigma_{\textsf{feat}}^{\textsf{cat}}}$ for ${\textsf{subcat} \in \textsf{cat}}$ and ${\textsf{cat}\in\{\textsf{cars},\textsf{dogs},\textsf{cities}\}}$. 

To learn from the $\textsf{Top-6}$ images, we consider the samples as being generated from a 6-DPP.  To emphasize the human component of the $\textsf{5-DPP + human}$ annotation sets, we examine a conditional 6-DPP \cite{kulesza2012determinantal} that fixes the five images from the partial results set and only considers the probability of adding the human annotated image.  The Supplementary Material provides details on this conditional $k$-DPP. \comment{and its modification to learning.}

All subcategory samples within a category are assumed to be independent draws from a DPP defined on $\Omega_{\sf{subcat}}$ with kernel $L^{\sf{subcat}}$ parameterized by a shared set of~$\sigma^{\textsf{cat}}_{\textsf{feat}}$, for $\textsf{subcat} \in \textsf{cat}$.  As such, each of these samples equally informs the posterior of $\sigma_{\textsf{feat}}^{\textsf{cat}}$.  We perform posterior sampling of the 6-DPP or conditional 6-DPP kernel parameters using slice sampling with weakly informative inverse gamma priors on the $\sigma^{\textsf{cat}}_{\textsf{feat}}$.  Details are in the Supplementary Material.

Fig.~\ref{fig:imagesigma} shows a comparison between $\sigma^{\textsf{cat}}_{\textsf{feat}}$ learned from the human annotated samples (conditioning on the 5-DPP partial result sets) and the $\textsf{Top-6}$ samples for different categories.  The results indicate that the $\textsf{5-DPP + human}$ annotated samples differs significantly from the $\textsf{Top-6}$ samples in the features judged by human to be important in diversity in each category. For $\textsf{cars}$ and $\textsf{dogs}$, human annotators deem $\textsf{color}$ to be a more important feature for diversity than the Google search engine based on their $\textsf{Top-6}$ results. For $\textsf{cities}$, on the other hand, the $\textsf{SIFT}$ features are deemed important for diversity by human annotators, while the Google search engine puts a much lower weight on them. Keep in mind, though, that this result only highlights the diversity components of the results while ignoring quality. In real life applications, it is desirable to combine both the quality of each image (as a measure of relevance of the image to the query) and the diversity between the top results.  Regardless, we have shown that DPP kernel learning can be informative of judgements of diversity, and this information could be used (for example) to tune search engines to provide results more in accordance with human judgement.  

\comment{\begin{figure}
\centering
    \includegraphics[scale=0.16]{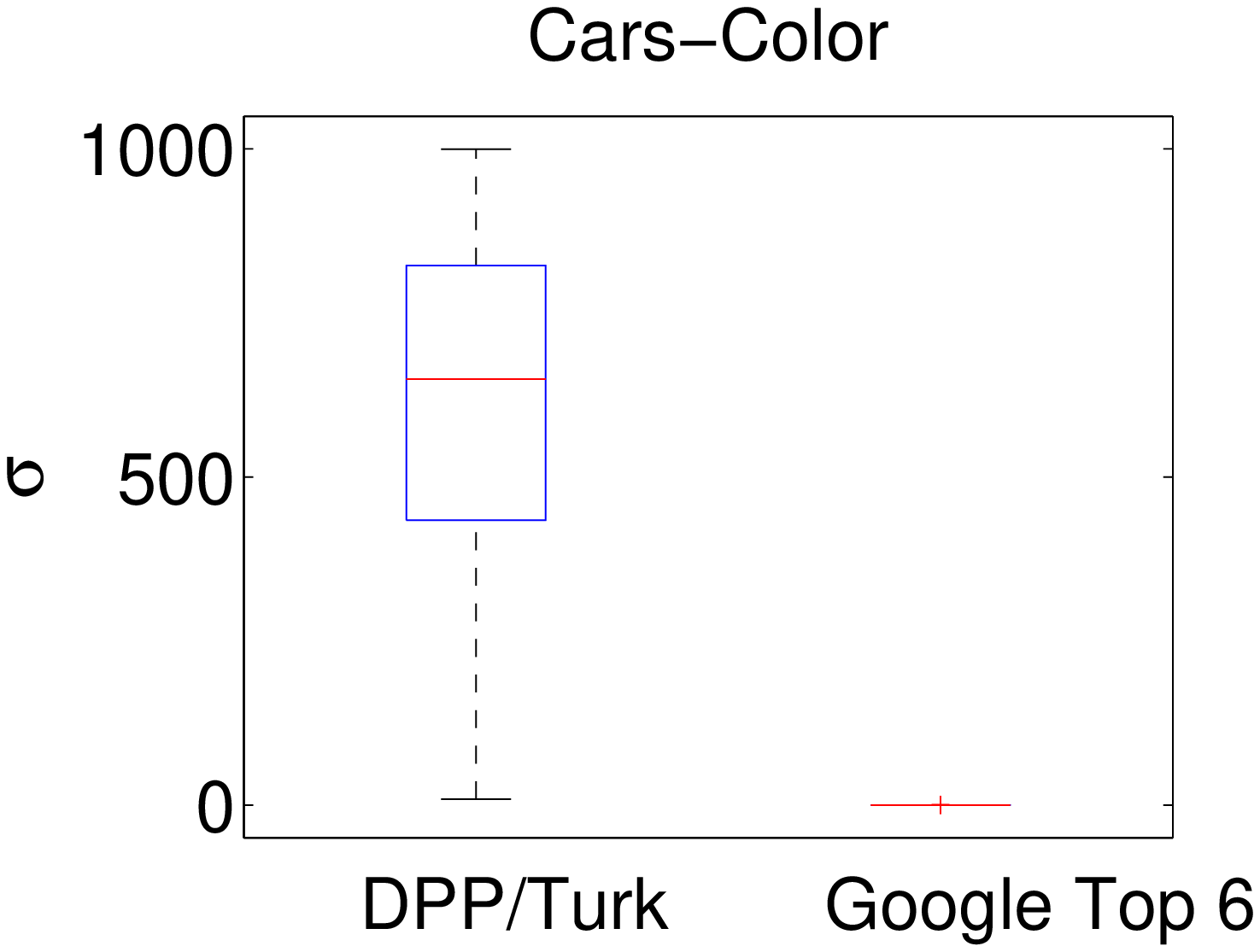} 
    \includegraphics[scale=0.16]{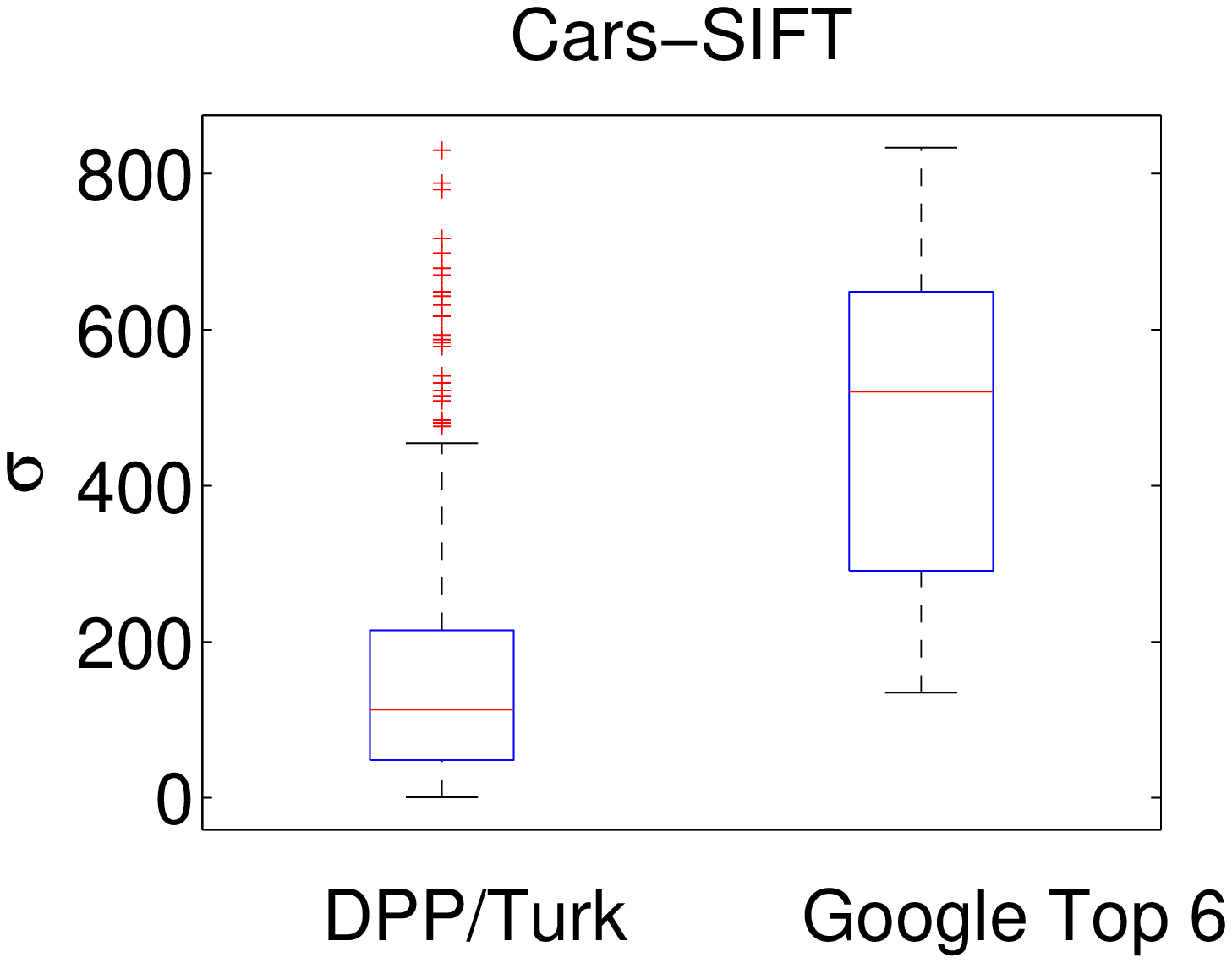} 
    \includegraphics[scale=0.16]{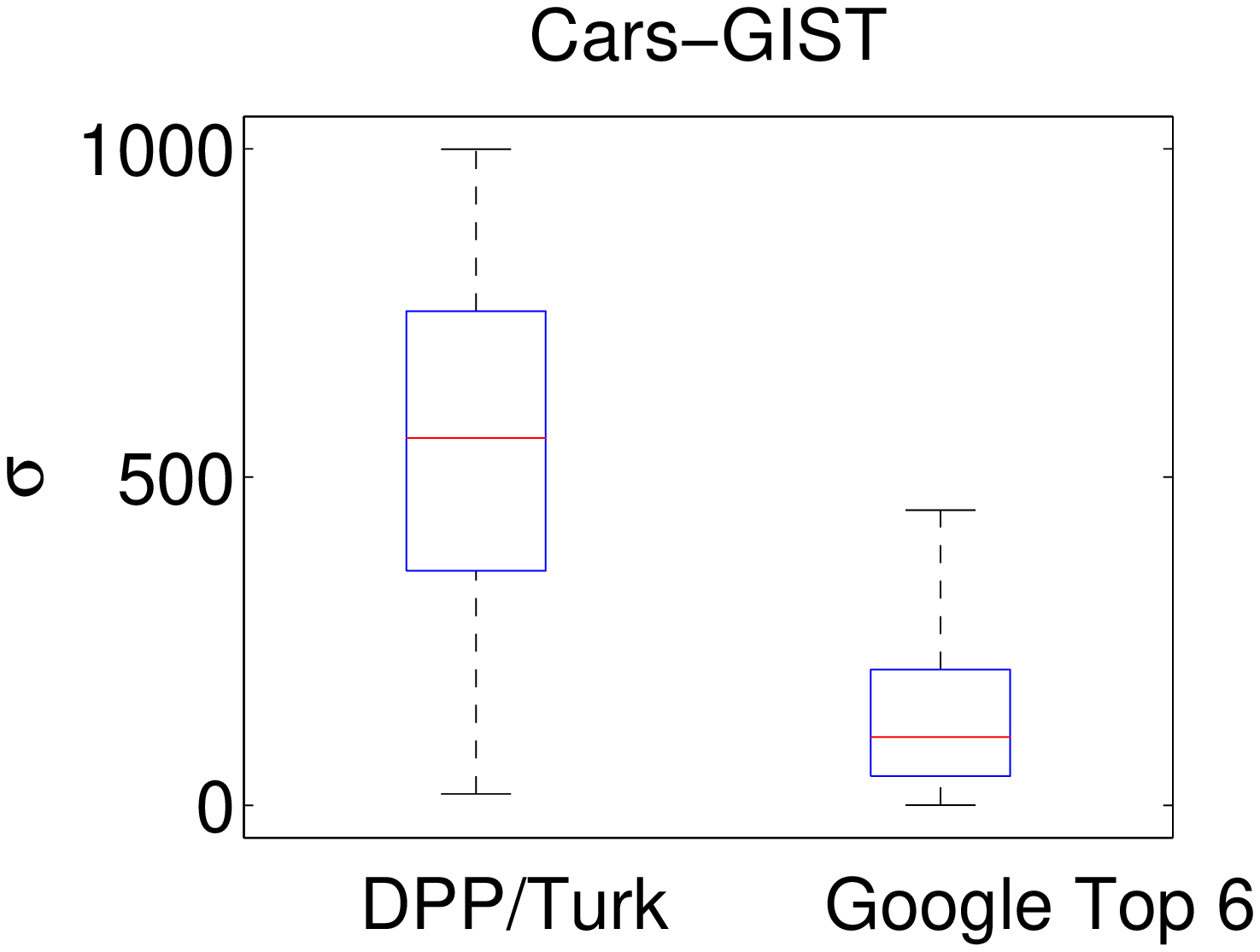}\\
    \includegraphics[scale=0.16]{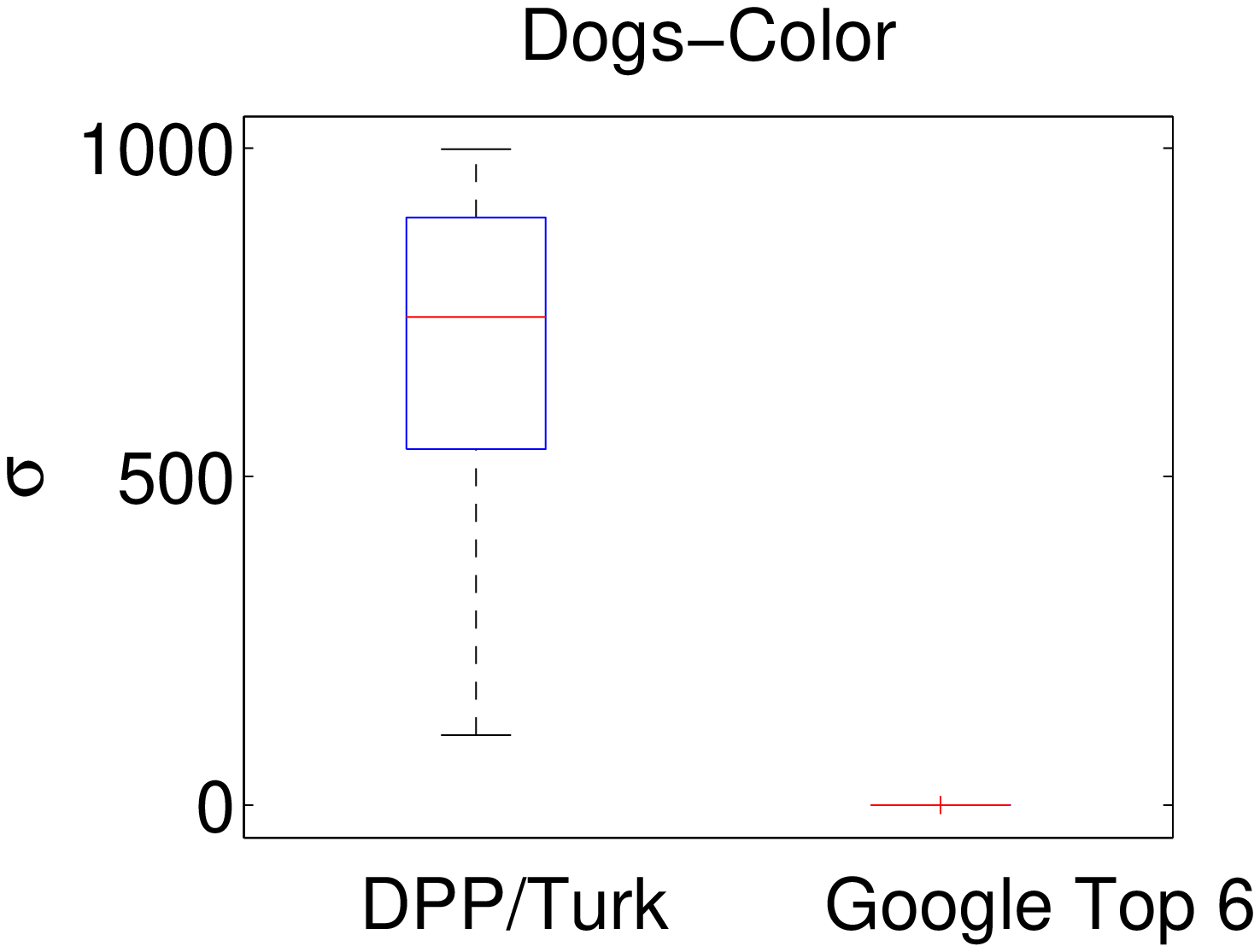} 
    \includegraphics[scale=0.16]{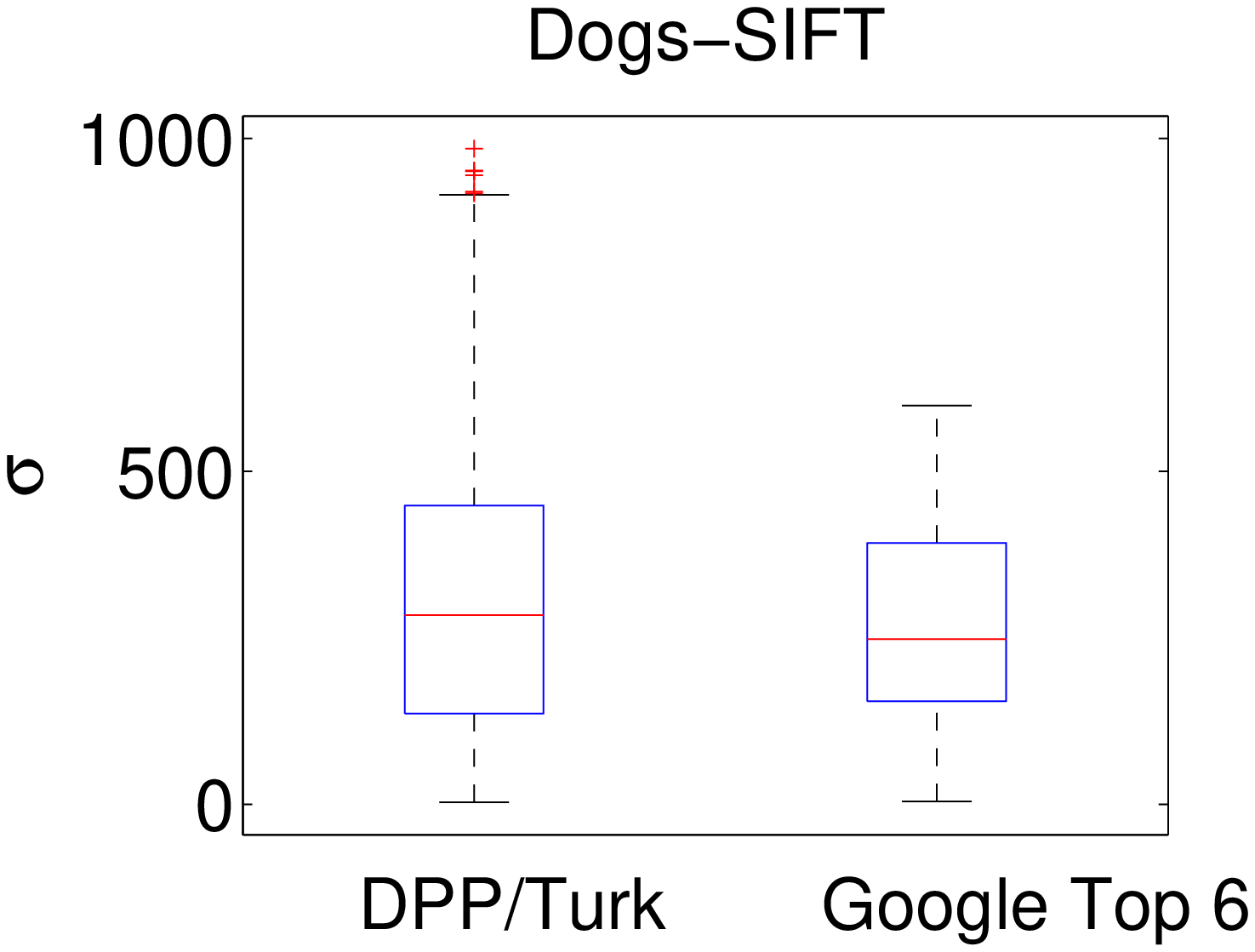} 
    \includegraphics[scale=0.16]{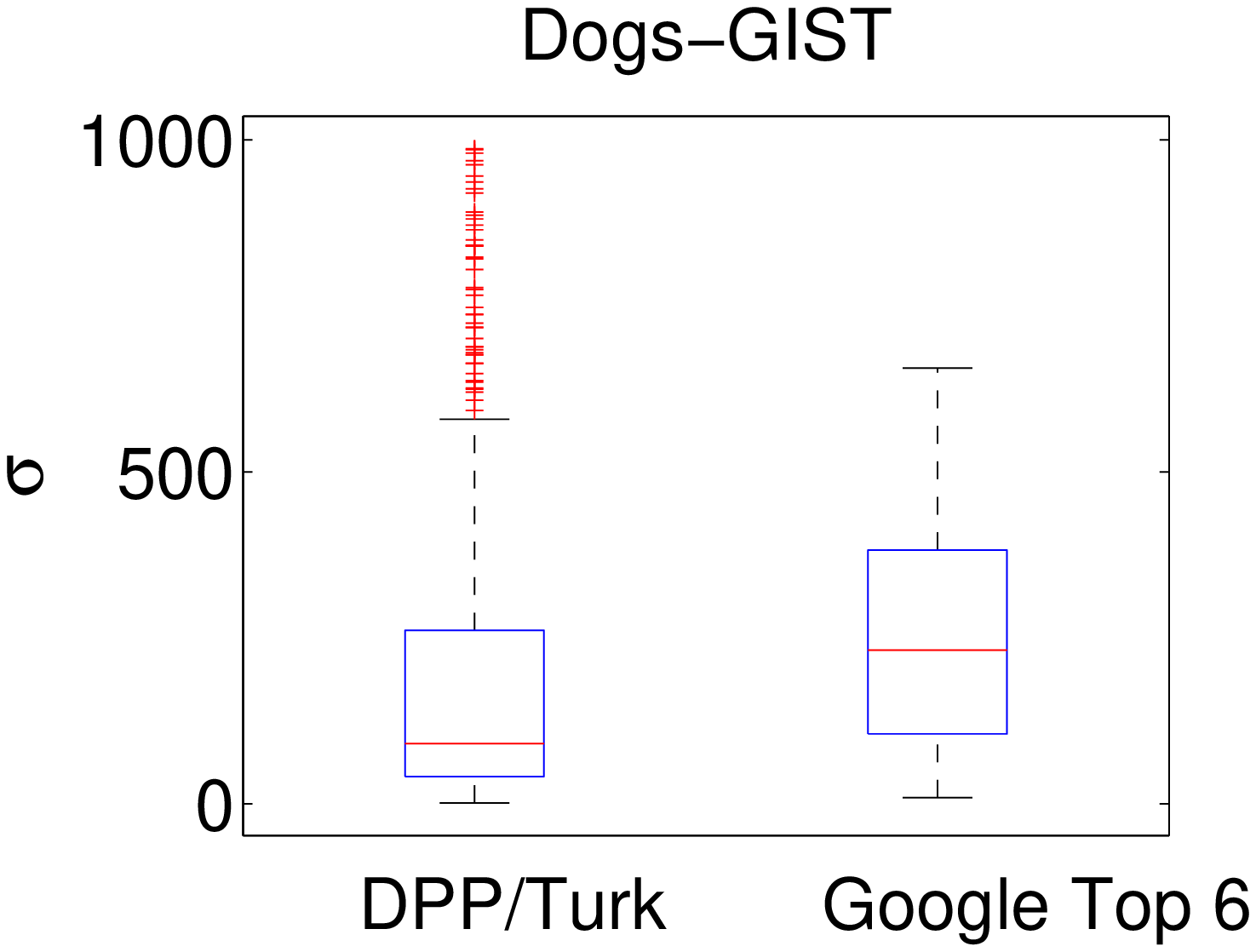}\\        
	\includegraphics[scale=0.16]{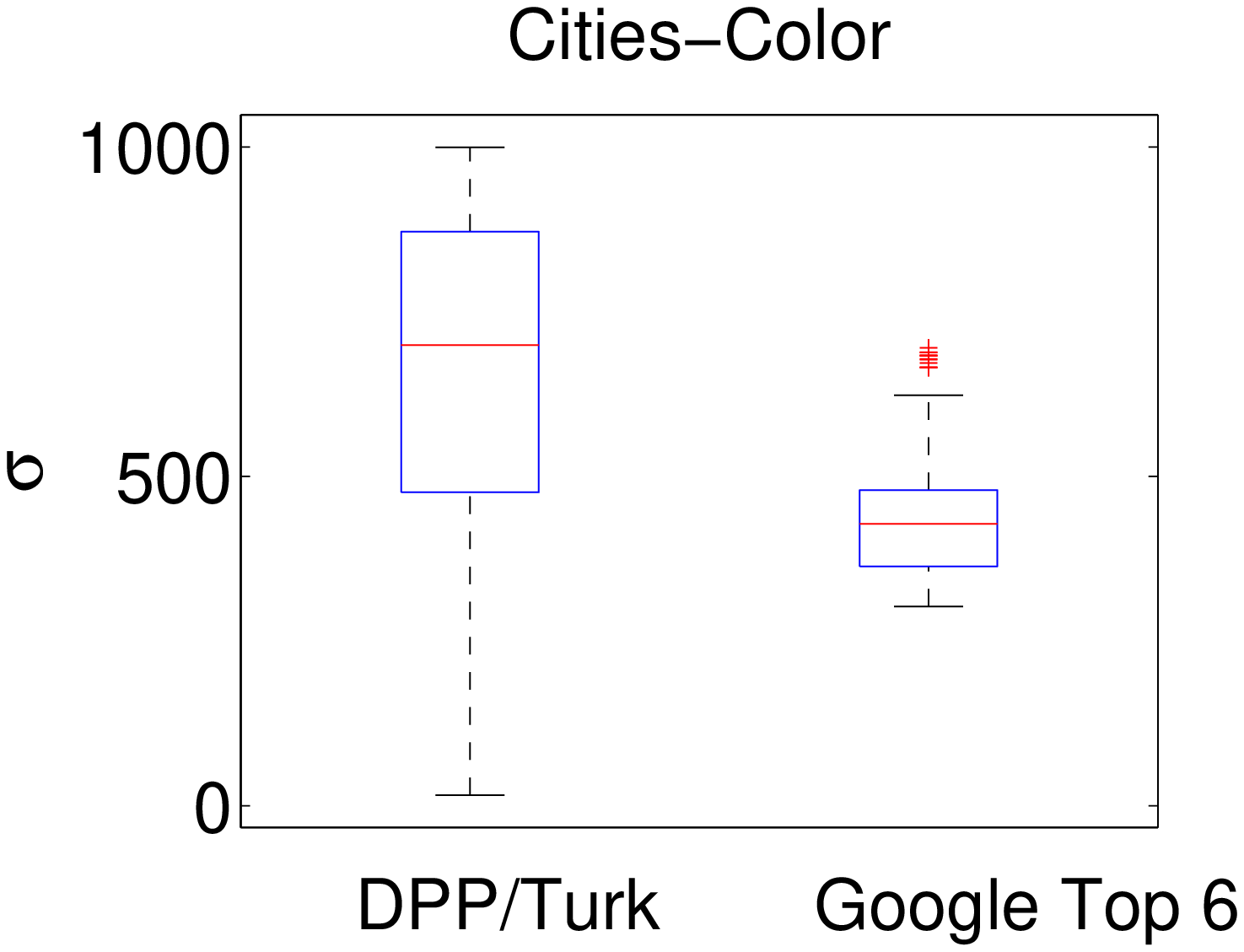} 
    \includegraphics[scale=0.16]{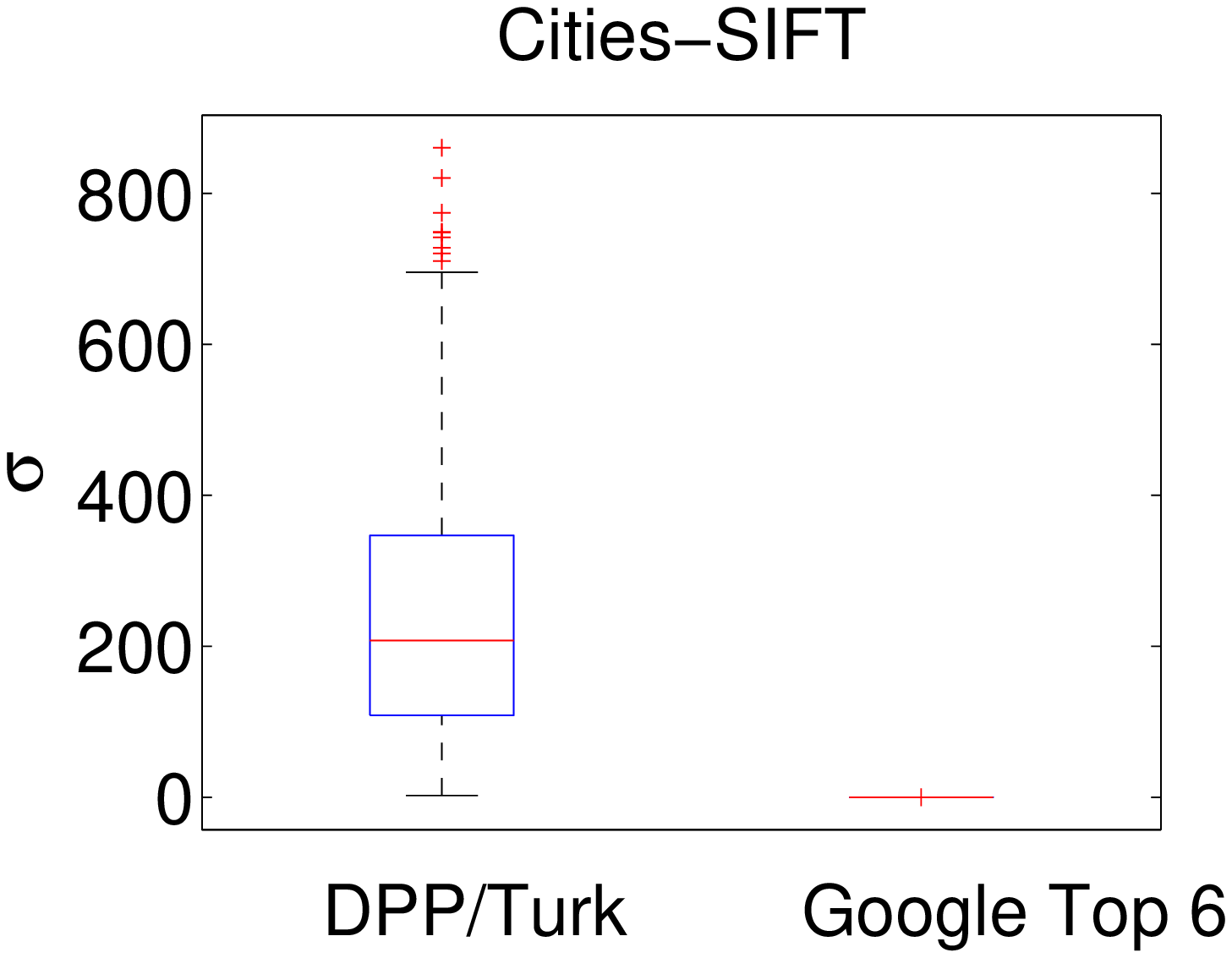} 
    \includegraphics[scale=0.16]{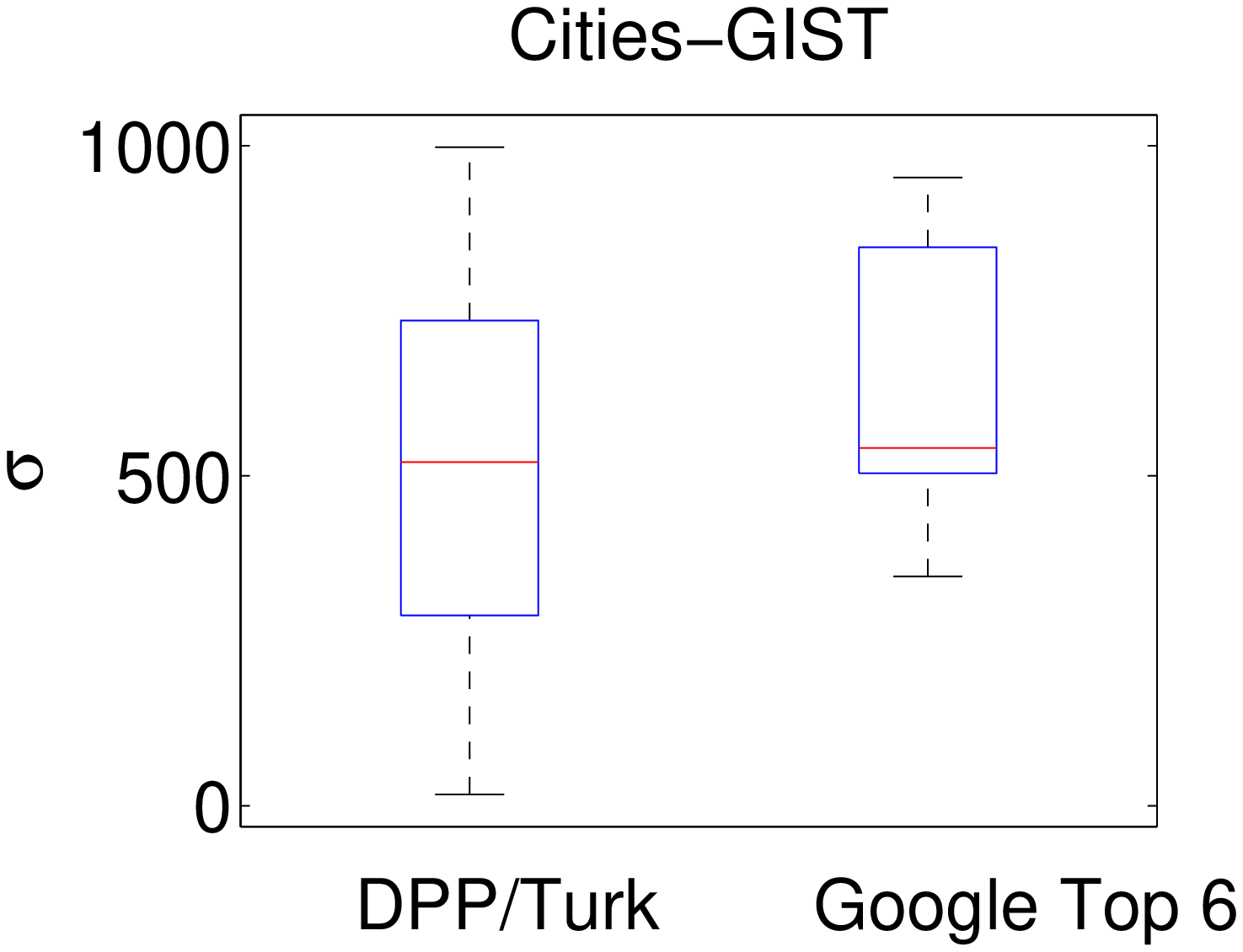}\\
\vspace{-10pt}
  \caption{For the image diversity experiment, boxplots of posterior samples of (rom left to right) $\sigma^{\textsf{cat}}_{\textsf{color}}$, $\sigma^{\textsf{cat}}_{\textsf{SIFT}}$ and $\sigma^{\textsf{cat}}_{\textsf{GIST}}$. Each plot shows results for human annotated sets (left) versus Google Top 6 (right). Categories from top to bottom: (a)\textsf{cars}, (b)\textsf{dogs} and (c)\textsf{cities}}
\vspace{-10pt}
  \label{fig:imagesigma}
\end{figure}}

\section{Conclusion}
\label{sec:conclusion}
\vspace{-5pt}
Determinantal point processes have become increasingly popular in machine learning and statistics. While many important DPP computations are efficient,\comment{---such as sampling and probability calculations---} learning the parameters of a DPP kernel is difficult due to the non-convexity of the likelihood.  We proposed Bayesian approaches using MCMC, in particular, for inferring these parameters.  In addition to being more robust and providing a characterization of the posterior uncertainty, these algorithms can be modified to deal with large-scale and continuous DPPs. We also showed how our posterior samples can be evaluated using moment matching as a model-checking method. Finally we demonstrated the utility of learning DPP parameters in studying diabetic neuropathy and evaluating human perception of diversity in images.
\begin{figure}
\centering
    \includegraphics[scale=0.16]{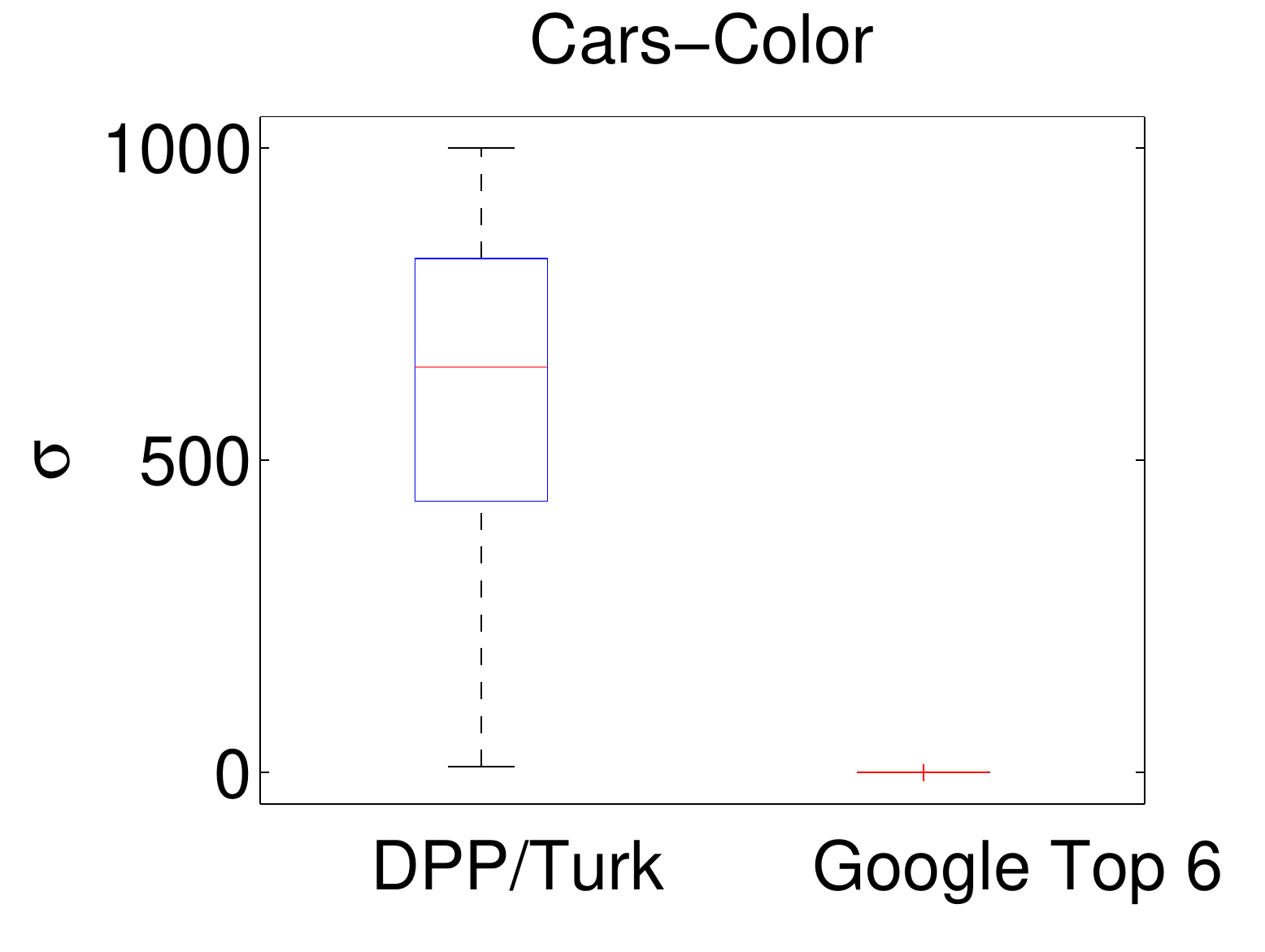} 
    \includegraphics[scale=0.16]{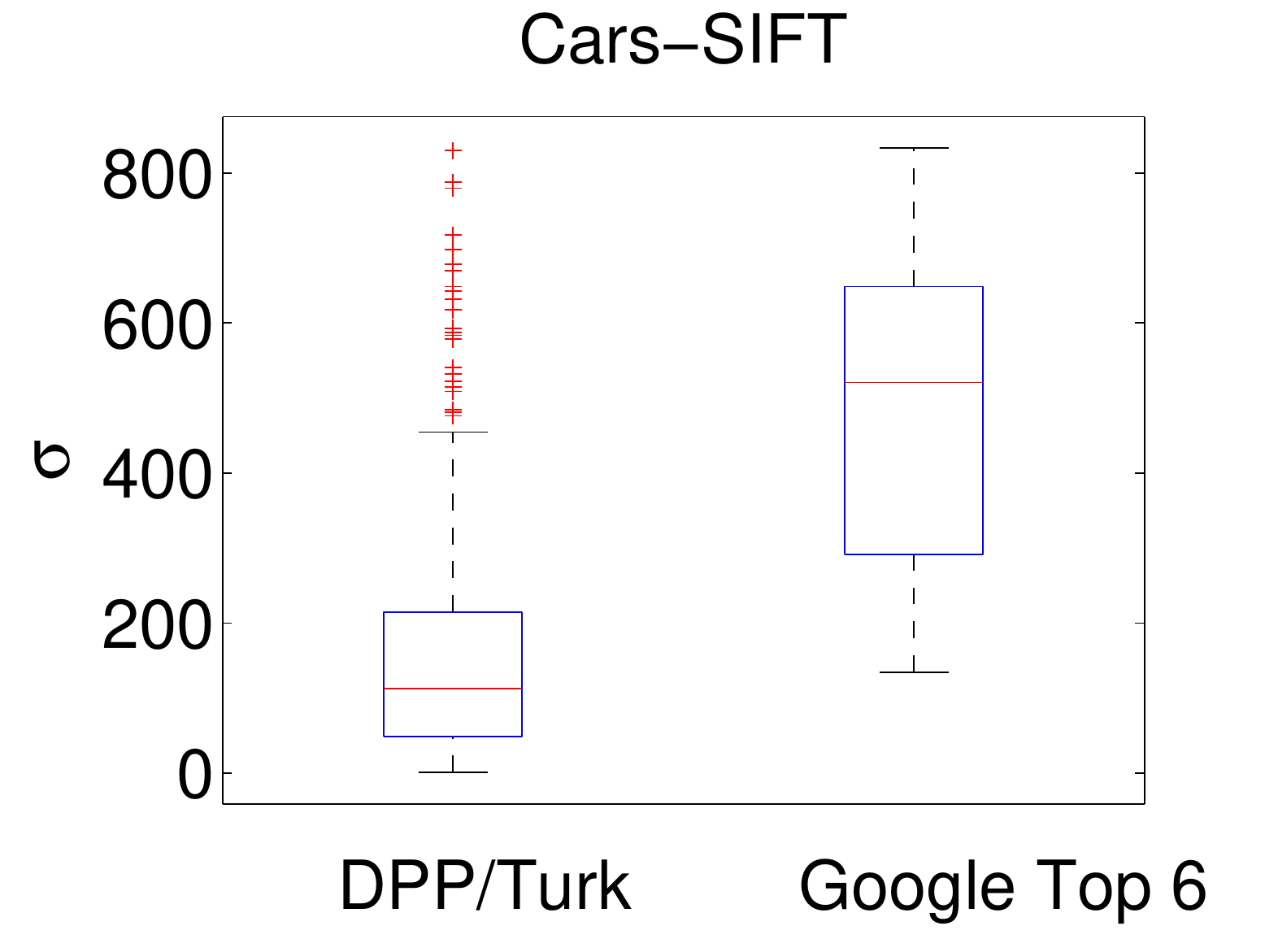} 
    \includegraphics[scale=0.16]{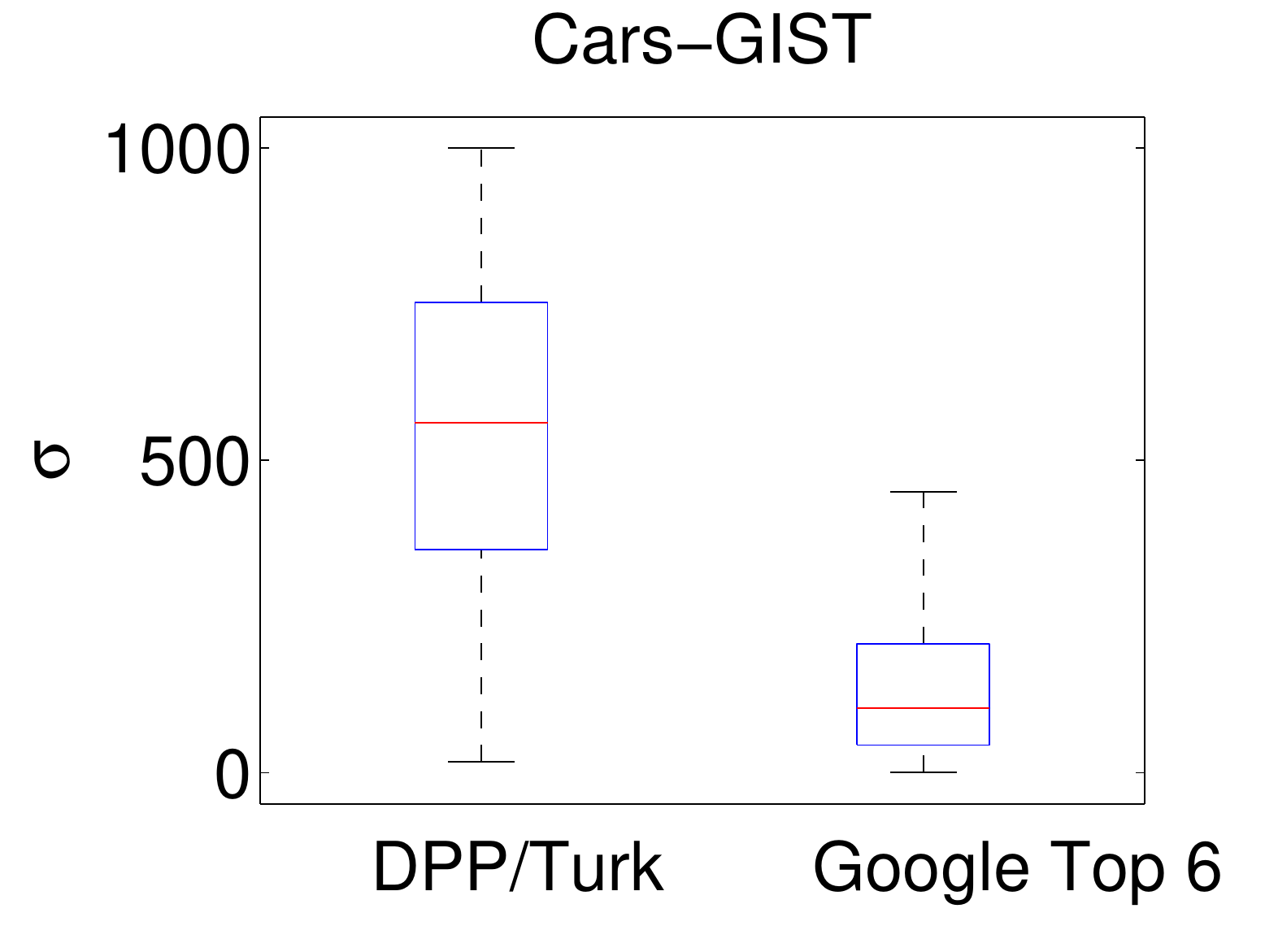}\\
    \includegraphics[scale=0.16]{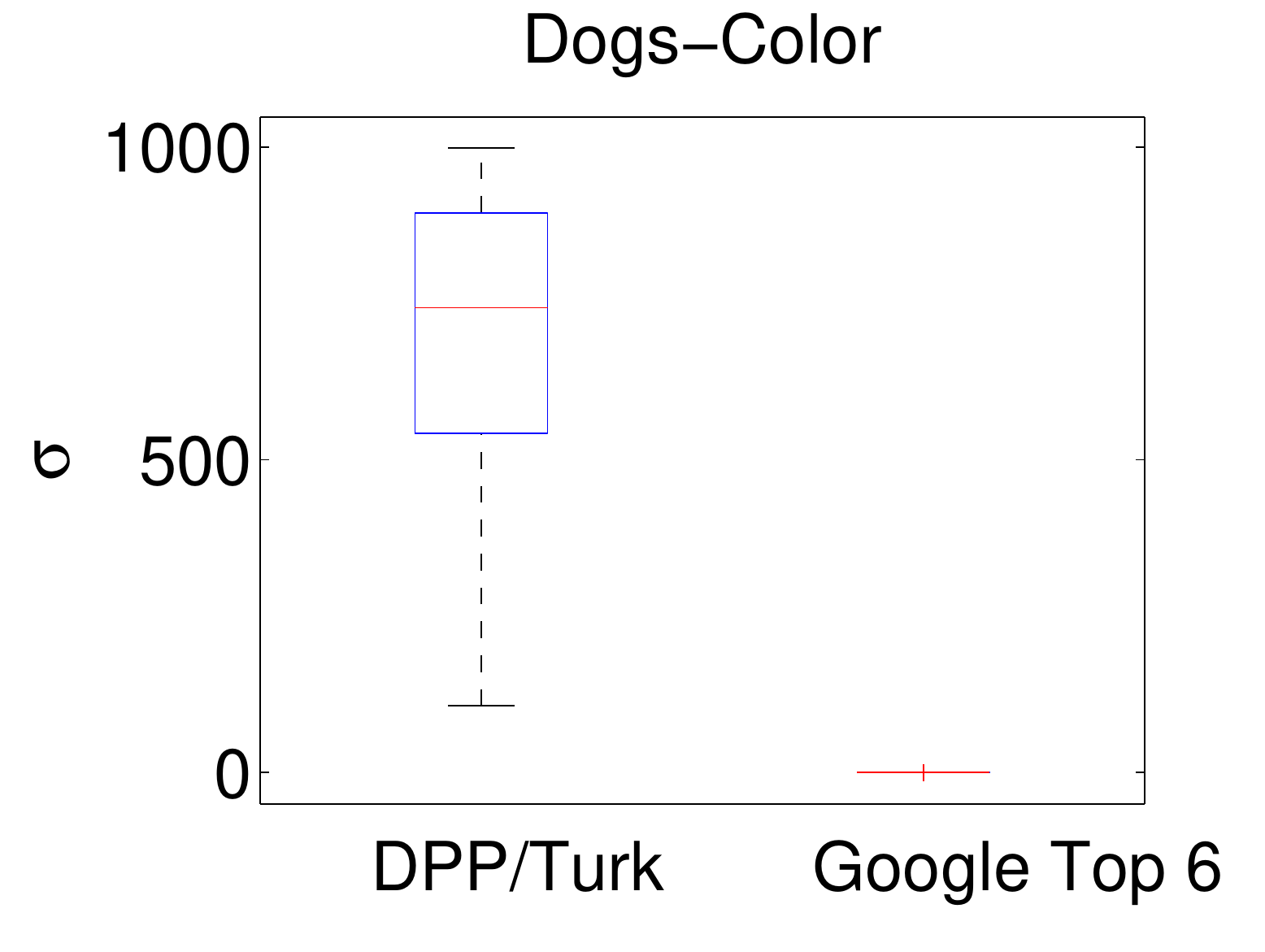} 
    \includegraphics[scale=0.16]{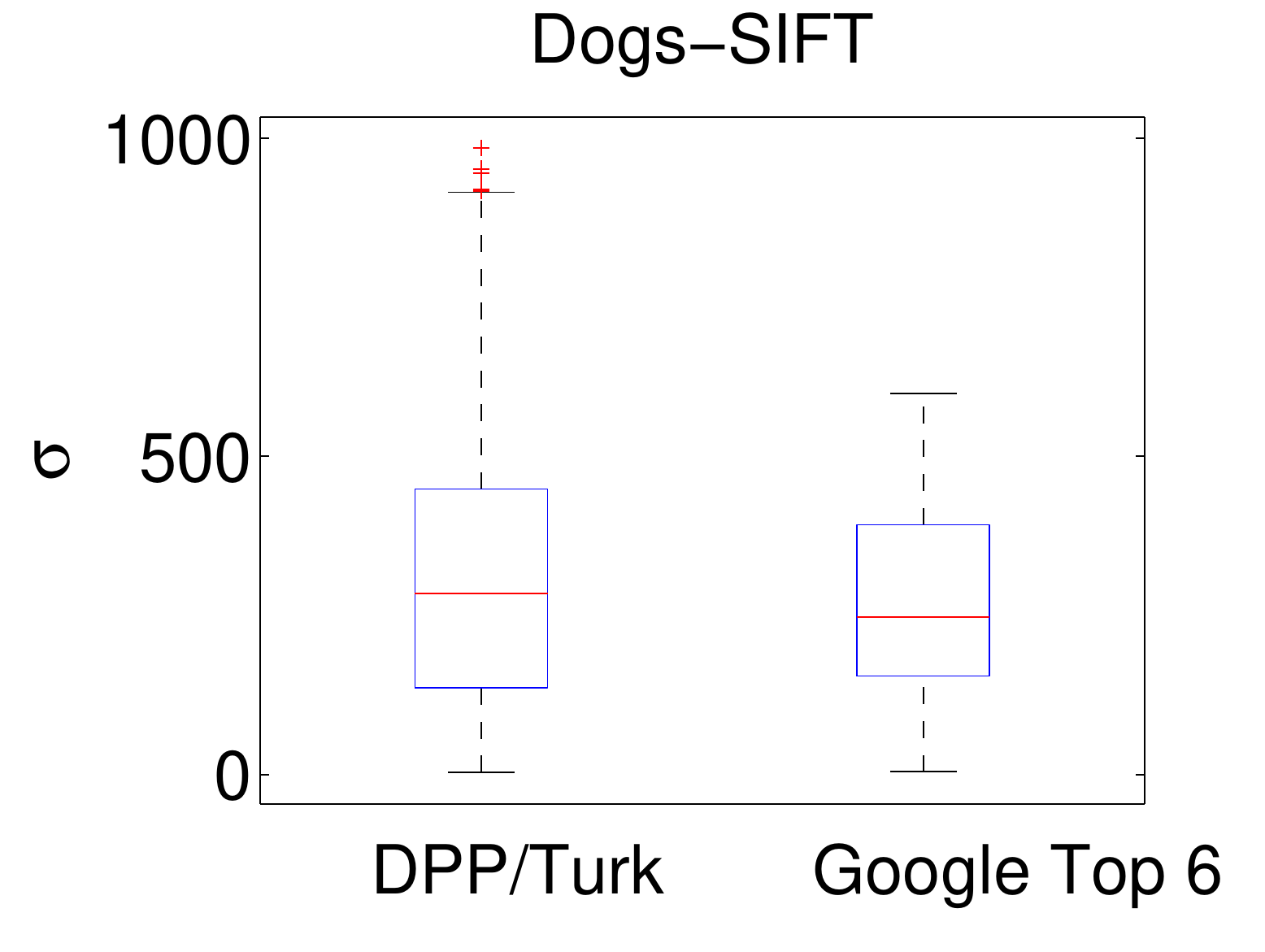} 
    \includegraphics[scale=0.16]{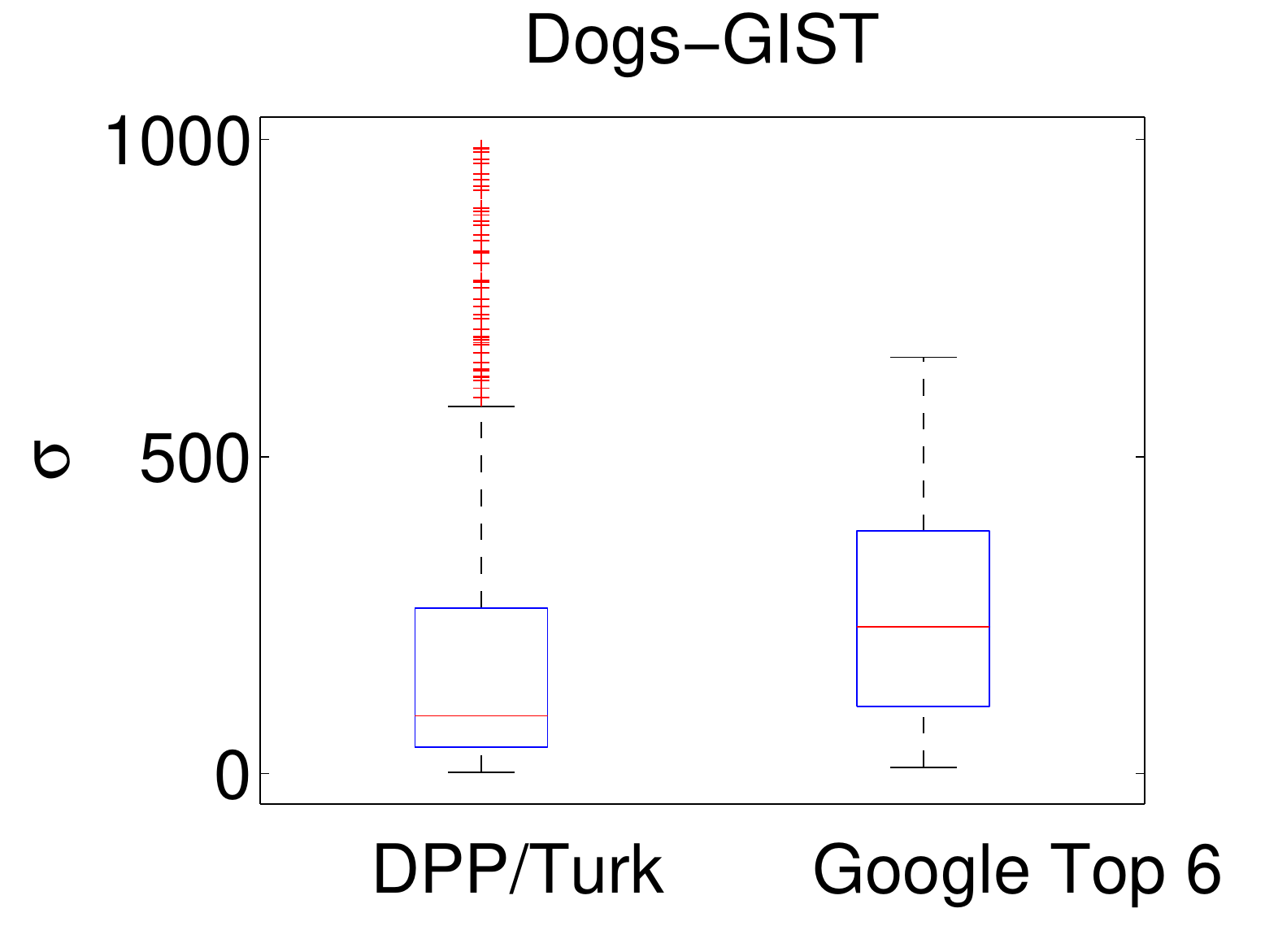}\\        
	\includegraphics[scale=0.16]{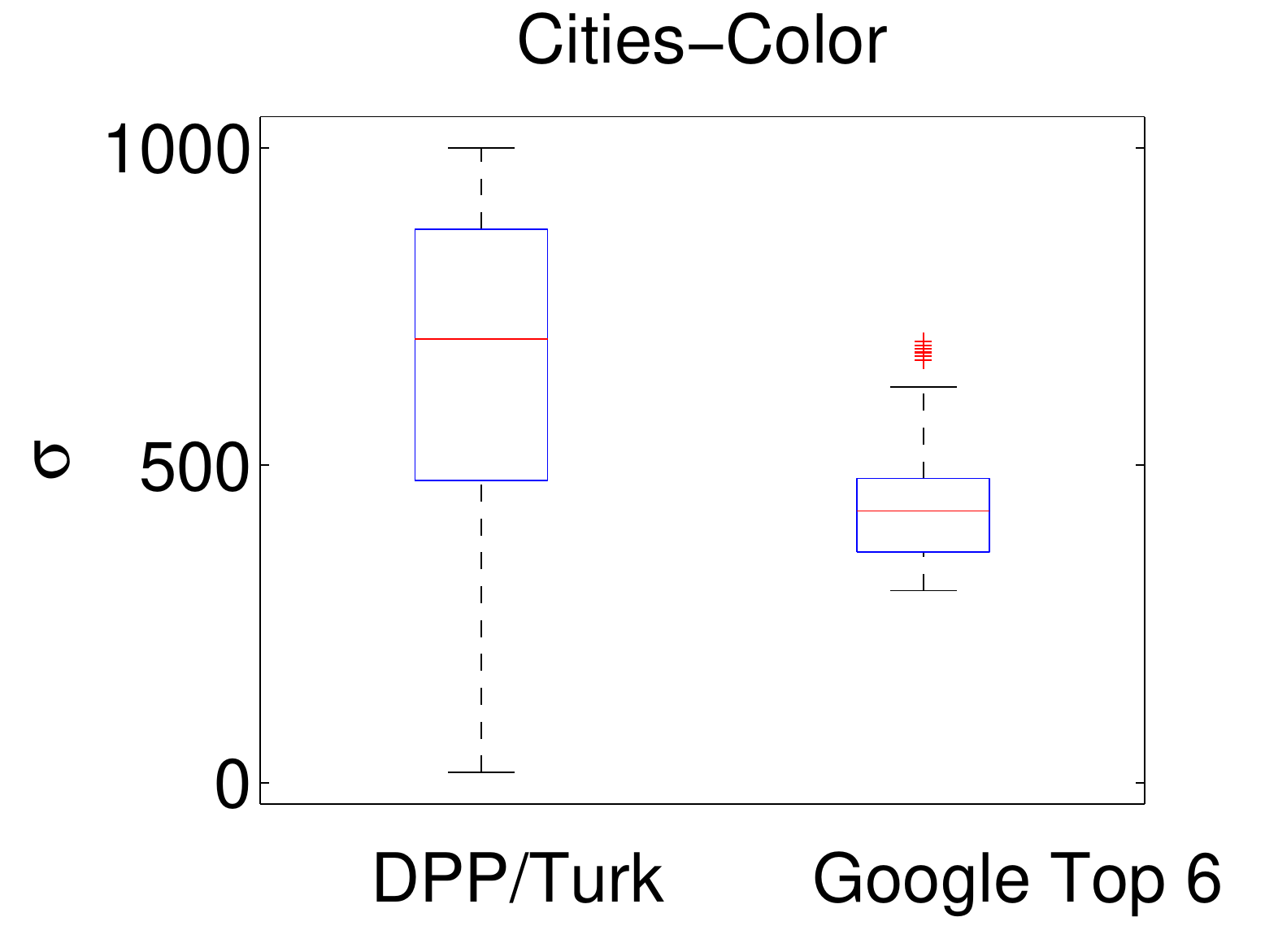} 
    \includegraphics[scale=0.16]{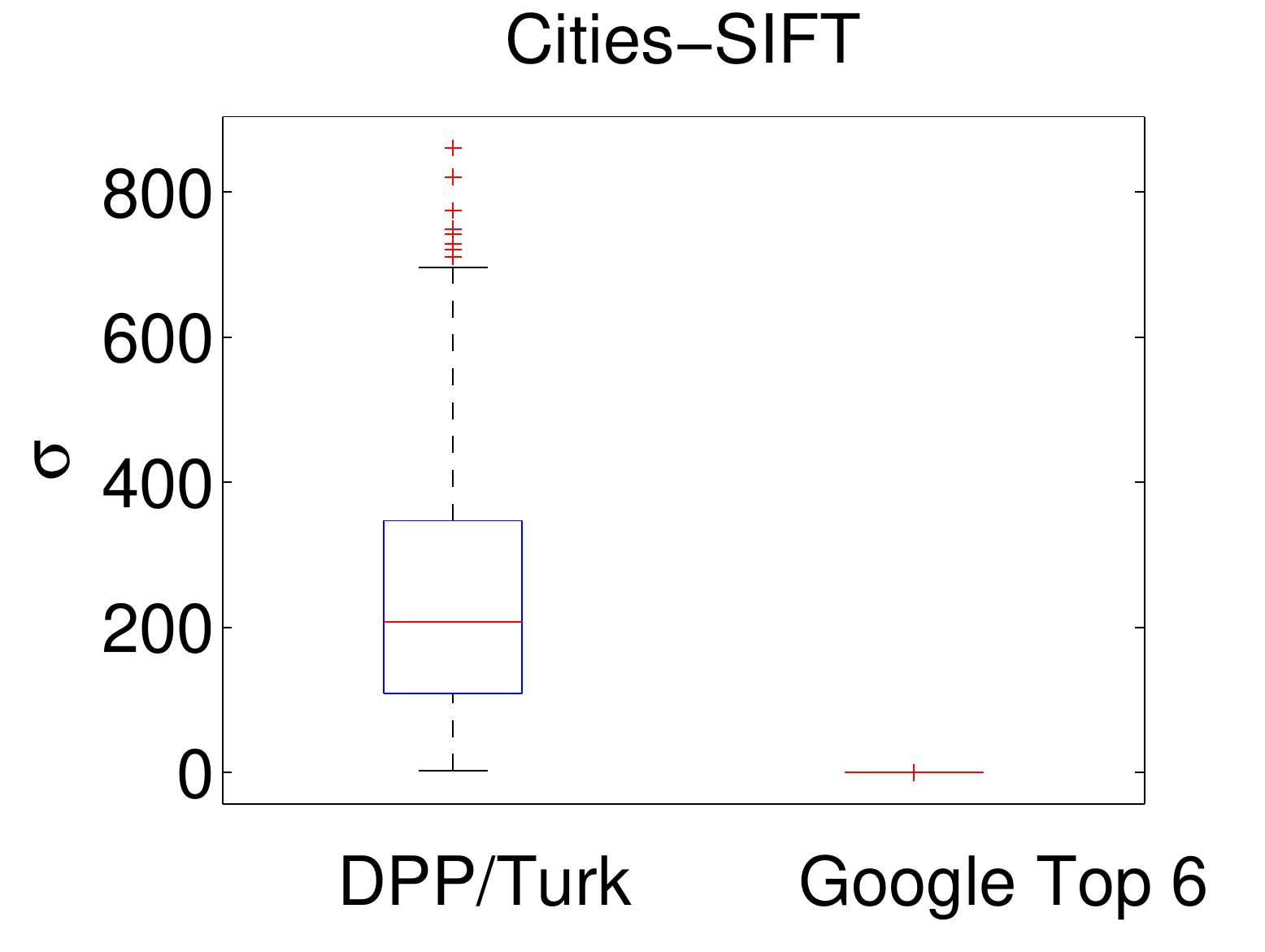} 
    \includegraphics[scale=0.16]{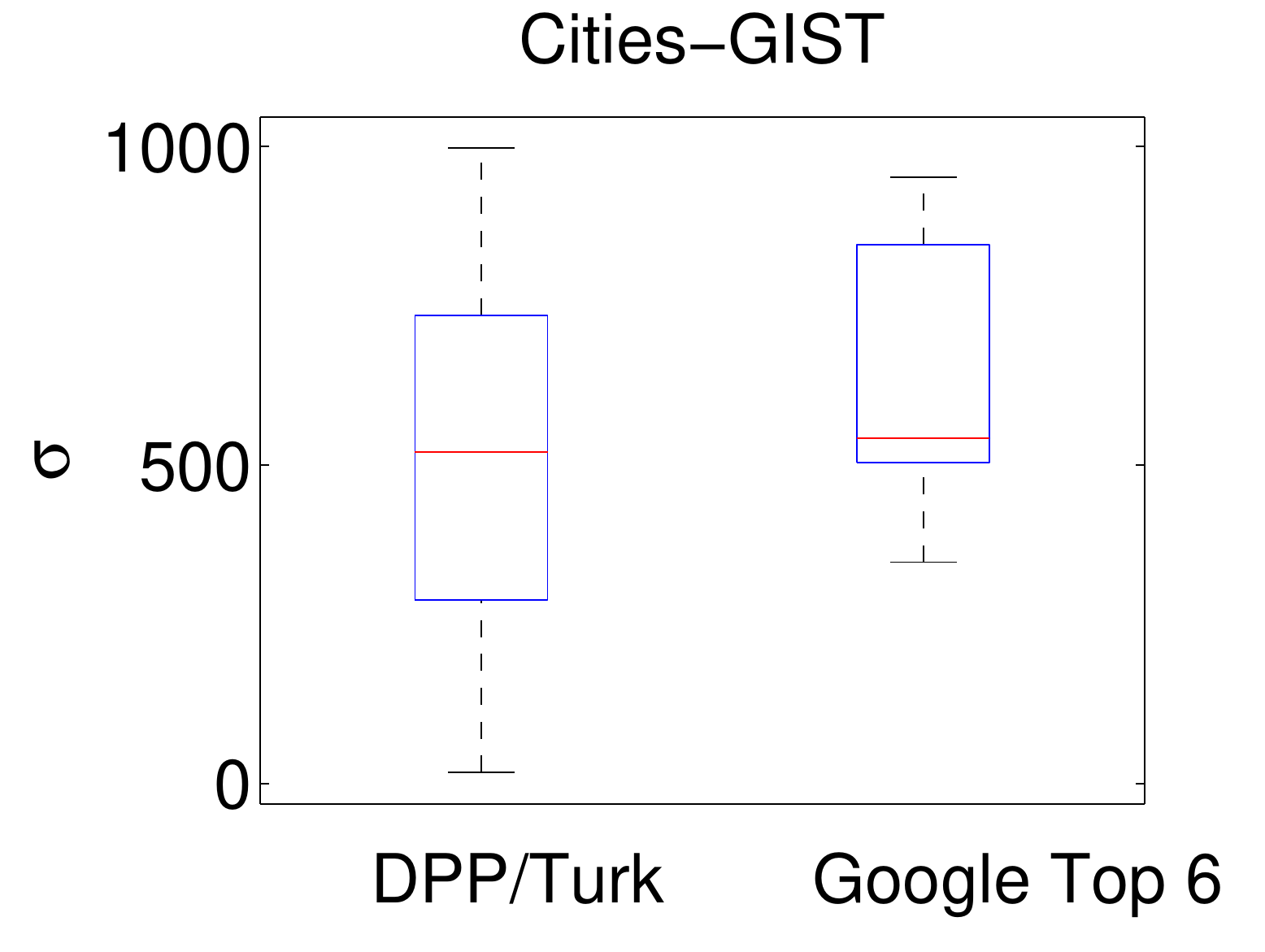}\\
\vspace{-10pt}
  \caption{For the image diversity experiment, boxplots of posterior samples of (rom left to right) $\sigma^{\textsf{cat}}_{\textsf{color}}$, $\sigma^{\textsf{cat}}_{\textsf{SIFT}}$ and $\sigma^{\textsf{cat}}_{\textsf{GIST}}$. Each plot shows results for human annotated sets (left) versus Google Top 6 (right). Categories from top to bottom: (a) \textsf{cars}, (b) \textsf{dogs} and (c) \textsf{cities}.}
\vspace{-10pt}
  \label{fig:imagesigma}
\end{figure}

\newpage
\bibliography{paper}
\bibliographystyle{icml2014}
\newpage
\appendix
\twocolumn[
\icmltitle{Supplementary Material: Learning the Parameters of Determinantal Point Process Kernels}

\icmlauthor{Raja Hafiz Affandi}{rajara@wharton.upenn.edu}
\icmladdress{University of Pennsylvania}
\icmlauthor{Emily B. Fox}{ebfox@stat.washington.edu}
\icmladdress{University of Washington}
\icmlauthor{Ryan P. Adams}{rpa@seas.harvard.edu}
\icmladdress{Harvard University}
\icmlauthor{Ben Taskar}{taskar@cs.washington.edu}
\icmladdress{University of Washington}

\vskip 0.2in
]
\section{Gradient for Discrete DPP }

Gradient ascent and stochastic gradient ascent provide attractive approaches in learning parameters, $
\Theta$  of DPP kernel $L(\Theta)$ because of their theoretical guarantees, but require knowledge of the gradient of the log-likelihood~$\mathcal{L}(\Theta)$.  In the discrete DPP setting, this gradient can be computed straightforwardly and we provide examples for discrete Gaussian and polynomial kernels here.
\begin{equation}
 \mathcal{L}(\Theta)=\sum_{t=1}^T \log(\det(L_{A^t}(\Theta)))-T\log(\det(L(\Theta)+I))
\end{equation}

\begin{eqnarray*}
\frac{d\mathcal{L}(\Theta)}{d\Theta}=&\sum_{t=1}^T\tr\left(L_{A^t}(\Theta)^{-1}\frac{dL_{A^t}(\Theta)}{d\Theta}\right)\\
&-T\tr\left((L(\Theta)+I)^{-1}\frac{dL(\Theta)}{d\Theta}\right)
\end{eqnarray*}

To find the MLE, we can perform gradient ascent
\begin{equation}
\Theta_i=\Theta_{i-1}+\eta\frac{dl(\Theta)}{d\Theta}
\end{equation}

In the following examples, we denote\\ $\bx_i=(x_i^{(1)},x_i^{(2)},\ldots,x_i^{(d)})$, where $d$ is the number of dimension.
\subsection{Example I: Gaussian Similarity with Uniform Quality}
$L(\Sigma)=\exp\{-(\x-\y)^{\top}\Sigma^{-1}(\x-\y)\}$

Denote  $G_{ij}^{(lm)}=L_{ij}\frac{(x_i^{(l)}-x_j^{(l)})(x_i^{(m)}-x_j^{(m)})}{2\Sigma_{lm}^2}$

Then,
\begin{eqnarray*}
\frac{d\mathcal{L}(\Sigma)}{d\Sigma_{lm}}=&\sum_{t=1}^T\tr\left(L_{A^t}(\Sigma)^{-1}G_{A^t}^{(lm)}\right)\\
&-T\tr\left((L(\Sigma)+I)^{-1}G^{(lm)}\right)
\end{eqnarray*}

\subsection{Example II: Gaussian Similarity with Gaussian Quality}
$L(\Gamma,\Sigma)\\=\exp\{-\x^{\top}\Gamma^{-1}\x-(\x-\y)^{\top}\Sigma^{-1}(\x-\y)-\y^{\top}\Gamma^{-1}\y\}$
Denote  $C_{ij}^{(lm)}=L_{ij}\frac{(x_i^{(l)}x_i^{(m)}+x_j^{(l)}x_j^{(m)})}{2\Gamma_{lm}^2}$ and $G_{ij}^{(lm)}$ as in previous example.

Then,
\begin{eqnarray*}
\frac{d\mathcal{L}(\Gamma, \Sigma)}{d\Gamma_{lm}}=&\sum_{t=1}^T\tr\left(L_{A^t}(\Sigma)^{-1}C_{A^t}^{(lm)}\right)\\
&-T\tr\left((L(\Sigma)+I)^{-1}C^{(lm)}\right)
\end{eqnarray*}
and $\frac{dl(\Gamma, \Sigma)}{d\Sigma_{lm}}$ is the same as the previous example. 

\subsection{Example III: Polynomial Similarity with Uniform Quality}
$L(p,q)=\left(\x^{\top}\y +p \right)^{q}$

Denote $R_{ij}=qL_{ij}^{\frac{q-1}{q}}$ and $U_{ij}=L_{ij}\log(L_{ij}^{\frac{1}{q}})$.

Then,
\begin{eqnarray*}
\frac{d\mathcal{L}(p,q)}{dp}=&\sum_{t=1}^T\tr\left(L_{A^t}(p,q)^{-1}R_{A^t}\right)\\
&-T\tr\left((L(p,q)+I)^{-1}R)\right)
\end{eqnarray*}

\begin{eqnarray*}
\frac{d\mathcal{L}(p,q)}{dq}=&\sum_{t=1}^T\tr\left(L_{A^t}(p,q)^{-1}U_{A^t}\right)\\
&-T\tr\left((L(p,q)+I)^{-1}U)\right)
\end{eqnarray*}

\section{Bayesian Learning}

In the main paper, we highlight two techniques: random-walk Metropolis-Hastings (MH) and slice sampling to sample from the posterior distribution. We present the pseudo algorithms here (Alg. \ref{alg:RWMH} and Alg. \ref{alg:slice}). 
\begin{algorithm}[tb]
 \caption{Random-Walk Metropolis-Hastings}
  \label{alg:RWMH}
\begin{algorithmic}
\STATE{\textbf{Input}:} Dimension: $D$, Starting point: $\Theta_0$, Prior distribution: $\Prob(\Theta)$, Proposal distribution $f(\hat{\Theta}|\Theta)$ with mean $\Theta$, Samples: $A^{1},\ldots,A^{T}]$.
\STATE $\Theta=\Theta_0$
\FOR{$i=0:(\tau-1)$}
	\STATE $\hat{\Theta}\sim f(\hat{\Theta}|\Theta_i)$               
           \STATE
$r=\left(\frac{\Prob(\hat{\Theta}|A^1,\ldots,A^T)}{\Prob(\Theta_i|A^1,\ldots,A^T)}\frac{f(\Theta_i|\hat{\Theta})}{f(\hat{\Theta}|\Theta_i)}\right)$ 
             \STATE $u\sim$ Uniform[0,1] 
           \IF{$u<\min\{1,r\}$}
	\STATE $\Theta_{i+1}=\hat{\Theta}$
\ENDIF
\ENDFOR
\STATE{\textbf{Output}: $\Theta_{0:\tau}$}
\end{algorithmic}
\end{algorithm}

\begin{algorithm}[tb]
 \caption{Univariate Slice Sampling}
  \label{alg:slice}
\begin{algorithmic}
\STATE{\textbf{Input:}} Starting point: $\Theta_0$, Initial width: $w$, Prior distribution: $\Prob(\Theta)$, Samples: $X=[X^{1},\ldots,X^{T}]$.
\STATE $\Theta=\Theta_0$
\FOR{$i=0:(\tau-1)$}
	\STATE $y\sim \textrm{Uniform}[0,\Prob(\Theta_i|A^1,\ldots,A^T)]$       
            \STATE $z\sim \textrm{Uniform}[0,1]$       
   \STATE $L=\Theta_i-z*\frac{w}{2}$
   \STATE $R=L+\frac{w}{2}$
   \WHILE{$y>\Prob(L|A^1,\ldots,A^T)$}
			\STATE $L=L-\frac{w}{2}$
   \ENDWHILE
   \WHILE{$y>\Prob(R|A^1,\ldots,A^T)$}
			\STATE $R=R+\frac{w}{2}$
   \ENDWHILE
   \STATE $\hat{\Theta}\sim\textrm{Uniform}[L,R]$
   \IF{$\Prob(\hat{\Theta}|A^1,\ldots,A^T)<y$}
		\WHILE{$\Prob(\hat{\Theta}|A^1,\ldots,A^T)<y$}
			\IF{$\hat{\Theta}>\Theta$} 
				\STATE $R=\hat{\Theta}$
			\ELSE
				\STATE $L=\hat{\Theta}$
			\ENDIF
			\STATE $\hat{\Theta}\sim\textrm{Uniform}[L,R]$
		\ENDWHILE
	\ENDIF
	\STATE $\Theta_{i+1}=\hat{\Theta}$      
\ENDFOR
\STATE{\textbf{Output}: $\Theta_{0:\tau}$}
\end{algorithmic}
\end{algorithm}

\begin{figure}
  \centering
     \includegraphics[scale=0.35]{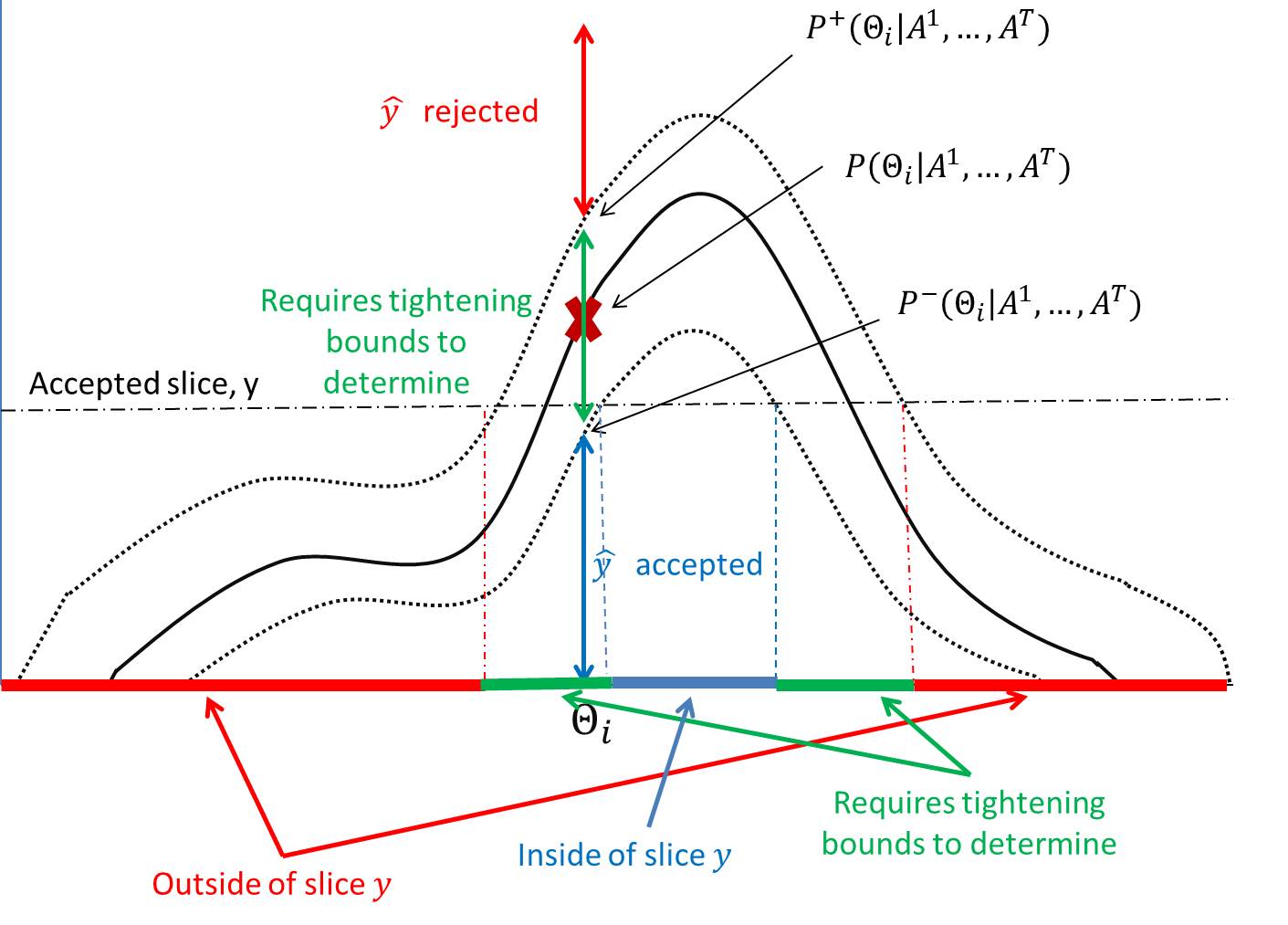}
  \caption{Illustration of slice sampling algorithm using posterior bounds. In the first step, a candidate slice $\hat{y}$ is generated. $\hat{y}$  is rejected if it is above the upper posterior bound and rejected if it is below the lower posterior bound. If $\hat{y}$ is in between the bounds, then the bounds are tightened until a decision can be made. Once a slice, $y$ is accepted, we need to sample new parameters inside the slice. To determine whether the endpoints of the interval or the new parameters are in the slice we decide that they are in the slice if the upper bound of posterior probability evaluated at the points are higher than the slice value and decide that they are outside of the slice if the lower bound of the posterior probability is lower than the slice value. Otherwise, we tighten the bounds until a decision can be made.}
  \label{fig:slice-EB}
\end{figure}

We also present the pseudo random walk MH algorithm for handling large-scale and continuous DPPs using posterior bounds in Alg. \ref{alg:RWMH-EB}. We also present and illustration of the slice sampling using posterior bounds in Figure \ref{fig:slice-EB}.

\begin{algorithm}[tb]
 \caption{Random-Walk Metropolis-Hastings with Posterior Bounds}
  \label{alg:RWMH-EB}
\begin{algorithmic}
\STATE{Input:} Dimension: $D$, , Starting point: $\Theta_0$, Prior distribution: $\Prob(\Theta)$,  Proposal distribution $f(\hat{\Theta}|\Theta)$ with mean $\Theta$, samples: $X=[X^{1},\ldots,X^{T}]$.
\STATE $\Theta=\Theta_0$
\FOR{$i=0:\tau$}
	\STATE $\hat{\Theta}\sim f(\hat{\Theta}|\Theta_i)$
	\STATE $r_{+}=\infty, r_{-}=-\infty$
\STATE $u\sim$ Uniform[0,1] 
\WHILE{$u\in[r_{-},r_{+}]$}               
\STATE $r^{+}=\left(\frac{\Prob^+(\hat{\Theta}|A^1,\ldots,A^T)}{\Prob^-(\Theta_i|A^1,\ldots,A^T)}\frac{f(\Theta_i|\hat{\Theta})}{f(\hat{\Theta}|\Theta_i)}\right)$
\STATE $r^{-}=\left(\frac{\Prob^-(\hat{\Theta}|A^1,\ldots,A^T)}{\Prob^+(\Theta_i|A^1,\ldots,A^T)}\frac{f(\Theta_i|\hat{\Theta})}{f(\hat{\Theta}|\Theta_i)}\right)$
\STATE Increase tightness on $\Prob^+$ and $\Prob^-$
\ENDWHILE
\IF{$u<\min\{1,r^-\}$}
	\STATE $\Theta_t=\hat{\Theta}$
\ENDIF
\ENDFOR
\STATE{Output: $\Theta_{0:\tau}$}
\end{algorithmic}
\end{algorithm}

\comment{\begin{algorithm}[tb]
 \caption{Univariate Slice Sampling with Posterior Bounds}
  \label{alg:slice-EB}
\begin{algorithmic}
\STATE{\textbf{Input:}} Starting point: $\Theta_0$, Initial width: $w$, Prior distribution: $\Prob(\Theta)$, Samples: $X=[X^{1},\ldots,X^{T}]$.
\STATE $\Theta=\Theta_0$
\FOR{$i=0:(\tau-1)$}
	\STATE $y\sim \textrm{Uniform}[0,\Prob(\Theta_i|A^1,\ldots,A^T)]$       
            \STATE $z\sim \textrm{Uniform}[0,1]$
                  
   \STATE $L=\Theta_i-z*\frac{w}{2}$
   \STATE $R=L+\frac{w}{2}$
   \WHILE{$y>\Prob(L|X^1,\ldots,X^T)$}
			\STATE $L=L-\frac{w}{2}$
   \ENDWHILE
   \WHILE{$y>\Prob(R|X^1,\ldots,X^T)$}
			\STATE $R=R+\frac{w}{2}$
   \ENDWHILE
   \STATE $\hat{\Theta}\sim\textrm{Uniform}[L,R]$
   \IF{$\Prob(\hat{\Theta}|X^1,\ldots,X^T)<y$}
		\WHILE{$\Prob(\hat{\Theta}|X^1,\ldots,X^T)<y$}
			\IF{$\hat{\Theta}>\Theta$} 
				\STATE $R=\hat{\Theta}$
			\ELSE
				\STATE $L=\hat{\Theta}$
			\ENDIF
			\STATE $\hat{\Theta}\sim\textrm{Uniform}[L,R]$
		\ENDWHILE
	\ENDIF
	\STATE $\Theta_{i+1}=\hat{\Theta}$      
\ENDFOR
\STATE{\textbf{Output}: $\Theta_{0:\tau}$}
\end{algorithmic}
\end{algorithm}}

\section{Proof of DPP/$k$DPP Bounds}

\begin{proposition}
\label{prop:DPP}
Let $\lambda_{1:\infty}$ be the eigenvalues of kernel $L$. Then 
\begin{equation}
\prod_{n=1}^M(1+\lambda_n)\le \prod_{n=1}^\infty(1+\lambda_n)
\end{equation}
and
\begin{equation}
\prod_{n=1}^\infty(1+\lambda_n)\le \exp\bigg\{ \textrm{\normalfont tr}(L)-\sum_{n=1}^M\lambda_n\bigg\}\left[\prod_{n=1}^M(1+\lambda_n)\right]~.
\end{equation}
\end{proposition}

\textbf{Proof:} The first inequality is trivial since the eigenvalues $\lambda_{1:\infty}$ are all nonnegative. 

To proof the second inequality, we use the AM-GM inequality: For any non-negative numbers, $\gamma_1,...,\gamma_M, (\prod_{n=1}^M\gamma_n)^{\frac{1}{M}}\le \sum_{n=1}^M\frac{\gamma_n}{M}$.

Let $\Lambda_M=\sum_{n=M+1}^\infty\lambda_n$ and $\gamma_n=1+\lambda_n$.
Then,
\begin{eqnarray*}
 \prod_{n=1}^\infty (1+\lambda_n)&=&\prod_{n=1}^\infty \gamma_n
=(\prod_{n=1}^M\gamma_n)(\prod_{n=M+1}^\infty\gamma_n)\\
&=&(\prod_{n=1}^M\gamma_n)(\lim_{l\rightarrow\infty}\prod_{n=M+1}^{M+l}\gamma_n)\\
&\le& (\prod_{n=1}^M\gamma_n)(\lim_{l\rightarrow\infty}(\sum_{n=M+1}^{M+l}\frac{\gamma_n}{l})^l)\\
&\le& (\prod_{n=1}^M(1+\lambda_n))\exp(\Lambda_M)~.\\
&&~~~~~~~~~~~~~~~~~~~~~~~~~~~~~~~~~~~~~~~\square
\end{eqnarray*}

\begin{proposition}
\label{prop:kDPP}
Let $\lambda_{1:\infty}$ be the eigenvalues of kernel $L$. Then 
\begin{equation}
e_k(\lambda_{1:M})\le e_k(\lambda_{1:\infty})
\end{equation}
and
\begin{equation}
e_k(\lambda_{1:\infty})\le \sum_{j=0}^k \frac{(\textrm{\normalfont tr}(L)-\sum_{n=1}^M\lambda_n)^j}{j!}e_{k-j}(\lambda_{1:M})~.
\end{equation}
\end{proposition}

\textbf{Proof:}
Let $e_k(\lambda_{1:m})$ be the $k$th elementary symmetric function:
$e_k(\lambda_{1:m}) =  \sum_{J \subseteq \{1,.\ldots,m\}, |J|=k} \prod_{j \in J} \lambda_j.$

Trivially, we have a lower bound since the eigenvalues $\lambda_{1:\infty}$ are non-negative:
$ e_k(\lambda_{1:m}) \le  e_k(\lambda_{1:n}) \quad \text{for } m \le n.$

For the upper bound we can use the Schur-concavity of elementary symmetric functions for non-negative arguments \cite{guan2006schur}.Thus for $\bar{\lambda}_{1:N}\prec\lambda_{1:N}$:
\begin{equation}
\sum_{i=1}^k \bar{\lambda}_n \leq \sum_{n=1}^k \lambda_n \quad \text{for } k=1,\dots,N-1
\end{equation}
and
\begin{equation}
\sum_{n=1}^N \bar{\lambda}_n = \sum_{n=1}^N \lambda_n
\end{equation}
we have  $e_k(\bar{\lambda}_{1:N})\geq e_k(\lambda_{1:N})$.

Now let $\Lambda_M=\sum_{n=M+1}^\infty$ and $\Lambda^N_M=\sum_{n=M+1}^N$. We consider 
\begin{equation}
\bar{\lambda}_{1:N}^{(M)} = (\lambda_1,\ldots,\lambda_M, \frac{\Lambda^N_M}{N-M},\ldots,\frac{\Lambda^N_M}{N-M}).
\end{equation}

Note that $\bar{\lambda}_{1:N}^{(M)}\prec\lambda_{1:N}$ and so $e_k(\bar{\lambda}_{1:N}^{(M)})\geq e_k(\lambda_{1:N})$ for $M<N$.

We now compute $e_k(\bar{\lambda}_{1:N}^{(M)})$. Note that for $e_k(\bar{\lambda}_{1:N}^{(M)})$, the terms in the sum are products of $k$ factors, each containing some of the $\lambda_{1:M}$ factors and some of the $\frac{\Lambda^N_M}{N-M}$  factors. The sum of the terms
that have $j$ factors of type $\frac{\Lambda^N_M}{N-M}$ is ${N-M\choose j} \left(\frac{\Lambda^N_M}{N-M}\right)^j e_{k-j}(\Lambda(m))$, so we have:
$$e_k(\bar{\lambda}_{1:N}^{(M)}) =   \sum_{j=0}^{k}  {N-M\choose j} \left(\frac{\Lambda^N_M}{N-M}\right)^j e_{k-j}(\lambda_{1:M})~.$$ 

Using ${N-M\choose j} \le \frac{(N-M)^j}{j!}$, we get

$$e_k(\bar{\lambda}_{1:N}^{(M)}) =   \sum_{j=0}^{k} \left(\frac{(\Lambda^N_M)^j}{j!}\right) e_{k-j}(\lambda_{1:M})~.$$

Letting $N\rightarrow \infty$, we get out upper bound
$$ e_k(\lambda_{1:\infty}) \le  \sum_{j=0}^{k}   \left(\frac{(\Lambda_M)^j}{j!}\right) e_{k-j}(\lambda_{1:M}) \quad \text{for } m \le n.$$

\section{Moments for Continuous DPP with Gaussian Quality and Similarity}

In the continuous case, given the eigendecomposition of the kernel operator, $L(\bx,\by)=\sum_{n=1}^\infty \lambda_n\phi_n(\bx)^*\phi_n(\by)$ (where $\phi_n(\bx)^*$ denotes the complex conjugate of the $n$th eigenfunction), the $m$th moment can be evaluated as
\begin{equation}
\label{eq:momCont}
E[\bx^m]=\int_{\Omega}\sum_{n=1}^\infty\frac{\lambda_n}{\lambda_n+1}\bx^m\phi_n(\bx)^2d\bx~.
\end{equation} 

Here we present the derivation for moments when 
\begin{equation}
\label{eq:qGaussian}
q(\bx)=\sqrt{\alpha}\prod_{d=1}^D\frac{1}{\sqrt{\pi\rho_d}}\exp\left\{-\frac{x_d^2}{2\rho_d}\right\}
\end{equation}
and
\begin{equation}
\label{eq:kGaussian}
k(\bx,\by)=\prod_{d=1}^D\exp\left\{-\frac{(x_d-y_d)^2}{2\sigma_d}\right\}, \bx,\by\in\mathbb{R}^D.
\end{equation}
In this case, the eigenvalues and eigenvectors of the operator $L$ are given by \citet{fasshauer2012stable},
\begin{equation}
\lambda_\mathbf{n}=\alpha\prod_{d=1}^D\sqrt{\frac{1}{\frac{\beta_d^2+1}{2}+\frac{1}{2\gamma_d}}}\bigg(\frac{1}{\gamma_d(\beta_d^2+1)+1}\bigg)^{n_d-1}, 
\end{equation}
and

$\phi_\n(\x)=$
\begin{equation}
\prod_{d=1}^D\left(\frac{1}{\pi\rho_d^2}\right)^{\frac{1}{4}}\sqrt{\frac{\beta_d}{2^{n_d-1}\Gamma(n_d)}}\exp\left\{-\frac{\beta_d^2x^2}{2\rho_d^2}\right\}H_{n_d-1}\left(\frac{\beta_d x_d}{\sqrt{\rho_d^2}}\right)~,
\end{equation} 

where $\gamma_d=\frac{\sigma_d}{\rho_d}$ , $\beta_d=(1+\frac{2}{\gamma_d})^{\frac{1}{4}}$ and $\n=(n_1,n_2,\ldots,n_D)$ is a multi index.

In the case of  DPPs (as opposed to $k$-DPPs), we can use the number of items as an estimate of the 0th moment. The 0th moment is given by $\sum_{\n=1} \frac{\lambda_\n}{1+\lambda_\n}$. Denote $\bx=(x_1,x_2,\ldots,x_d)$. The for higher moments, note that
\begin{eqnarray*}
E[x_j^m]&=\int_{\mathbb{R}}\sum_{n=1}^\infty\frac{\lambda_n}{\lambda_n+1}x_j^m\phi_n(\bx)^2dx_j\\
&=\sum_{n=1}^\infty\frac{\lambda_n}{\lambda_n+1}\int_{\mathbb{R}}x_j^m\phi_n(\bx)^2dx_j
\end{eqnarray*} 

Using the results of moment integrals involving a product of two Hermite polynomials \cite{paris2010asymptotics}, we get that
\begin{equation}
E[x_j^m]=\int_{\mathbb{R}^d}\sum_{\n}^\infty\frac{\lambda_\n}{\lambda_\n+1}(\frac{\sqrt{\rho_j}}{\sqrt{2}\beta_j})^m \wp_{\frac{m}{2}}(n_j-1) 
\end{equation}
for $m$ even and 0 otherwise. The polynomial $\wp_{\frac{m}{2}}(n_j-1)$ is given in Eq. (4.8) in \citet{paris2010asymptotics}.
For example, the second and fourth moments are given by
\begin{itemize}
\item[(i)] $E[x_j^2]=\sum_{\n}^\infty\frac{\lambda_\n}{\lambda_\n+1}(\frac{\sqrt{\rho_j}}{\sqrt{2}\beta_j})^2 (2n_j-1)$ 

\item[(ii)]$E[x_j^4]=\sum_{\n}^\infty\frac{\lambda_\n}{\lambda_\n+1}(\frac{\sqrt{\rho_j}}{\sqrt{2}\beta_j})^4 3(2n_j^2-2n_j+1)$
\end{itemize}

For a low dimensional setting, we can learn the parameters by using grid search such that the moments agree.

\section{Details on Simulation}

In the main paper, we use our Bayesian learning algorithms to learn parameters from (i) simulated data generated from a 2-dimensional isotropic discrete kernel ($\sigma_d=\sigma$, $\rho_d=\rho$ for $d=1,2$), (ii) nerve fiber data using 2-dimensional isotropic continuous kernel ($\sigma_d=\sigma$, $\rho_d=\rho$ for $d=1,2$) and (ii) image diversity data using 3600-dimensional discrete kernel with Gaussian similarity. In all of these experiments, we use weakly informative inverse gamma priors on $\sigma$,$\rho$ and $\alpha$. In particular, for all three parameters, we used the same priors for all three parameters
\begin{equation}
\Prob(\alpha)=\Prob(\rho)=\Prob(\sigma)= \textrm{Inv-Gamma}(0.001,0.001)
\end{equation}
We then learn the parameters using hyperrectangle slice sampling.

\section{Details on Image Diversity}

In studying the diversity in images, we extracted 3 different types of features from the images---color features, SIFT-descriptors \cite{lowe1999object,vedaldi2010vlfeat} and GIST-descriptors \cite{oliva2006building} described in the supplementary material. We describe these features below.

\textbf{Color}: Each pixel is assigned a coordinate in three-dimensional Lab color space. The colors are then sorted into axis-aligned bins, producing a histogram of either 8 (denoted color8) or 64 (denoted color64) dimensions.

\textbf{SIFT}: The images are processed  to obtain sets of 128-dimensional SIFT descriptors. These descriptors are commonly used in object recognition to identify objects in images and are invariant to scaling, orientation and minor distortions. The descriptors for a given category are combined, subsampled to set of 25,000, and then clustered using k-means into either 256 (denoted SIFT256) or 512 (denoted SIFT512) clusters. The feature vector for an image is the normalized histogram of the nearest clusters to the descriptors in the image.

\textbf{GIST}: The images are processed to obtain 960-dimensional GIST feature vectors that is commonly used to describe scene structure.

We also extracted the features above from the center of the images, defined as the centered rectangle
with dimensions half those of the original image. This yields a total of 10 different feature vectors. Since we are only concerned with the diversity of the images, we ensure that the quality across the images are uniform by normalizing each feature vector such that their $L_2$ norm equals to 1. We then combine the feature vectors into 3 types of features- color, SIFT and GIST.

For the Google top 6 images, we model the samples, $X^t_{Top6}$ as though they are generated from a 6-DPP with kernel $L^{subcat}(X^t)$. To highlight the effect of the human annotation in the partial results sets, we model the samples as though they are generated from a conditional 6-DPP. 

In general, given a partial set of observations A and $k$-DPP kernel $L$, we can define the conditional $k$-DPP probability of choosing a set B given the inclusion of set A (with $|A|+|B|=k$)as
\begin{equation}
\Prob_L^k(Y=A\cup B|A\in Y)\propto \det(L_B^A)
\end{equation}
with
\begin{equation}
\label{eq:condkernel}
 L^A=\left(\left[\left((L+I_{A^c}\right)^{-1}\right]_{A^c}\right)^{-1}-I
\end{equation}
where $I_{A^c}$ denotes the identity matrix with 0 for diagonal corresponding to elements in $A$. Here, following the $N \times N$ inversion, the matrix is restricted to rows and columns indexed by elements not in $A$, then inverted again. The normalizer is given by \citet{kulesza2012determinantal}.
\begin{equation}
\sum_{|Y'|=k-|A|}\det(L^A_{Y'})
\end{equation}

In our experiment, our samples can be seperated into the partial result sets and human annotations,
\begin{equation}
X^t_{\textrm{DPP+human}}=(A^t,b^t)
\end{equation}
 where $A^t$ is the partial result sets and $b^t$ is the human annotated result, we model the data from the conditional $6$-DPP $L^{subcat}(b^t|A^t)$. In this case, the likelihood is given by  
\begin{equation}
L^i(\Theta^{cat})=\frac{\det(L^{i~A^t}_{b_t}(\Theta^{cat}))}{\sum_{i=1}^NL^{i~A^t}_{x_i}(\Theta^{cat})}
\end{equation}
for each subcategory, $i$. That is, for each subcategory, i, we compute $L^i(\Theta^{cat})$ and use Eq. \ref{eq:condkernel} to compute the conditional kernel.

\end{document}